\documentclass[runningheads]{llncs}
\RequirePackage{silence}  
\usepackage{graphicx}
\usepackage{comment}
\usepackage{amsmath,amssymb}
\usepackage{color}
\usepackage{url}
\usepackage{hyperref}
\usepackage{caption}
\usepackage{overpic}
\definecolor{cvprblue}{rgb}{0.21,0.49,0.74}
\definecolor{lightboldcolor}{gray}{0.6}

\usepackage{makecell}
\usepackage{cleveref}
\usepackage{pifont}
\usepackage{tabularx}
\usepackage{orcidlink}

\newif\ifreview
\reviewtrue
\reviewfalse

\ifreview
	\usepackage{lineno}

	\linenumbers
\fi

\begin{document}


\def\SubNumber{2}

\def\GCPRTrack{Main Track}

\title{Confidence matters: Leveraging Multi-view Geometric Priors for GS-based Reconstruction}
\titlerunning{Leveraging Multi-view Geometric Priors for GS-based Reconstruction}

\ifreview
\else

	\author{\thanks{Corresponding Author}Hongyu Zhou\inst{1,2}\orcidlink{0000-0002-0099-643X} \and
	Zorah Lähner\inst{1,2}\orcidlink{0000-0003-0599-094X}}
	\authorrunning{H. Zhou et al.}
	\institute{$^{1}\text{University of Bonn}$ \hspace{16pt} $^{2}\text{Lamarr Institute for Machine Learning and Artificial Intelligence}$ \\
\email{\{hzhou, laehner\}@uni-bonn.de}
   }
   
\fi

\maketitle              

\begin{center}
    \centering
    \captionsetup{type=figure}
    \begin{overpic}[width=\linewidth, height=0.4\linewidth]{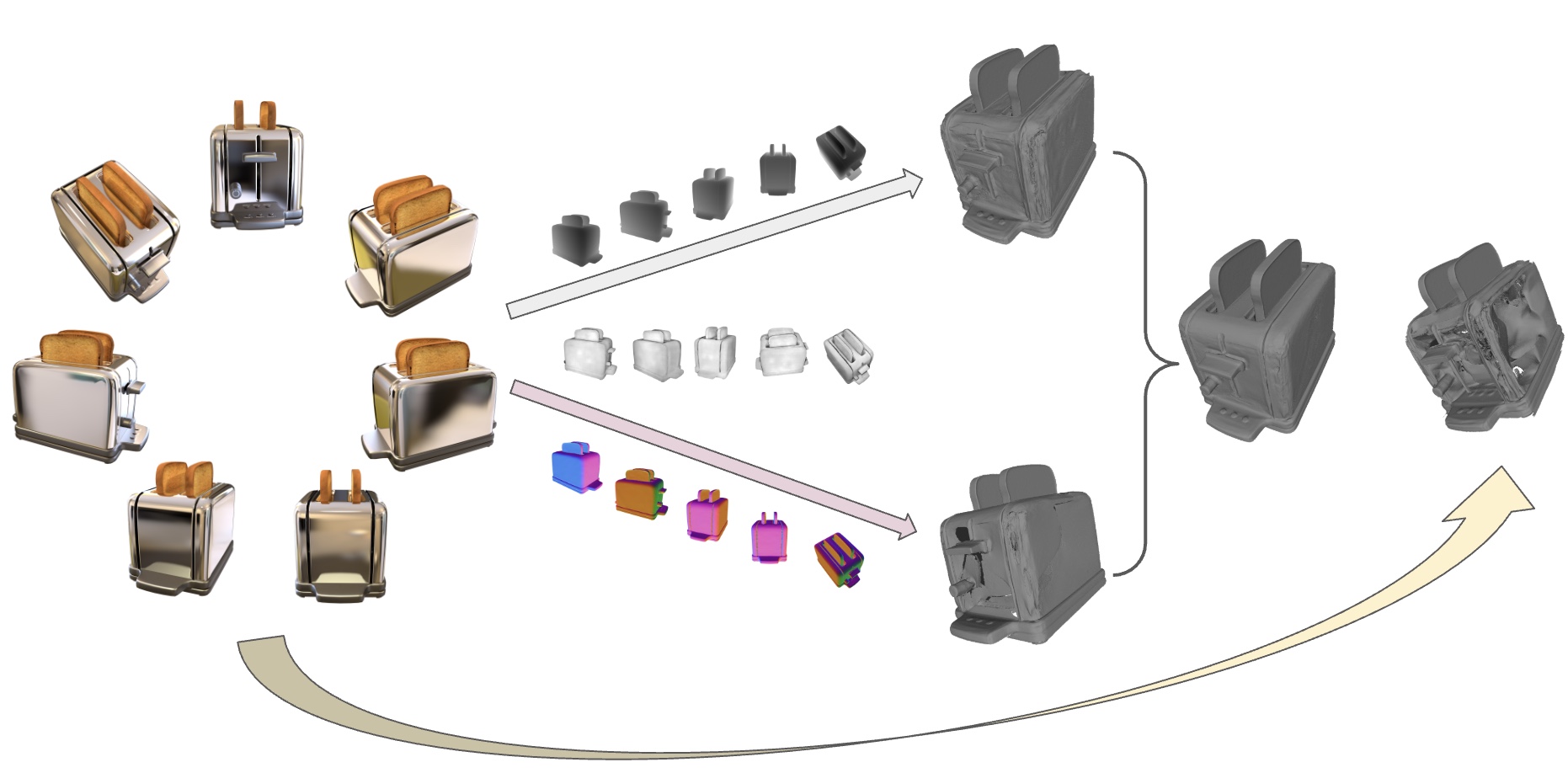}
    \put(49,-0.5){without geometric priors}
    
    \put(33,8.5){normal maps}
    \put(33,33.5){depth maps}
    \put(57,22){confidence}
    \put(61,20){maps}

    \put(75.5,15){improved}
    \put(73.5,12){reconstruction}
    \end{overpic}
    \captionof{figure}{\textbf{Overview.} We present a framework for integrating geometric priors (normal maps, depth maps) into 3D Gaussian splatting pipelines with the goal of increasing the geometric fidelity after extracting meshes. 
    Our analysis shows that multi-view priors, in combination with confidence maps, provide significant advantages in the presence of complex material properties, especially specularity. }
    \label{fig:teaser}
\end{center}%

\begin{abstract}
3D Gaussian splatting (3DGS) has emerged as a widely-used tool for novel view synthesis, offering real-time rendering in a sparse representation. However, the method's reliance on structure-from-motion initialization and photometric optimization can lead to suboptimal geometric reconstruction, particularly for objects with high specularity. 
In this work, we investigate the integration of geometric priors, in the form of predicted normal and depth maps, into the 3DGS framework to improve the reconstruction quality. 
We analyze the effect of incorporating these priors into GS-based methods and our evaluation reveals that multi-view predictions, as they are done by the recent visual geometry grounded transformer (VGGT), outperform single-view alternatives. 
A major factor is the existence of a confidence map for the estimations, which comes as a by-product of multi-view models and which can significantly improve the effectiveness of priors by weighting each prediction appropriately.  
Extensive experiments on standard benchmarks show consistent improvement in reconstruction quality and significant gains in complex scenes including specular objects. 

{\bf Website}: \href{https://github.com/Zero-4869/ConfidenceMattersGS}{https://github.com/Zero-4869/ConfidenceMattersGS}
\keywords{3D Gaussian Splatting  \and Reconstruction \and Multi-View.}
\end{abstract}

\section{Introduction}
\label{sec:intro}

3D Gaussian Splatting (3DGS)~\cite{kerbl3Dgaussians} has recently emerged as one of the major solutions for novel view synthesis. 
By representing scenes as collections of 3D Gaussians with learnable position, covariance, opacity and color, the method is able to produce high fidelity renderings while remaining computationally efficient due to the sparsity of its representation. 
During optimization, the parameters are updated using a photometric loss on the training views. 
While this is optimal for novel view synthesis, it limits the geometric reconstruction quality, especially in complex cases with reflective materials. 

\begin{figure}
    \centering
    \includegraphics[width=0.3\linewidth]{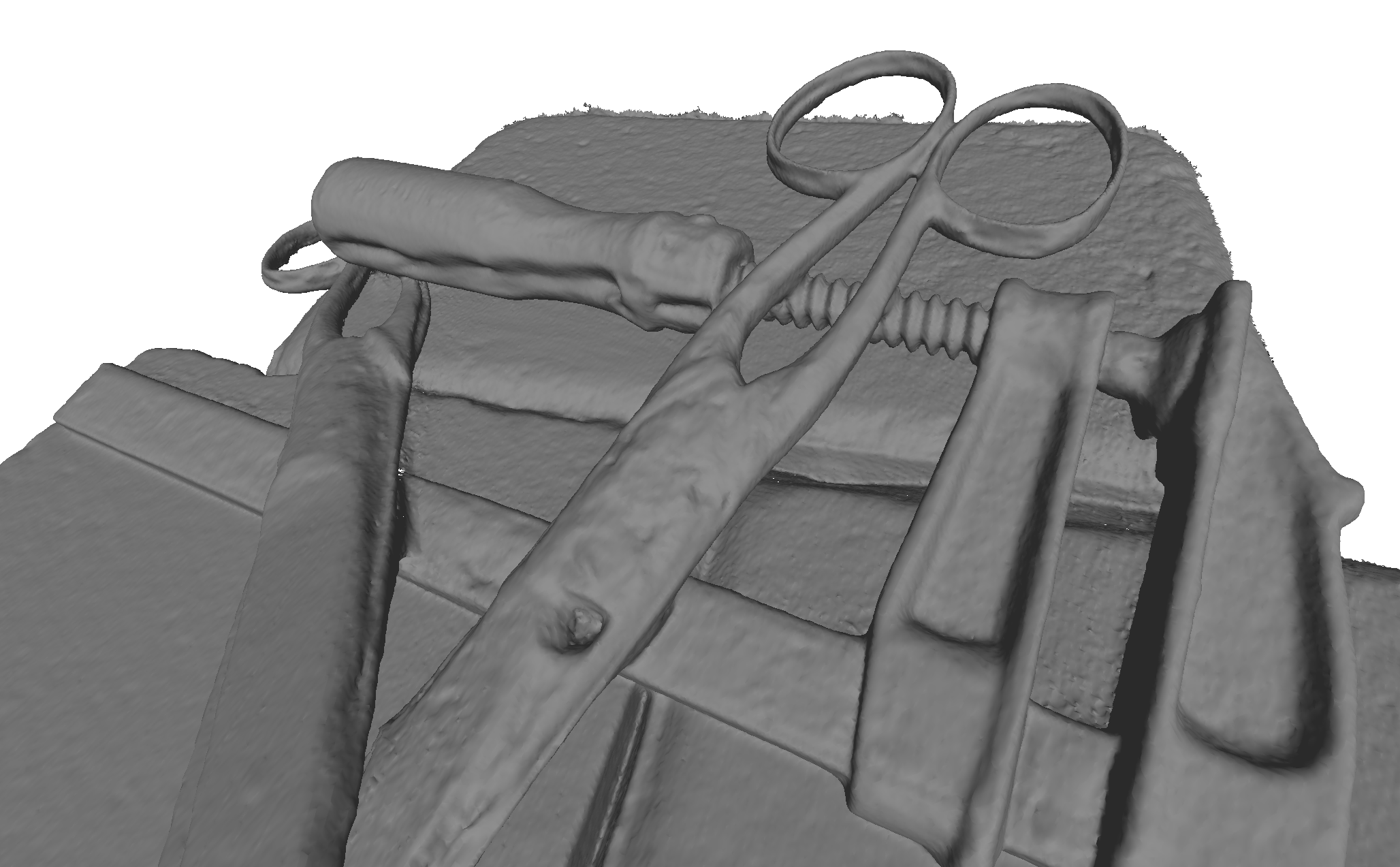}
    \begin{overpic}[width=0.3\linewidth]{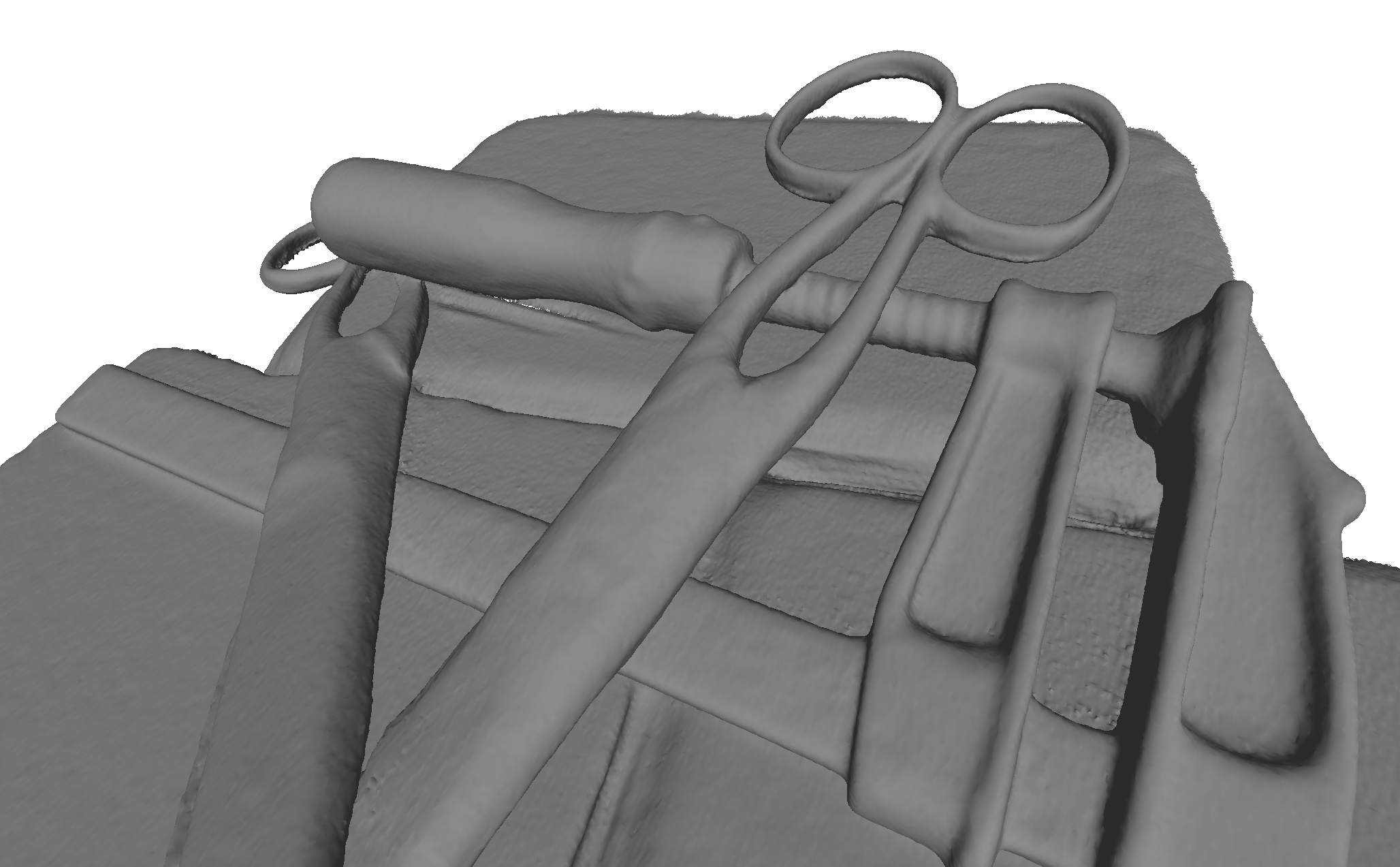}
    \put(63,41){\linethickness{5pt}\color{red}\circle{20}}
    \end{overpic}
    \includegraphics[width=0.3\linewidth]{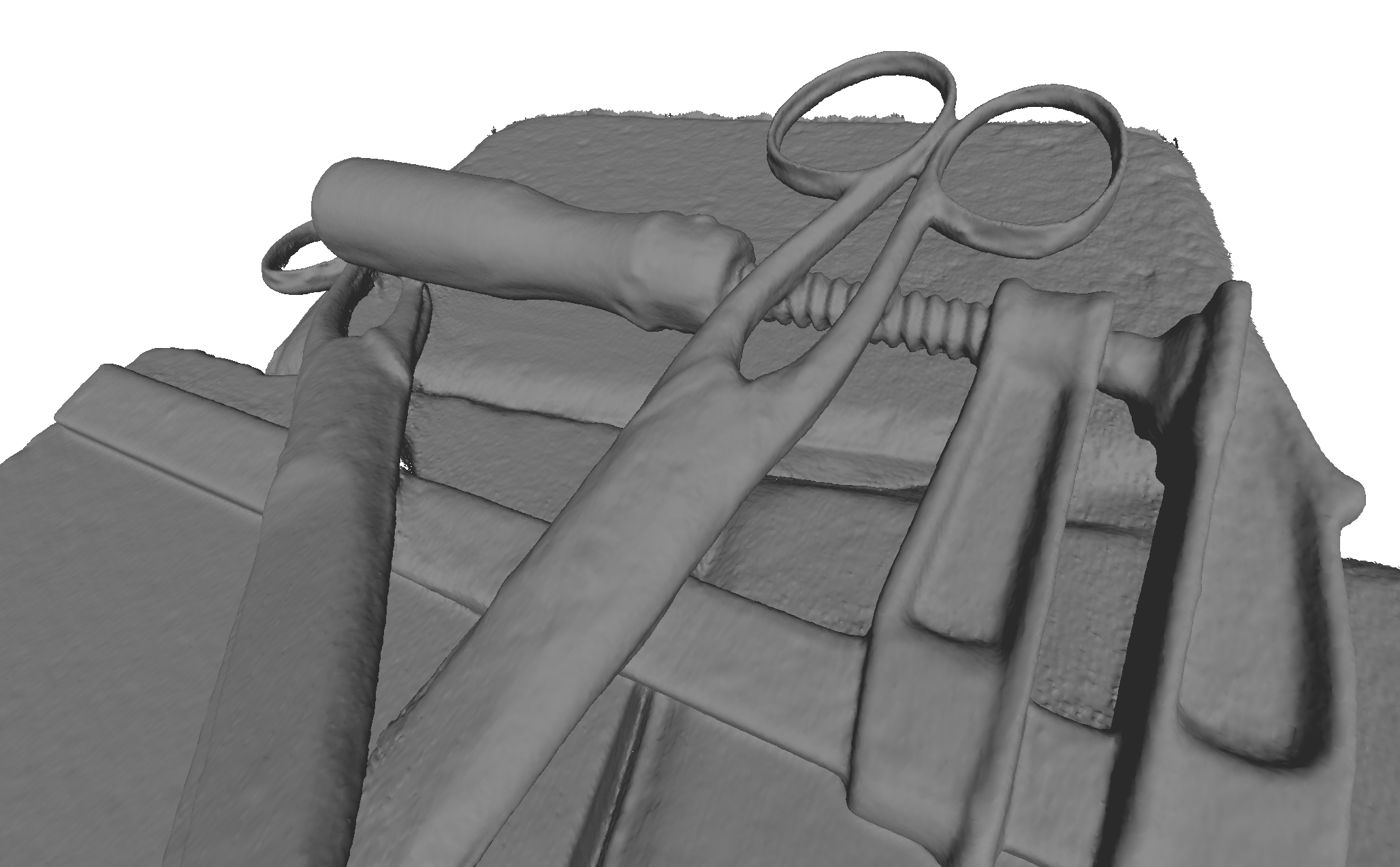}
    \caption{Monocular geometric priors often improve reconstruction quality but can also lead to degradation of results. (Left) PGSR. (Middle) PGSR with a StableNormal prior without any masking. (Right) PGSR + VGGT priors with confidence-based weighting (\Cref{sec:method}). While the pure monocular prior smooths out the results, the multi-view prior with proper weighting preserves all details.}
    \label{fig:motivation}
\end{figure}

Recent advances in depth and surface normal estimation have shown remarkable progress in predicting per-pixel geometric information in images. 
In the past, these methods were mostly monocular, and while they work well on most scenes, they struggle with reflective material and large texture-less areas which are hard to understand from a single view.
Recently, VGGT (Visual Geometry Grounded Transformer~\cite{wang2025vggt}) was introduced as a multi-view approach that can infer complex geometric properties from a set of images and generalize to a wide variety of scenes. 
Due to the inference from multiple images, predictions in complicated cases, especially for specularity, are more stable and it is possible to compute a confidence score based on how consistent the predictions from different views are. 
These are exactly the cases in which 3DGS reconstruction methods also struggle because, while the photometric loss can make up for reflection effects using the view dependency, the geometry needs to be consistent for all views. 
The incorporation of monocular geometric priors like StableNormal~\cite{ye2024stablenormal} is quite common in 3DGS and leads to significant improvements in reconstruction quality. 
However, the risk that this approach propagates errors introduced by the pretrained models, an issue that is in general unavoidable
\Cref{fig:motivation} shows an example in which the monocular prior actually degrades the reconstruction quality, a behavior our multi-view priors with confidence prediction can prevent.

In this work, we conduct a systematic analysis on the integration of geometric priors into 3D Gaussian splatting. 
The focus lies on the following questions:
1)~What is the best way to incorporate the priors into arbitrary 3DGS methods?
2)~Do multi-view priors provide considerable advantages over monocular geometric priors?
3)~It is unlikely that wrong predictions from direct inference can be completely avoided. Can masking of the geometric priors or a tailored optimization framework during regularization prevent huge failure cases due to propagation of these errors?

\paragraph{Contributions.} Our main contributions are as follows:
\begin{itemize}
    \item We propose a simple but effective way of leveraging confidence masks to integrate multi-view geometric priors into 3D Gaussian splatting methods and show its robustness across multiple datasets. 
    \item Our results determine that multi-view geometric priors in combination with confidence masks and a proper optimization framework lead to the best results, especially on specular objects.
    \item We provide an extensive experimental evaluation of combining different geometric priors with 3DGS methods on multiple datasets with varying focus, such as specularity. 
\end{itemize}

\section{Related Work} \label{sec:relatedwork}

We focus this section on works related to ours and refer to \cite{chen2024survey3dgaussiansplatting} for an in-depth survey of 3D Gaussian splatting. 

\subsection{3D Gaussian Splatting}

3D Gaussian splatting~\cite{kerbl3Dgaussians} is an efficient novel view synthesis method which represents the scene as a sparse collection of 3D Gaussian distributions with color and opacity information.
Even though it is fairly recent, it has been widely adopted for new applications~\cite{SplattingAvatar:CVPR2024,yan2024street,yan2023gsslam}, made even more efficient~\cite{chen2024mvsplat}, and combined with other representations, like signed distance functions, or operators to produce more accurate geometry~\cite{guedon2023sugar,yu2024gsdf,lyu20243dgsr,Huang2DGS2024,zhou2026laplace}. Other methods, such as PGSR~\cite{chen2024pgsr}, propose to use a better estimation of the depth and normal maps for single- and multi-view regularization during optimization.

However, while non-Lambertian effects are well captured in the view-dependent rendering equation, geometric reconstruction from 3DGS often struggles with specular effects which lead to artifacts~\cite{yao2025reflective}. 
This is because volume rendering does not capture surface properties well in these cases.
Directly regularizing (approximated) surface properties like normals improves the reconstruction by providing additional guidance in geometric domains~\cite{yu2024gaussian}.
Another solution is using separate geometric priors trained on large image datasets and predicted directly from images (also see \Cref{sub:rw:priors}).
Due to the learned prior knowledge, this produces reliable information about the surface even under very inconsistent visual cues or illumination settings~\cite{wang2025gsi3}.

\subsection{Reconstruction of Shiny Objects}
3D reconstruction of shiny objects has been well explored together with inverse rendering. 
NeRF-based methods such as VolSDF~\cite{yariv2021volume}, NeuS~\cite{wang2021neus}, RefNeuS~\cite{ge2023ref} and AniSDF~\cite{gao2025anisdf} leverage SDFs to reconstruct the geometry. While effective, they are time-consuming. 
3DGS-DR~\cite{ye20243d} and Ref-Gaussian~\cite{yao2025reflective} use normal propagation to reconstruct the geometry. MaterialRefGS~\cite{zhang2026materialrefgs} enforces 3D multi-view consistency to eliminate geometry aliasing, but BRDF modeling is required to reconstruct the geometry. 
Our work shows that, when used in the right way, geometry and confidence from multi-view vision models naturally improve the reconstruction for shiny materials.

\subsection{Geometric Priors} \label{sub:rw:priors}

Pretrained models for predicting geometric priors from images have been widely explored. 
StableNormal~\cite{ye2024stablenormal} is able to produce monocular normal estimation from a single image, and Depth Anything~\cite{yang2024depthanything} and MoGe~\cite{wang2025moge} show that nowadays accurate depth estimation is possible for almost all ordinary scenes. 
Monocular depth estimation is normally an ill-posed problem mathematically, but neural networks can learn priors from large datasets that allow meaningful predictions anyway~\cite{Ranftl2022}. Recently, more efforts have been put into multi-view geometry estimation. MASt3R~\cite{leroy2024grounding} and DUSt3R~\cite{wang2024dust3r} provide multi-view depth estimation, but they rely on pair-wise estimation that requires post-processing for further usage. 
On the other hand, VGGT~\cite{wang2025vggt} and Depth Anything 3~\cite{lin2025depth} enable multi-view depth estimation and point estimation through a single forward pass. In this paper, we will use these to integrate multi-view depth estimations into 3DGS to produce accurate and robust results.

\subsection{Geometric Regularization}
Using geometric priors in NeRF or Gaussian Splatting is a widely explored direction. MVPGS~\cite{xu2024mvpgs} uses the multi-view priors for Gaussian Splatting, but they are limited to rendering and do not study 3D reconstruction. 
MVG-splatting~\cite{li2025mvg} uses pretrained feature extractors for finer depth estimation and multi-view priors for 3D reconstruction. MonoSDF~\cite{yu2022monosdf} uses monocular depth and normal priors to supervise the SDF training for 3D reconstruction. GausSurf~\cite{wang2024gaussurf}, TSGS~\cite{li2025tsgs}, Reflections Unlock~\cite{song2025reflections} and 2DGS-Room~\cite{zhang20242dgs} leverage monocular normal estimation to train Gaussian splatting. MILo~\cite{guedon2025milo} uses geometric priors to supervise the joint training of Gaussian Splatting and a mesh. However, they do not study how artifacts in the predicted geometric priors affect the results. 

In this work, we study how multi-view geometric priors and confidence maps derived from them can improve 3DGS reconstruction quality, and propose an optimization strategy for geometric regularization with wide applicability. 

\section{Method} \label{sec:method}
\label{sec:method}
Monocular depth estimation and normal estimation have been widely used in enhancing the quality of 3D reconstruction for volumetric representations such as NeRF and Gaussian Splatting.
With the rise of pretrained models that support multi-view depth and normal estimation, the advantages of multi-view geometric priors over monocular geometric priors become worth exploring (\Cref{sub:method:multiview}). 
Recent investigations on VGGT~\cite{bratulic2025geometric} show that fundamental principles of multi-view geometry, such as correspondences and epipolar geometry, emerge in the representation, giving potential theoretical guarantees for using multi-view priors.
While it remains unclear how to leverage the priors while mitigating errors induced by inaccurate predictions, we propose to use the confidence map predicted by the model~(VGGT) as the weight for geometric regularization~(\Cref{sub:method:reg}).

\subsection{Multi-View Geometric Priors} \label{sub:method:multiview}

Monocular priors are easy to incorporate in 3DGS optimization as they can be directly applied and compared to the training images. 
But 3DGS optimization in its entirety is a multi-view approach and, thus, it would be natural to use this additional information in multi-view priors. 
Areas that are ambiguous in one training image (and lead to wrong prior estimation there) might be resolved in another, and this information should be propagated appropriately. 

\paragraph{VGGT.} Recently, the Visual Geometry Grounded Transformer (VGGT)~\cite{wang2025vggt} was introduced as a full multi-view inference method for a diverse set of geometric properties, including camera parameters and depth maps. VGGT is able to simultaneously generate depth estimations of all views and attach a confidence map $C$ to each estimation.

\paragraph{Alignment.} Even though the depth maps are multi-view consistent, they can have different scales due to scale ambiguity of the depicted objects. 
To overcome this, we assume that the relation between the estimated depth map and the ground truth depth is affine and will approximate this relationship by the affine transformation $l_i \in \mathbb{R}^{3\times3}$ for the $i$-th view. 
To avoid using error-prone estimations in the alignment, we take the mask $M$ to be $C_i > 0.5$, where $C_i$ is the normalized confidence map of the $i$-th view. 
Then the aligning function is 
\begin{equation}
    M\odot D_{gs} = l_{i}(M\odot D_{prior}) + t_{i},
\end{equation}
where $D_{gs}$ is the currently estimated depth from the 3DGS solution, $D_{prior}$ is the predicted depth from the geometric prior, and $t_i$ the displacement. 
The solution has the closed form
\begin{align}
    l_{i} &= \frac{(D_p^M - \bar{D_p^M})^\top(D_{gs}^M - \bar{D_{gs}^M})}{(D_p^M - \bar{D_p^M})^\top(D_p^M - \bar{D_p^M})}\\
    t_{i} &= \bar{D_{r}^M} - l_{i}\bar{D_p^M}
\end{align}
where $ D_p^M= M\odot D_{prior}$ is the masked depth prior and $D_{gs}^M = M\odot D_{gs}$ is the masked rendered depth.

A point map $P$ can be derived from the aligned depth priors and ground truth camera pose. We then compute the normal priors from $P$
\begin{equation}
\label{eq:normal_prior}
N_{prior}(p) = \frac{(P_1 - P_0)\times(P_3 - P_2)}{\|(P_1 - P_0)\times(P_3 - P_2)\|}.
\end{equation}
where $P_0, P_1, P_2, P_3$ are the neighboring points (up, left, down, right, respectively) of the pixel $p$ in the point map.
 
\subsection{Geometric regularization} \label{sub:method:reg}
In this section, we explain how to incorporate the priors as regularization.
An overview of the approach is shown in~\Cref{fig:teaser}. 
We use PGSR~\cite{chen2024pgsr} as the base GS method to which we apply all regularization. 
PGSR flattens the splats to obtain a stronger geometric signal, which means the normal $n_g$ of a Gaussian splat can be approximated by the lowest magnitude
variance direction. 
The normal map is then rendered with $\alpha$-blending as 
\begin{equation}\label{rendered_normal}
    N_{gs} = \sum_i \alpha_{g_i} n_{g_i}\prod_{j<i}(1 - \alpha_{g_j})
\end{equation}
Similarly, we estimate the unbiased depth map $D_{gs}$ in accordance with \cite{chen2024pgsr}. 
Specifically, the depth to the splat's plane is computed as $d_g = (\mu_g - T_c)^T n_g$. Then, the distance map $D$ is rendered and the unbiased depth estimation $D_{gs}$ at pixel $p$ can be computed as 
\begin{equation}\label{rendered_distance}
    D = \sum_i \alpha_{g_i} d_{g_i}\prod_{j<i}(1 - \alpha_{g_j}), \hspace{4pt} D_{gs}(p) = \frac{D(p)}{N_{gs}(p)\cdot K^{-1}\tilde{p}}
\end{equation}
where $K$ is the intrinsic matrix of the camera and $\tilde{p}$ is the homogeneous coordinate of $p$. The depth normal, denoted by $N_d$, can be computed similarly to \cref{eq:normal_prior}, with the point map $P^d$ derived from $D_{gs}$ and the camera pose:
\begin{equation}
\label{eq:depth_normal}
N_{d}(p) = \frac{(P^d_1 - P^d_0)\times(P^d_3 - P^d_2)}{\|(P^d_1 - P^d_0)\times(P^d_3 - P^d_2)\|}.
\end{equation}

\paragraph{Confidence-based weighting.}
Even the most advanced geometric prior estimators will struggle with certain predictions, for example due to complicated scene structure, complex material interactions or unseen instances. 
In the optimal case, the wrongly predicted pixel would be masked out and the optimization falls back to the default photometric losses only. 
Unlike monocular vision models such as StableNormal~\cite{ye2024stablenormal} or DepthAnything~\cite{yang2024depth}, VGGT explicitly models uncertainty during training.
The predicted confidence map $C$ is inversely proportional to an uncertainty map; thus, we use the normalized confidence map as part of the geometric regularization.   
 
The geometric priors are applied by geometric regularization in the form of 

\begin{align}
\begin{aligned}
\mathcal{L}_{geo} = &\lambda_{normal}(W_{conf}\odot \|N_{gs} - N_{prior}\| + W_{conf}\odot \|N_d - N_{prior\|})\\
&+ \lambda_{depth}W_{conf}\|D_{gs} - D_{prior}\|, \label{eq:lgeo}
\end{aligned}
\end{align}
where $W_{conf} = C_{i}^{f(k)}$, $C_i$ is the normalized confidence map of view $i$, and $f(k)$ is a function of the training iteration that controls the decaying speed of the weight.

With $\mathcal{L}_{pgsr} = \mathbf{L}_{svgeom} + \mathbf{L}_{mvgeom}+\mathbf{L}_{mvrgb}$, the geometric regularization used in PGSR~\cite{chen2024pgsr}, we conclude our geometric regularization as 
\begin{equation}
    \mathcal{L} = \mathcal{L}_{pgsr} + \mathcal{L}_{geo}
\end{equation}

\begin{table*}[htb]
    \centering               
    \setlength{\tabcolsep}{0.4pt}
    \fontsize{7.5pt}{9pt}\selectfont
    \begin{tabularx}{\textwidth}{X|c c c c c c c c c c c c c c c|c}
    
    \hline
          \textbf{DTU} & 24 & 37 & 40 & 55 & 63 & 65 & 69 & 83 & 97 & 105 & 106 & 110 & 114 & 118 & 122 & Mean\\ \hline
          NeuS\cite{wang2021neus} & 1.00 & 1.37 & 0.93 & 0.43 & 1.10 & 0.65 & 0.57 & 1.48 & 1.09 & 0.83 & 0.52 & 1.20 & 0.35 & 0.49 & 0.54 & 0.84 \\
          Neuralangelo\cite{li2023neuralangelo} & 0.37 & 0.72 & 0.35 & 0.35 & 0.87 & 0.54 & 0.53 & 1.29 & 0.97 & 0.73 & 0.47 & 0.74 & 0.32 & 0.41 & 0.43 & 0.61 \\ \hline 
          2DGS\cite{Huang2DGS2024} & 0.48 & 0.91 & 0.39 & 0.39 & 1.01 & 0.83 & 0.81 & 1.36 & 1.27 & 0.76 & 0.70 & 1.40 & 0.40 & 0.76 & 0.52 & 0.80 \\ 
          GOF\cite{yu2024gaussian} & 0.50 & 0.82 & 0.37 & 0.37 & 1.12 & 0.74 & 0.73 & 1.18 & 1.29 & 0.68 & 0.77 & 0.90 & 0.42 & 0.66 & 0.49 & 0.74 \\
           PGSR\cite{chen2024pgsr} & 0.34 & 0.58 & \textbf{0.29} & \textbf{0.29} & 0.78 & 0.58 & 0.54 & \textbf{1.01} & 0.73 & \textbf{0.51} & 0.49 &  0.69 & \textbf{0.31} & \textbf{0.37} & 0.38 & 0.53 \\ 
            GausSurf\cite{wang2024gaussurf} & 0.35 & 0.55 & 0.34 & 0.34 & \textbf{0.77} & 0.58 & 0.51 & 1.10 & 0.69 & 0.60 & 0.43 & \textbf{0.49} & 0.32 & 0.40 & 0.37 & \textbf{0.52} \\
         Ours & \textbf{0.33} & \textbf{0.53} & 0.33 & 0.34 & 0.83 & \textbf{0.54} & \textbf{0.48} & 1.14 & \textbf{0.65} & 0.62 & \textbf{0.38} & 0.59 & \textbf{0.31} & \textbf{0.37} & \textbf{0.36} & \textbf{0.52} \\ \hline
    \end{tabularx}


\vspace{0.2cm}
\centering
\setlength{\tabcolsep}{6.4pt}
\begin{tabular*}{\textwidth}{@{\extracolsep{\fill}}l| c c c c |c}
    \hline
        \textbf{Shiny Blender} &  Car & Coffee & Helmet & Toaster & Mean\\ \hline
        NeuS~\cite{wang2021neus}    & 1.10 & 1.99 & 1.12 & 2.87 & 1.77 \\
        RefNeuS~\cite{ge2023ref} & 0.80 & 0.99 & 0.38 & 1.47 & 0.91 \\
        RefNeRF~\cite{verbin2024ref} & 14.93 & 12.24 & 42.87 & 29.48 & 24.88 \\
        AniSDF~\cite{gao2025anisdf} & 0.70 & 1.14 & 1.15 & 0.41 & 0.85\\ \hline

        VGGT~\cite{wang2025vggt} & 2.15	& 2.56	& 1.43	& 4.54 & 2.67 \\\hline 
        3DGS-DR~\cite{ye20243d} & 1.33 & 1.35 & 1.65 & 4.80 & 2.28\\
        Ref-Gaussian~\cite{yao2025reflective} & \bf{0.93} & 1.32 & 3.44 & 2.86 & 2.14\\
        PGSR~\cite{chen2024pgsr} & 1.93 & 1.24 & 2.76 & 6.99 & 3.23\\ 
        
        Ours & 1.20 & \bf{1.00} & \bf{0.87} & \bf{1.87} & \bf{1.23}\\ \hline
    
    \end{tabular*}

\vspace{0.2cm}
\centering
\begin{tabular*}{\textwidth}{@{\extracolsep{\fill}}l|c c c c c c|c}
    \hline
     \textbf{TnT}   & Barn & Courthouse & Caterpillar & Meetingroom & Ignatius & Truck & Mean \\ \hline
    2DGS~\cite{Huang2DGS2024} & 0.36 & 0.13 & 0.23 & 0.16 & 0.44 & 0.26 & 0.30 \\
    GOF~\cite{yu2024gaussian} & 0.51 & 0.28 & 0.41 & 0.28 & 0.68 & 0.58 & 0.46 \\
    PGSR~\cite{chen2024pgsr} & \bf{0.66} & 0.21 & 0.41 & 0.29 & \bf{0.80} & 0.60 &\bf{0.50} \\ 
    GausSurf~\cite{wang2024gaussurf} & 0.50 & \bf{0.30} & 0.42 & \bf{0.39} & 0.73 & \bf{0.65} & \bf{0.50} \\
    Ours & 0.65 & 0.19 & \bf{0.44} & 0.37 & 0.74 & 0.61 & \bf{0.50} \\ \hline
    \end{tabular*}
 \caption{Quantitative evaluation on DTU reporting chamfer distance, Shiny Blender reporting MAE~(degree) and TnT reporting F1. Ours using VGGT including the confidence weighting works best on average in all of the dataset.}
        \label{exp:dtu} \label{exp:shinyblender} \label{exp:tnt}
\end{table*}

\section{Experiments} \label{sec:experiments}

We provide an experimental evaluation of our proposed method and put the focus on the geometric reconstruction quality.
This is often neglected in novel view synthesis work which focuses on rendering; thus, we also report PSNR to show that our method performs on par with previous approaches in rendering quality as well. 
Quantitative results can be found in \Cref{subsec:quantitative}, qualitative results in \Cref{subsec:qualitative}, and an ablation study in \Cref{subsec:ablation}.

\subsection{Datasets} \label{sec:exp:datasets}
We conduct our experiments on a variety of datasets with different challenges:
\begin{itemize}
    \item \textbf{DTU~\cite{jensen2014large}}: a multi-view capture of 80 different everyday objects, some of which are specular. We use the 15 most commonly used objects. 
    \item \textbf{Shiny Blender~\cite{verbin2022refnerf}}: a dataset of 8 objects with highly reflective surfaces rendered from Blender. We use 4 objects: car, coffee, helmet and toaster, to keep consistency with previous works~\cite{ge2023ref}~\cite{gao2025anisdf}.
    \item \textbf{Tanks and Temples (TnT)~\cite{knapitsch2017tanks}}: a dataset of 14 different larger indoor and outdoor scenes with varying material complexity. We use 6 commonly used scenes: Barn, Courthouse, Caterpillar, Meetingroom, Ignatius, and Truck.
\end{itemize}

\subsection{Baselines} \label{sub:exp:baselines}
We evaluate the effect of our proposed geometric prior techniques on state-of-the-art reconstruction methods for each dataset. 
\begin{itemize}
    \item \textbf{2DGS~\cite{Huang2DGS2024}} and \textbf{GOF~\cite{yu2024gaussian}}. Two GS methods focusing on 3D reconstruction used as baselines for DTU and TnT.
    \item \textbf{PGSR~\cite{chen2024pgsr}}. A robust state-of-the-art method for reconstruction with 3DGS. We build our model based on PGSR and compare with it on DTU, Shiny Blender and TnT. 
    \item \textbf{GausSurf~\cite{wang2024gaussurf}}. The method is built upon PGSR with monocular geometric estimation. Since the code is not available, we will just copy the results reported in the paper and compare with it on DTU and TnT quantitatively.
    \item \textbf{3DGS-DR~\cite{ye20243d}} and \textbf{Ref-Gaussian~\cite{yao2025reflective}}. These are state-of-the-art 3DGS-based methods specifically designed for reflective scenes. We compare our method with them on Shiny Blender quantitatively and qualitatively. 
\end{itemize}
For DTU, Shiny Blender and TnT, we use PGSR~\cite{chen2024pgsr} as the baseline as it is a robust state-of-the-art method for reconstruction with 3DGS.

\paragraph{Evaluation metrics. }
We report results in the default metrics for each dataset to make them comparable among publications. 
To be consistent with previous methods, DTU uses Chamfer distance, TnT uses F1 score, and Shiny Blender uses MAE in degrees on normal estimation. We follow PGSR for the metric computation on DTU and TnT and follow RefNeuS~\cite{ge2023ref} for computing MAE on Shiny Blender. For results in novel view synthesis, we report PSNR.

\begin{figure*}
\setlength{\tabcolsep}{1pt} 
    \renewcommand{\arraystretch}{1} 
    \centering
    \begin{tabular}{cccccc}
    {\footnotesize \text{VGGT}} & {\footnotesize \text{PGSR}} & {\footnotesize \text{3DGS-DR}} & {\footnotesize \text{Ref-Gaussian}} & {\footnotesize Ours} & {\footnotesize GT}\\
    \includegraphics[width=0.16\linewidth]{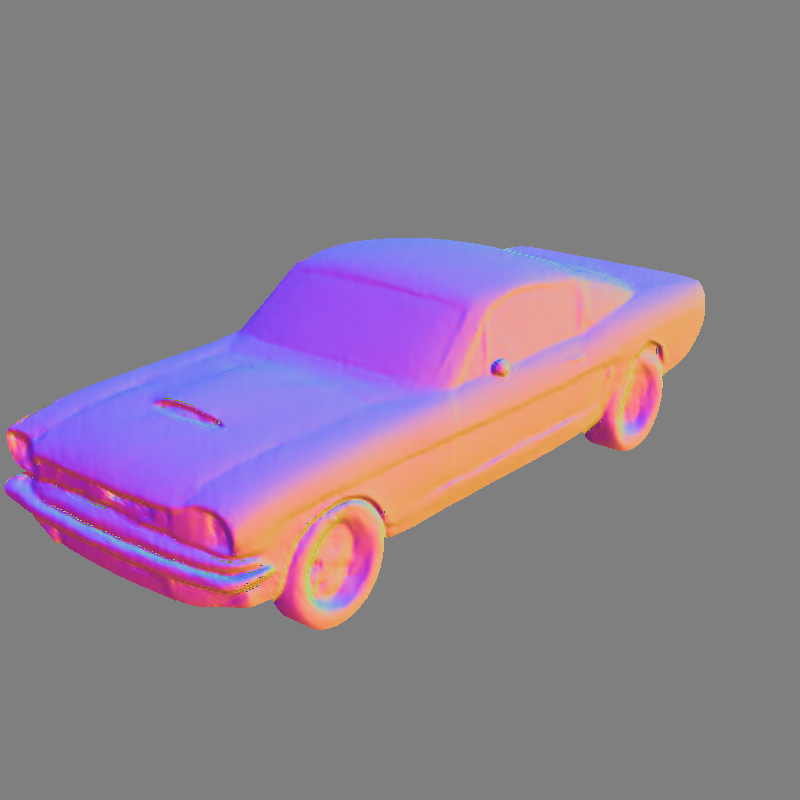} &
    \includegraphics[width=0.16\linewidth]{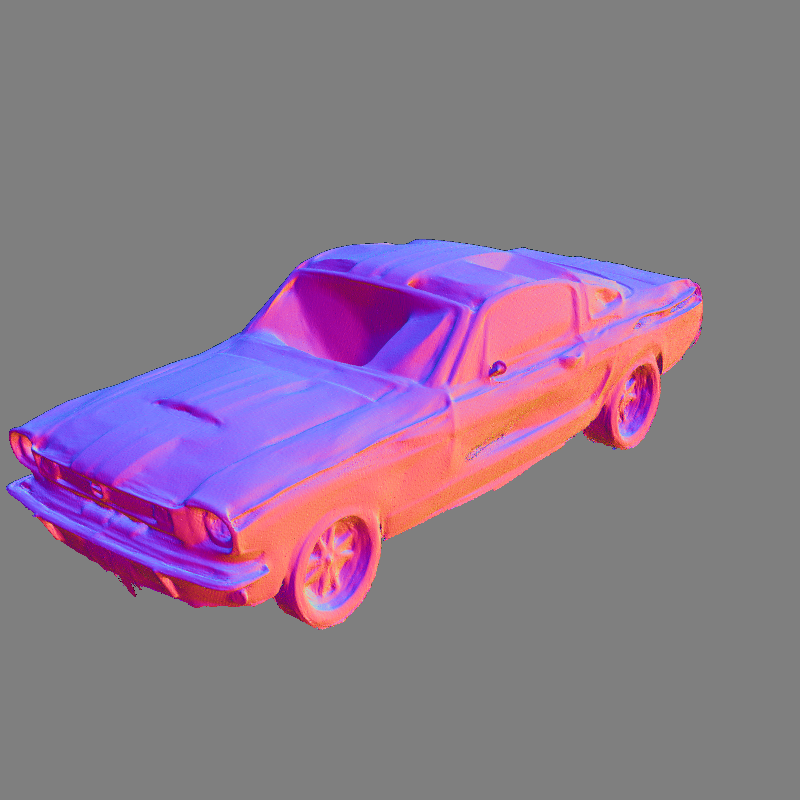} &
    \includegraphics[width=0.16\linewidth]{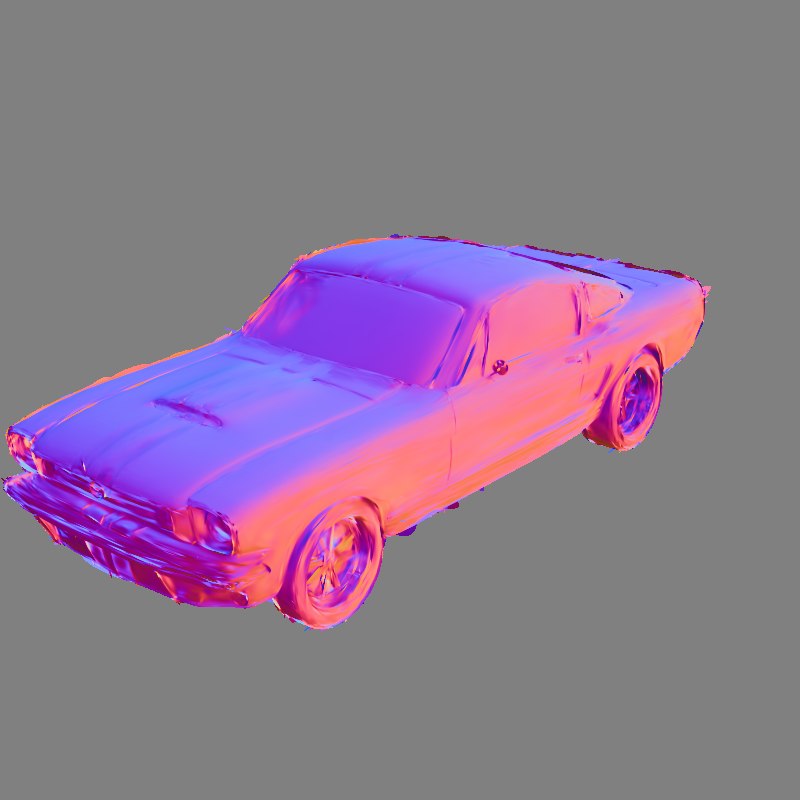} &
    \includegraphics[width=0.16\linewidth]{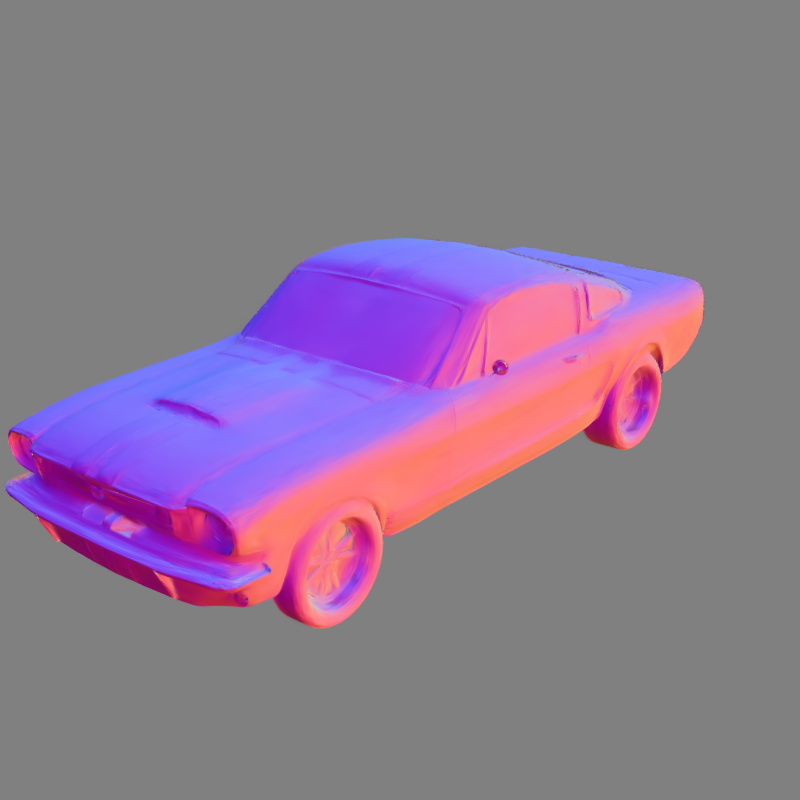} &
    \includegraphics[width=0.16\linewidth]{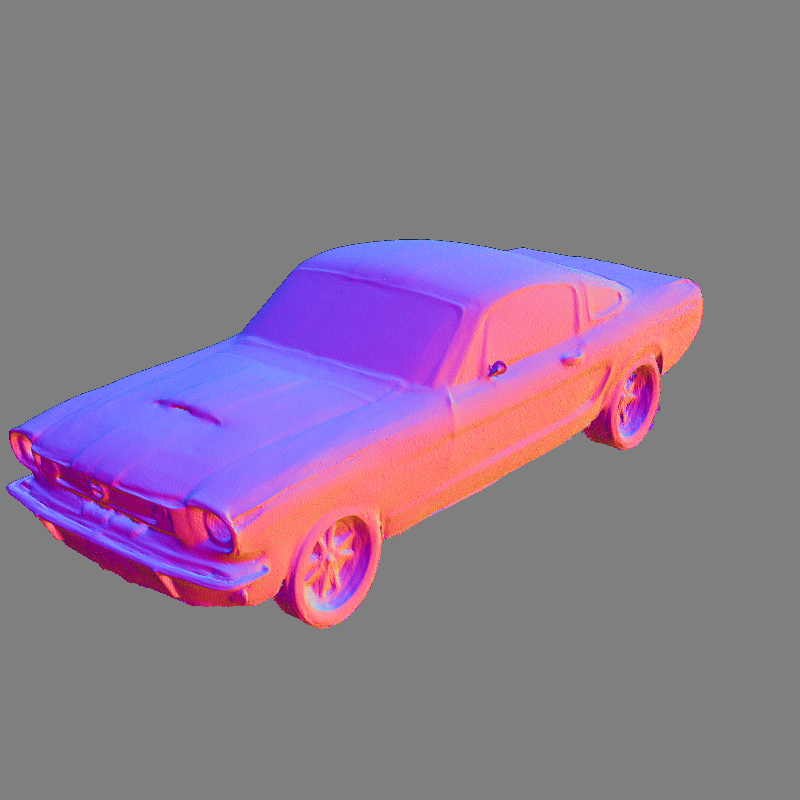} &
    \includegraphics[width=0.16\linewidth]{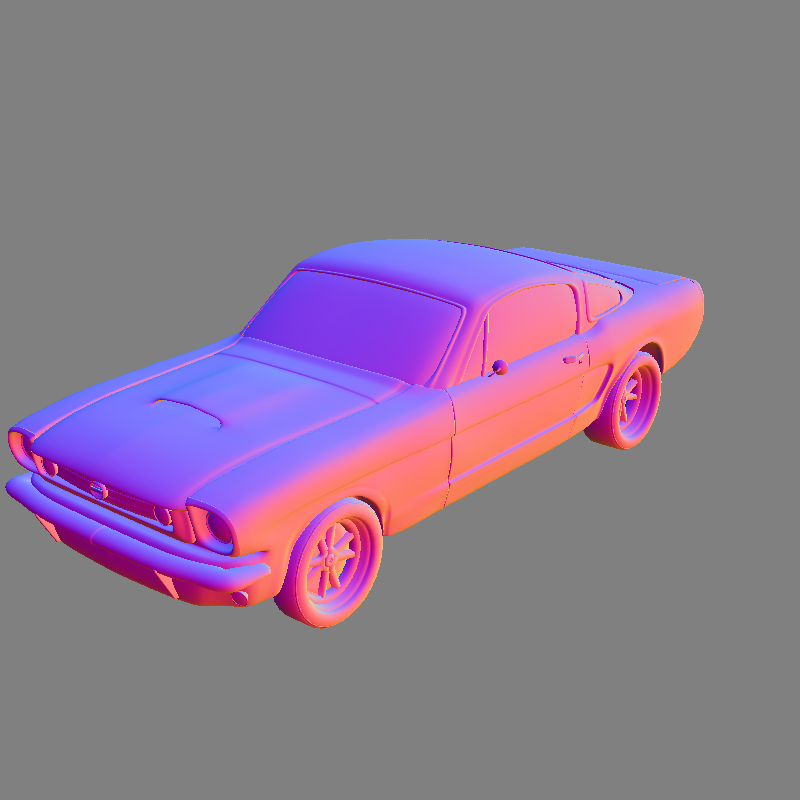} \\

    \includegraphics[width=0.16\linewidth]{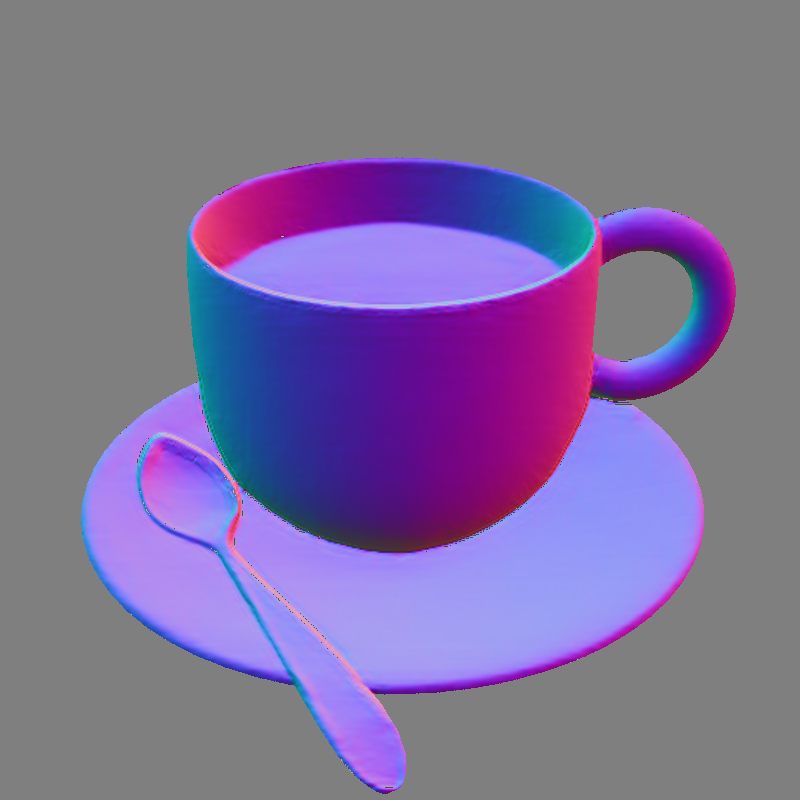} &
    \includegraphics[width=0.16\linewidth]{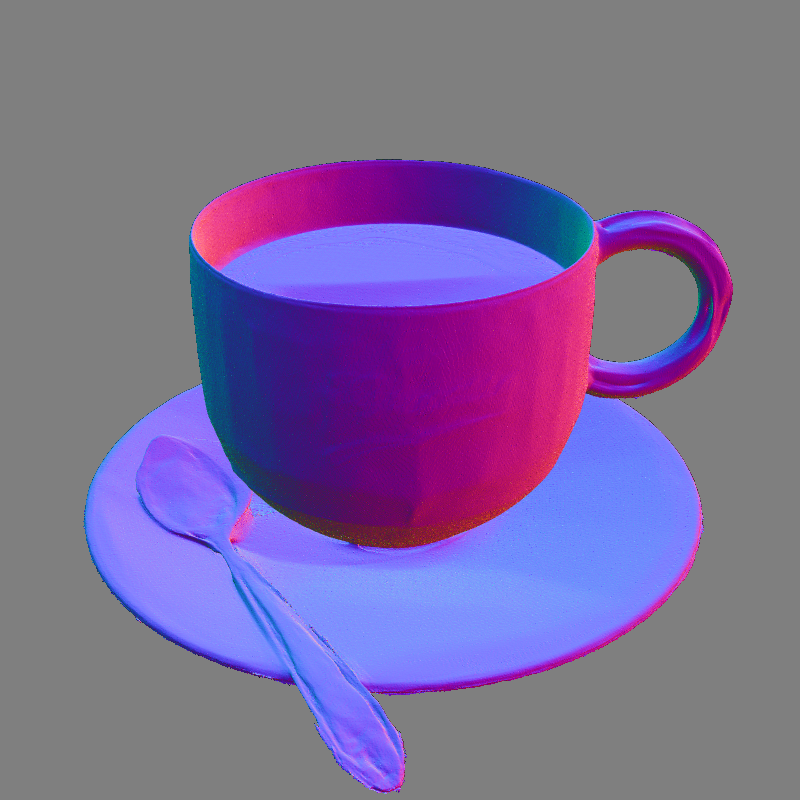} &
    \includegraphics[width=0.16\linewidth]{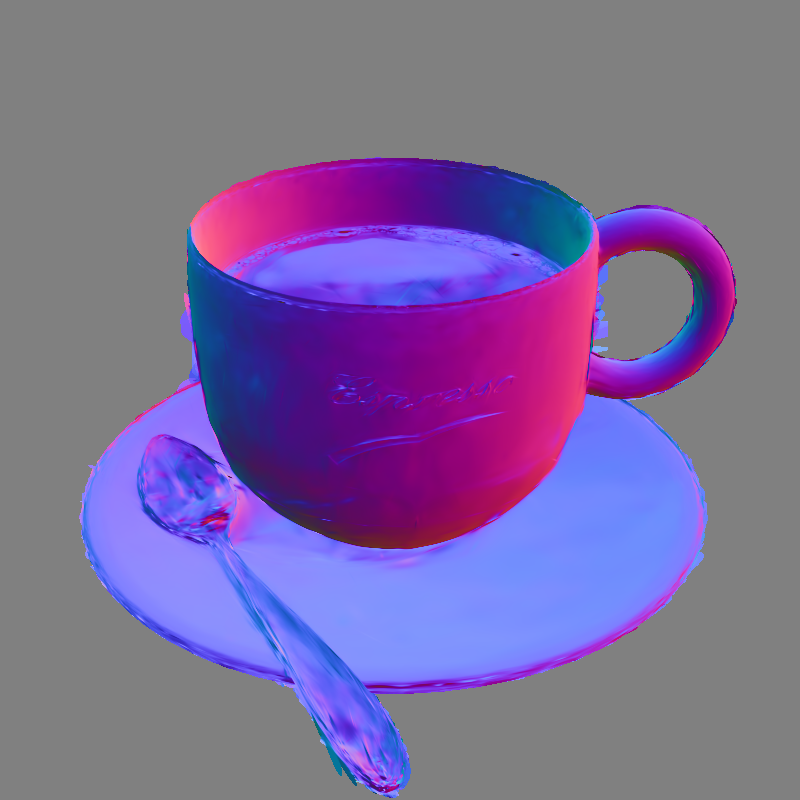} &
    \includegraphics[width=0.16\linewidth]{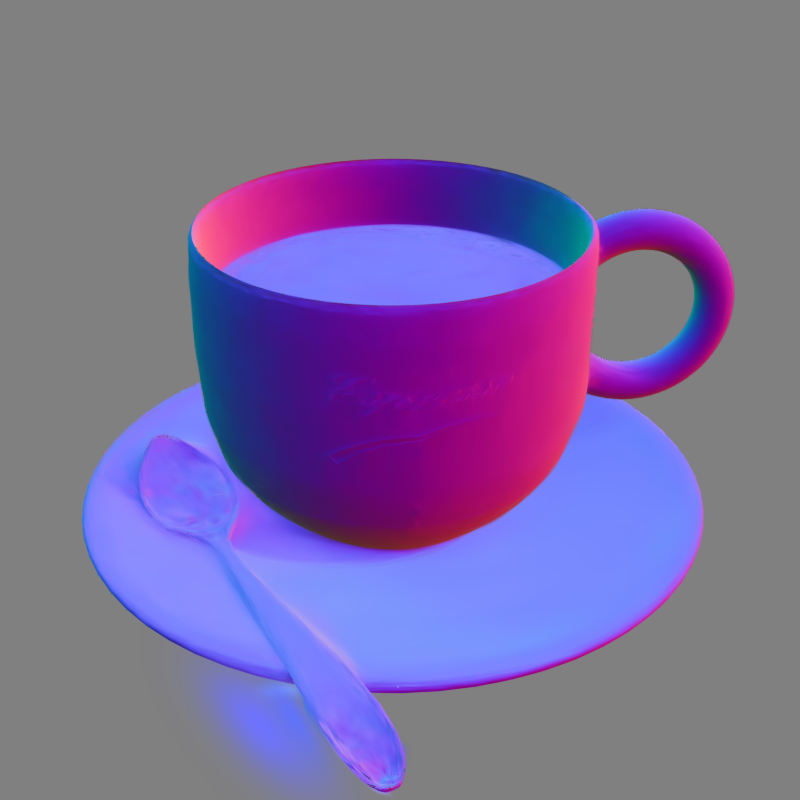} &
    \includegraphics[width=0.16\linewidth]{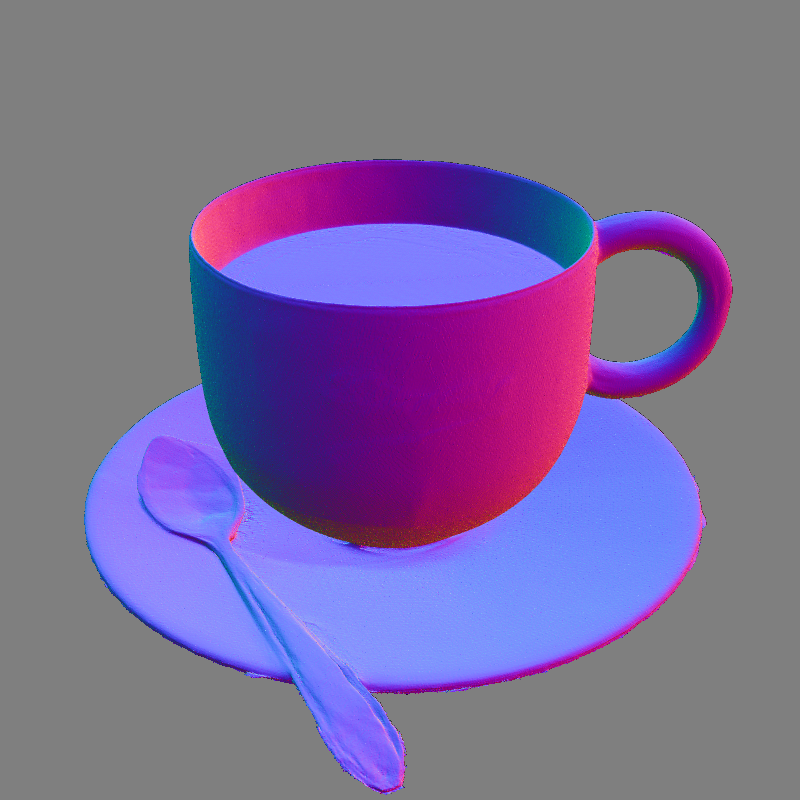} &
    \includegraphics[width=0.16\linewidth]{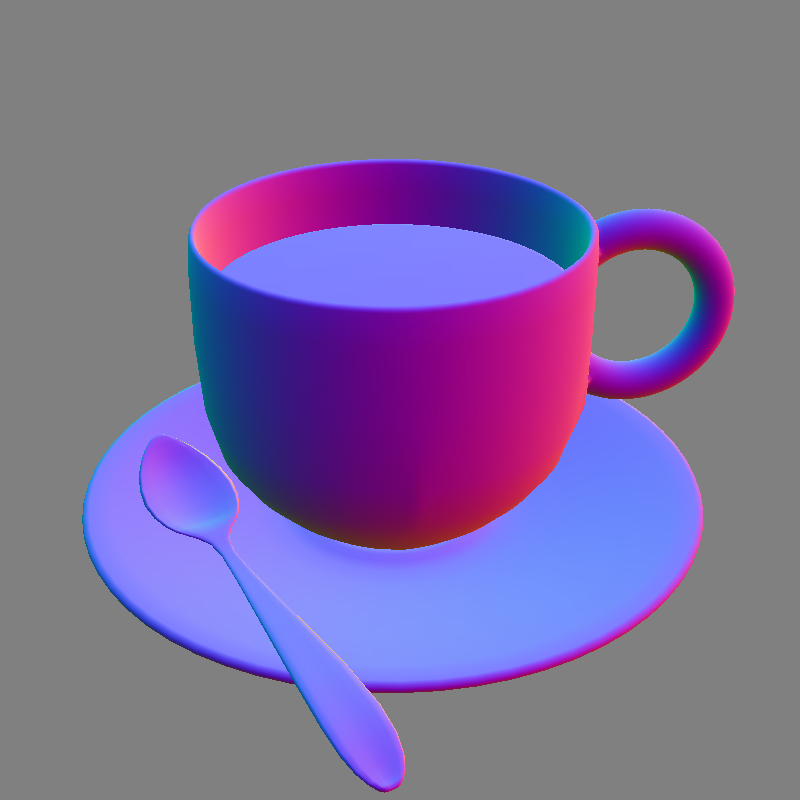} \\

    \includegraphics[width=0.16\linewidth]{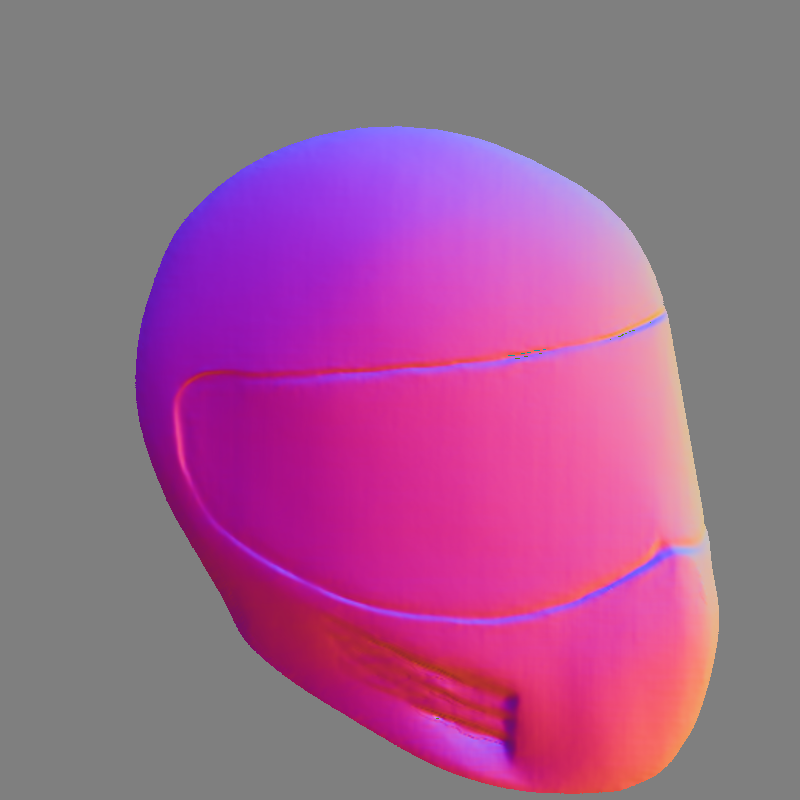} &
    \includegraphics[width=0.16\linewidth]{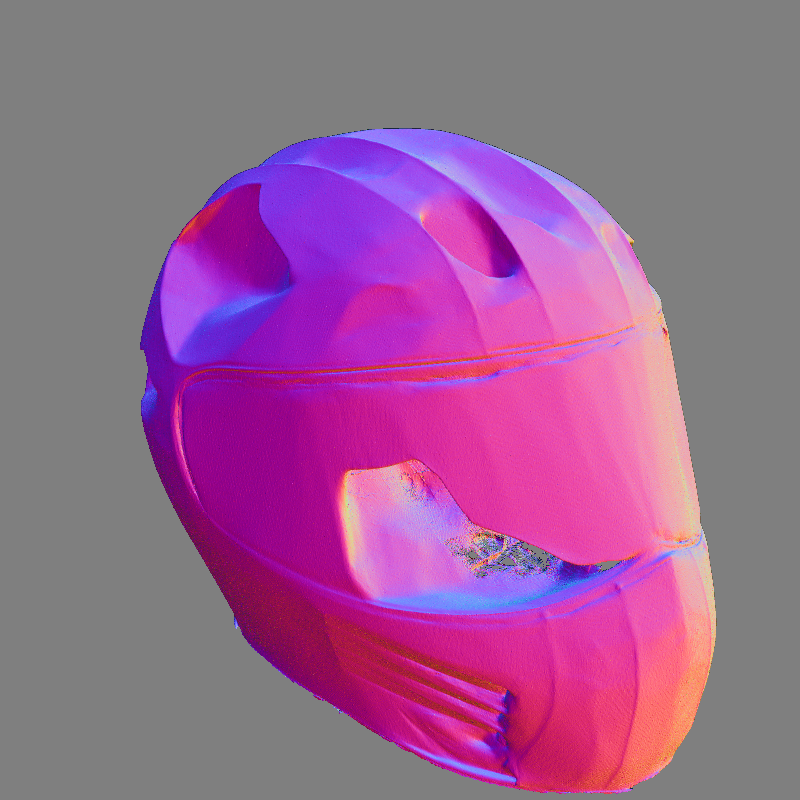} &
    \includegraphics[width=0.16\linewidth]{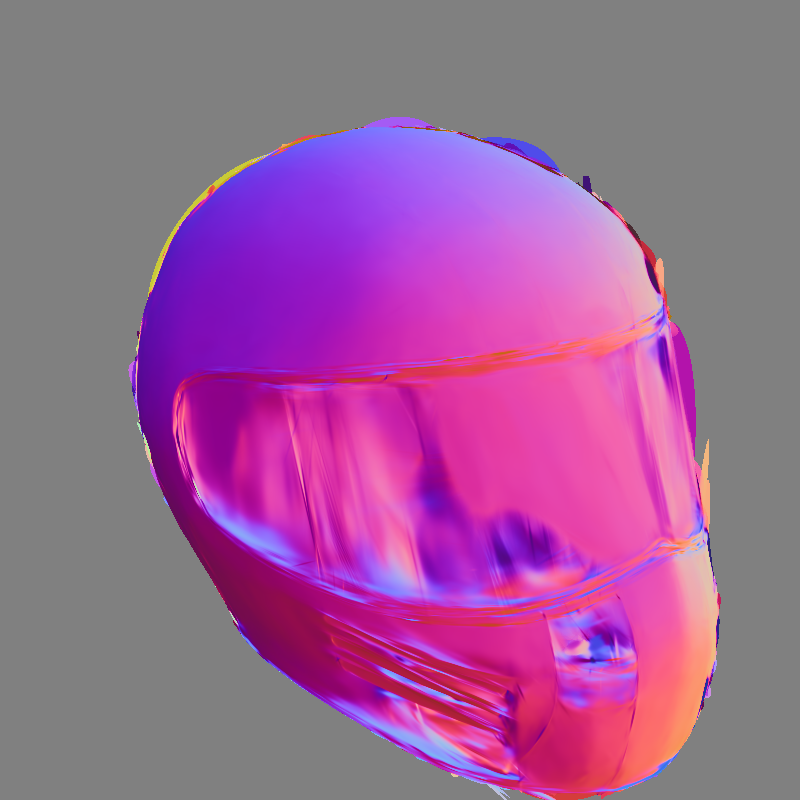} &
    \includegraphics[width=0.16\linewidth]{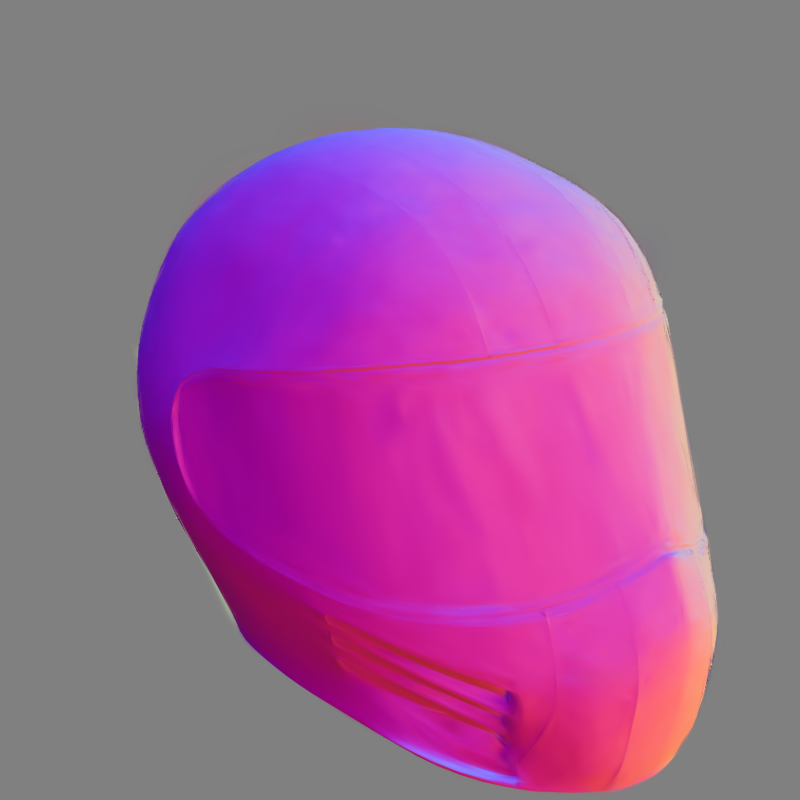} &
    \includegraphics[width=0.16\linewidth]{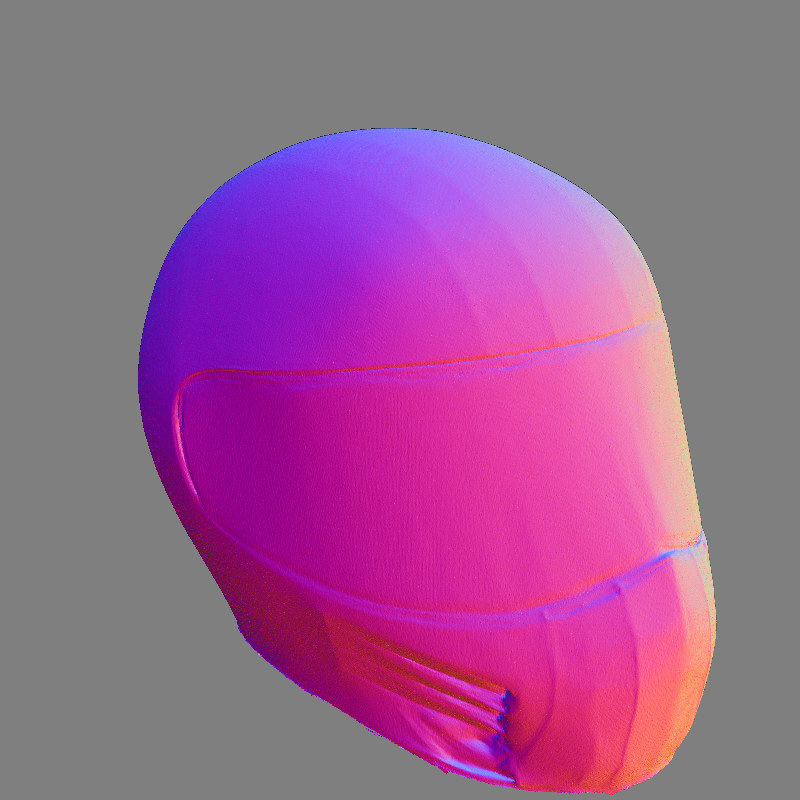} &
    \includegraphics[width=0.16\linewidth]{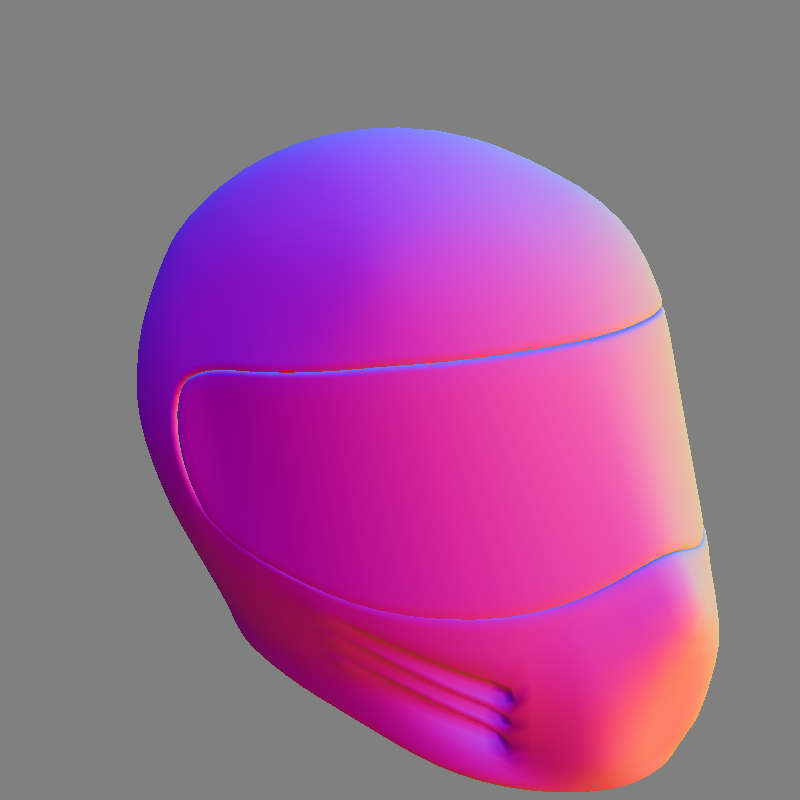} \\

    \includegraphics[width=0.16\linewidth]{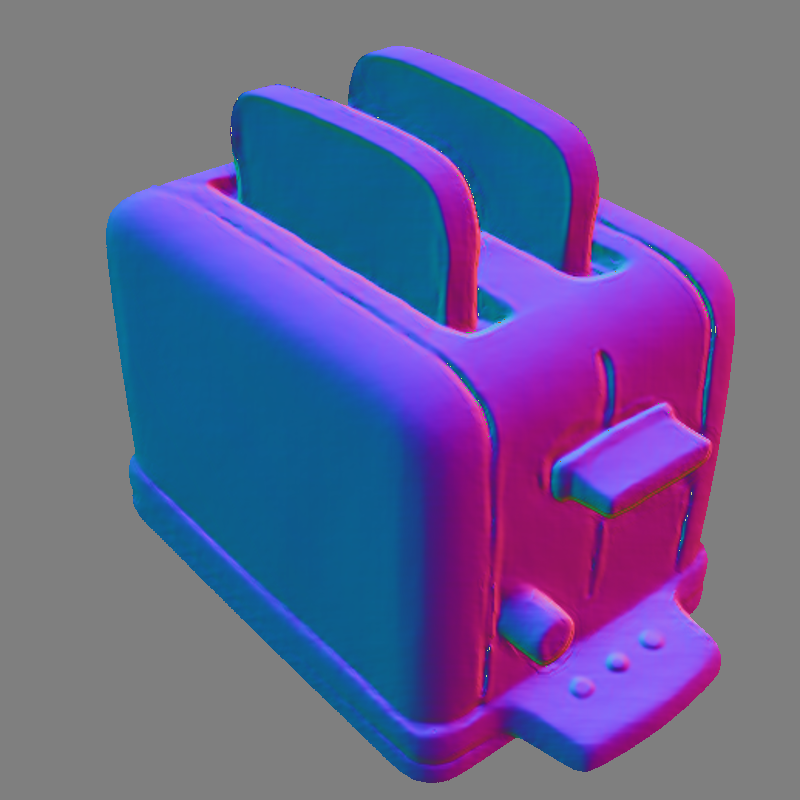} &
    \includegraphics[width=0.16\linewidth]{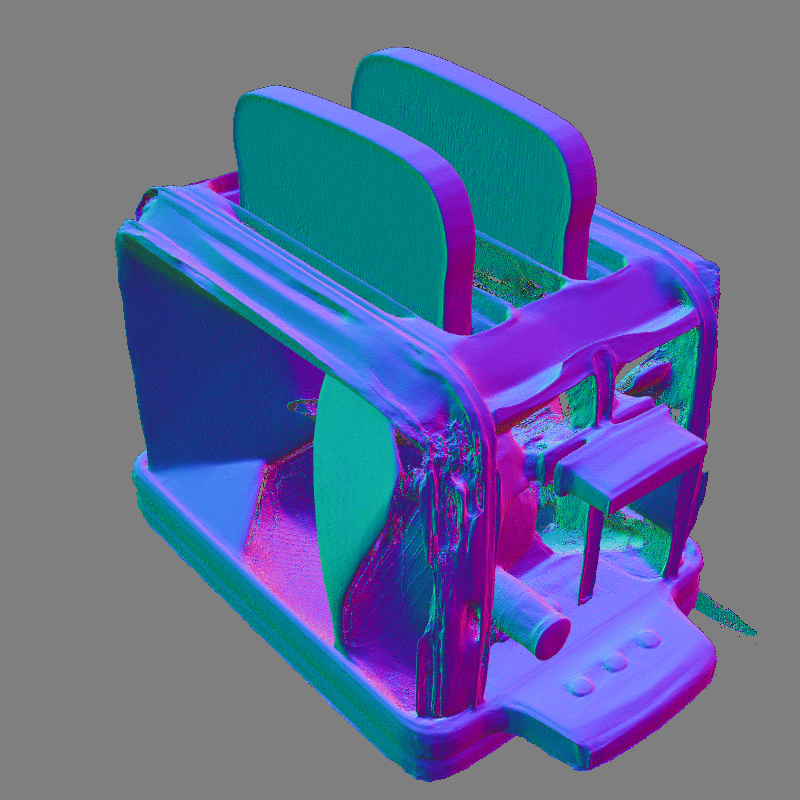} &
    \includegraphics[width=0.16\linewidth]{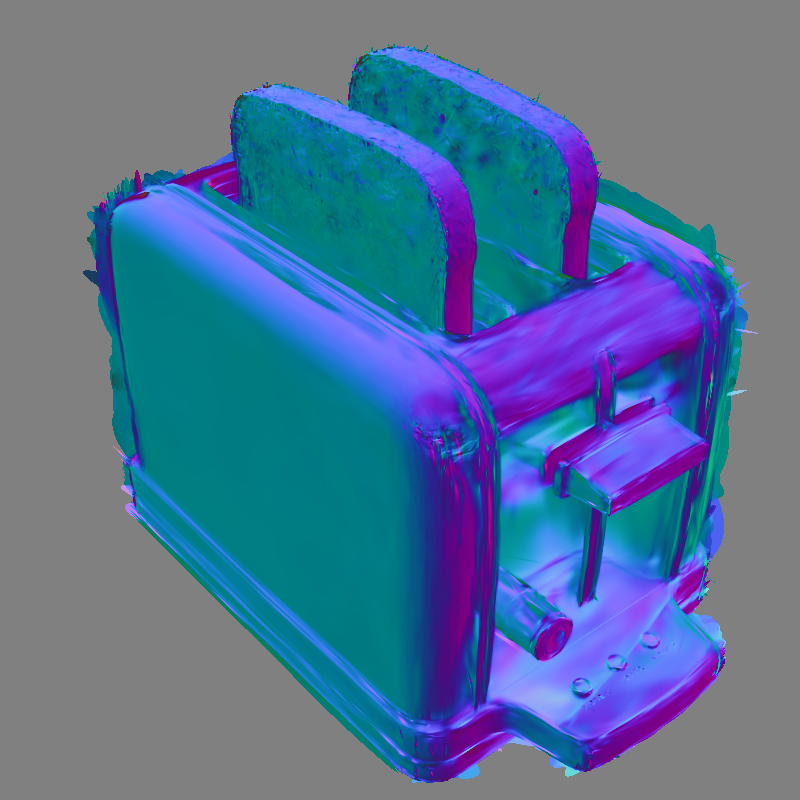} &
    \includegraphics[width=0.16\linewidth]{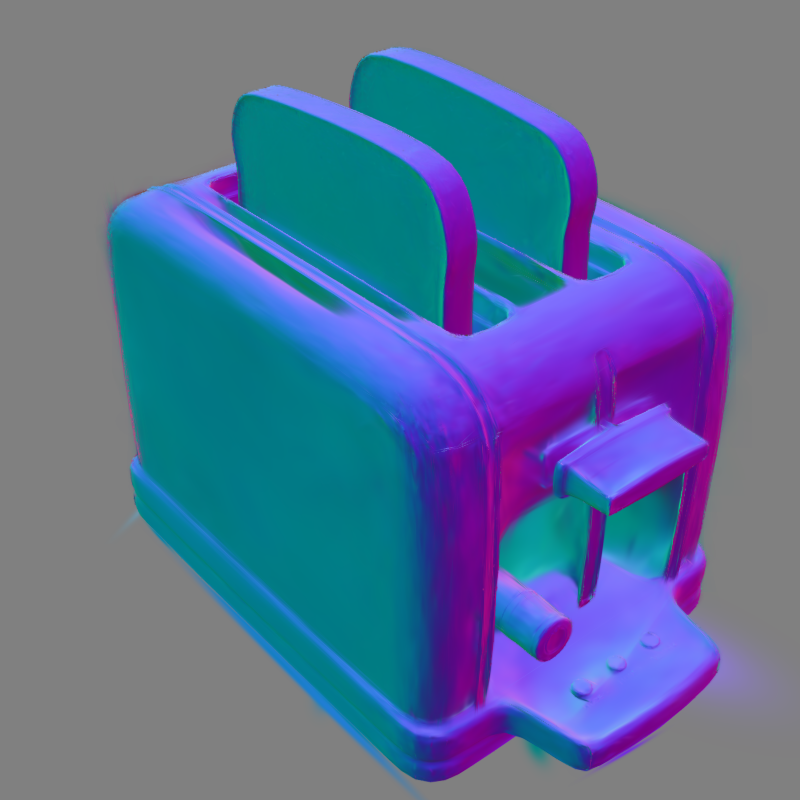} &
    \includegraphics[width=0.16\linewidth]{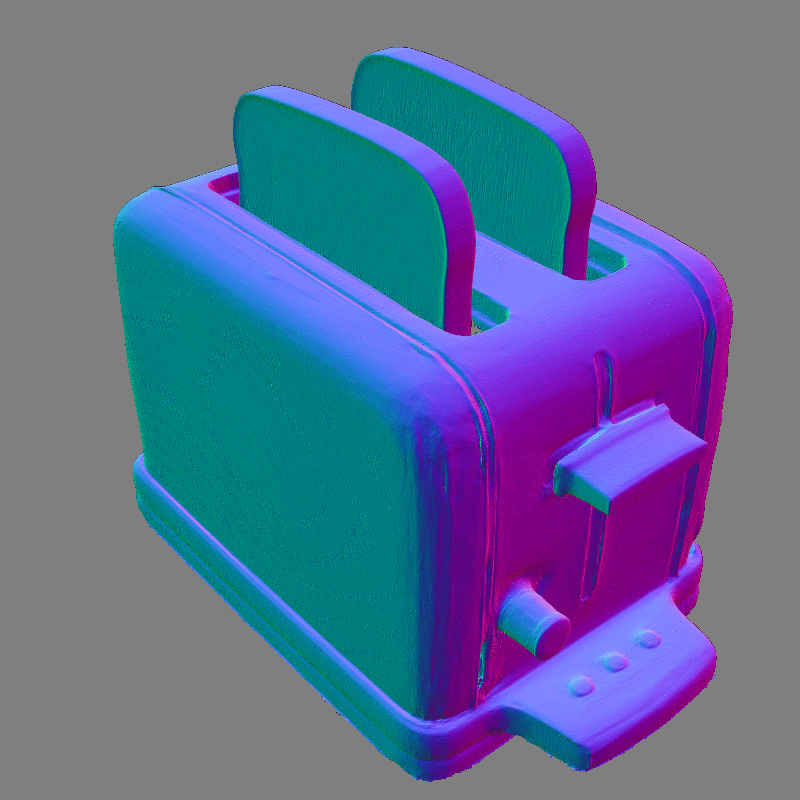} &
    \includegraphics[width=0.16\linewidth]{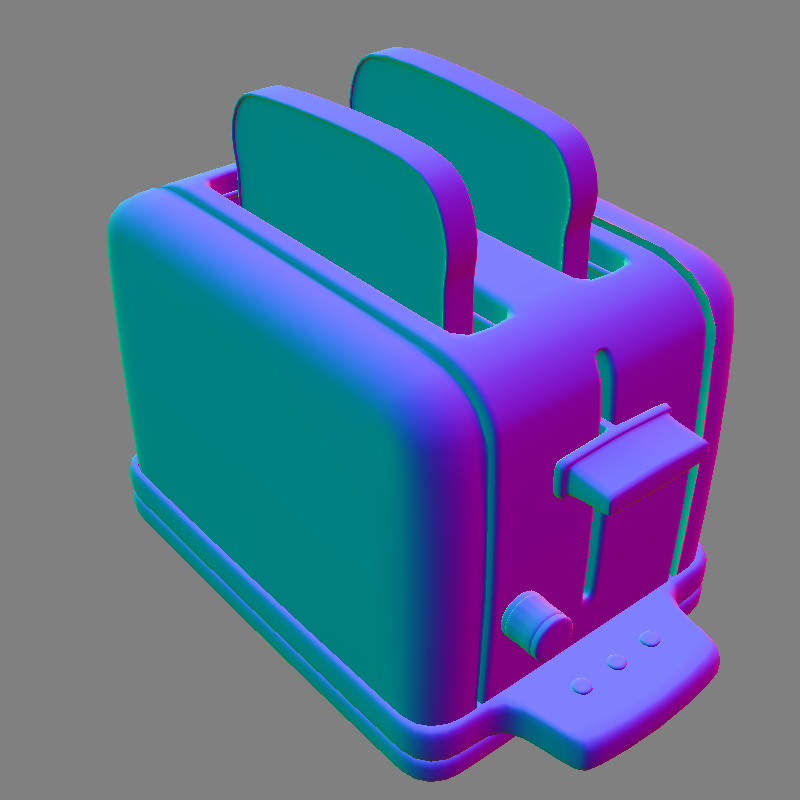} \\
 
    \end{tabular}
    \caption{Qualitative results on the Shiny Blender dataset. VGGT has erroneous prediction at side surface of the car and the toaster and PGSR has many holes for almost every shiny objects. When combining VGGT and PGSR, their artifacts are removed and we can achieve the SOTA performance.}
    \label{fig:qualtitative}
\end{figure*}

\begin{figure*}
\setlength{\tabcolsep}{1pt} 
    \renewcommand{\arraystretch}{1} 
    \centering
    \begin{tabular}{ccccc}
    {\footnotesize \text{PGSR}} & {\footnotesize \text{+SN}} & +DA & {\footnotesize Ours~(+VGGT)} & {\footnotesize GT}\\
    
    \includegraphics[width=0.18\linewidth]{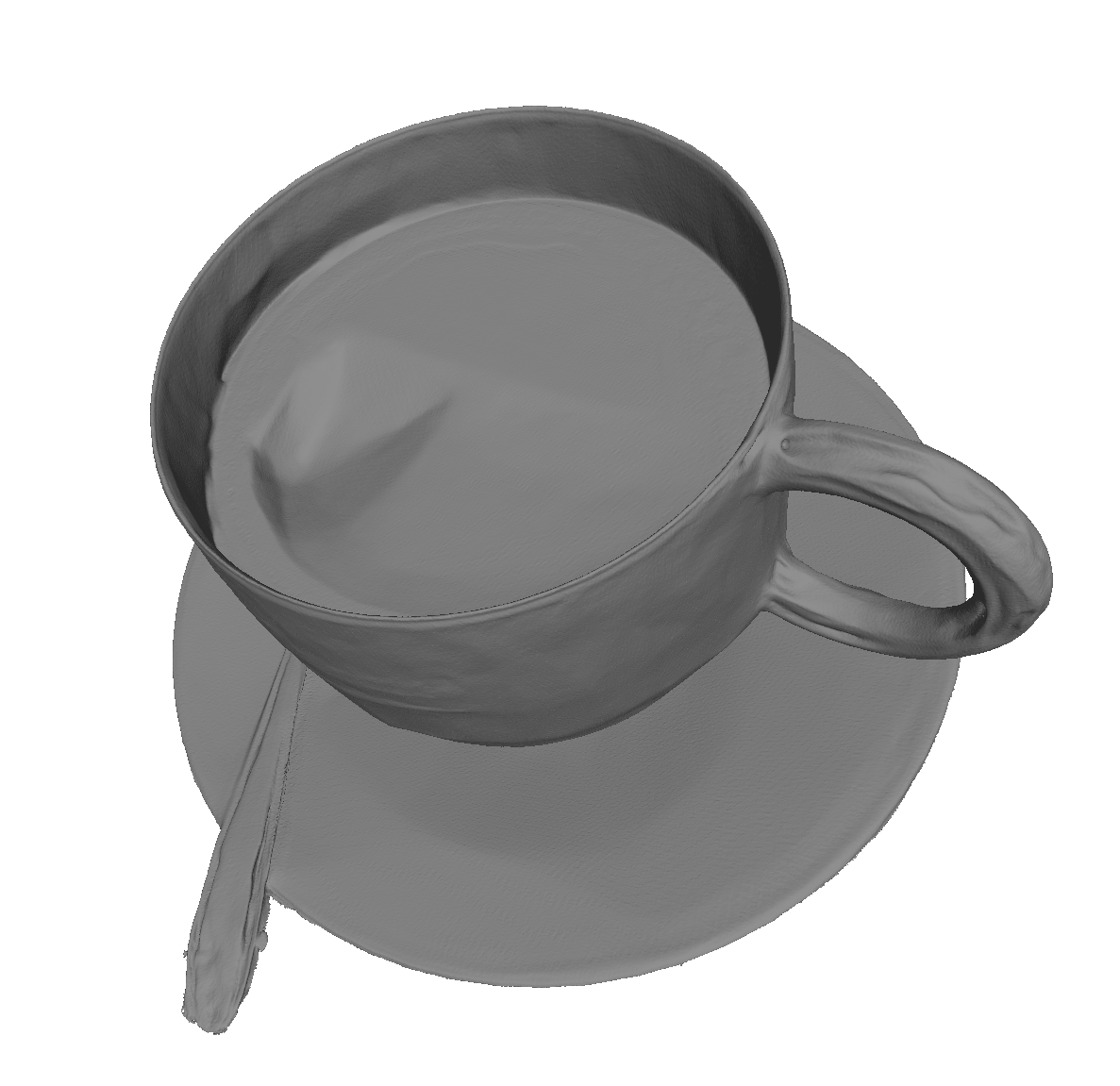} &
    \includegraphics[width=0.18\linewidth]{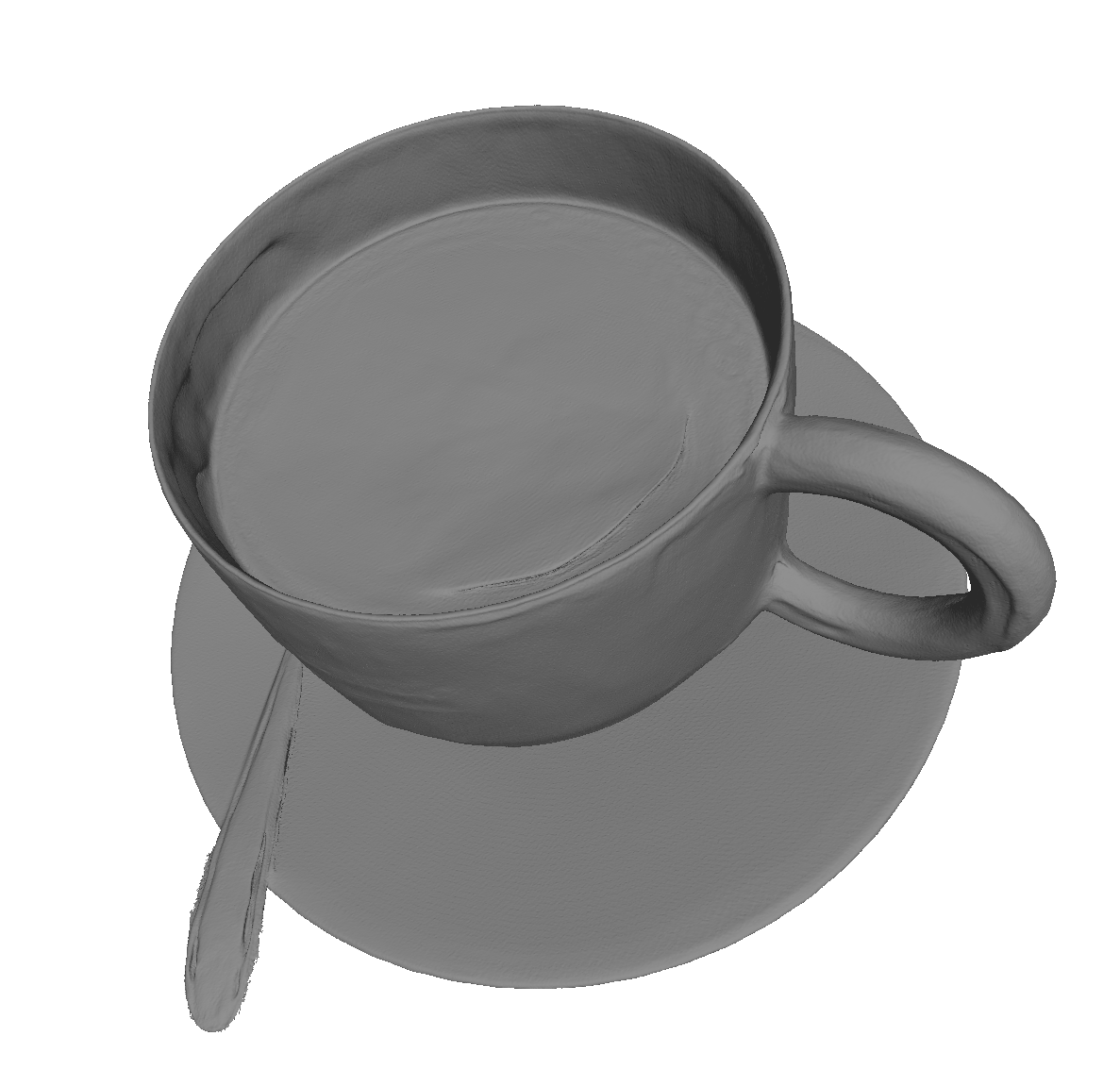} &
    
    \includegraphics[width=0.18\linewidth]{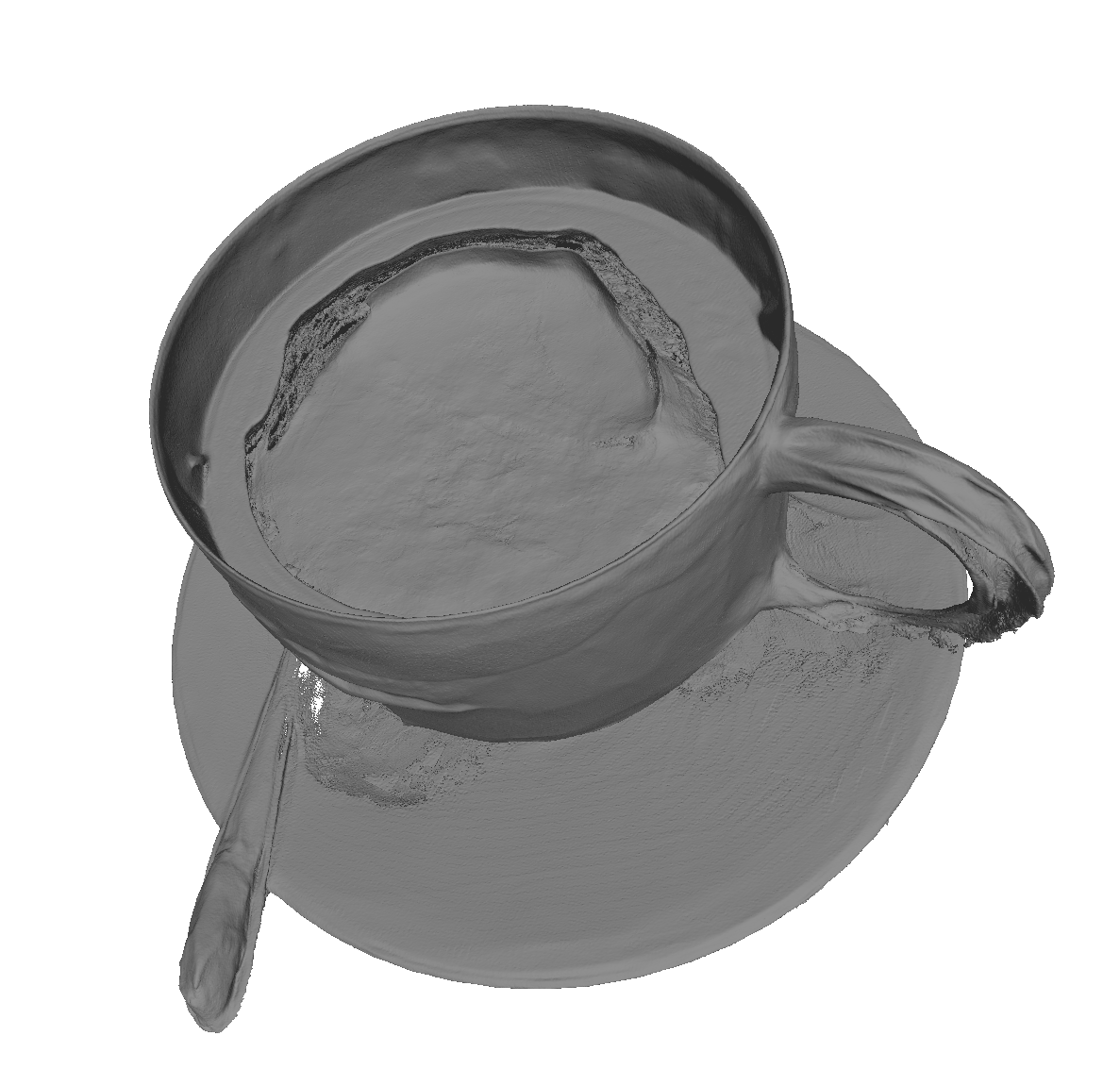} &
    
    \includegraphics[width=0.18\linewidth]{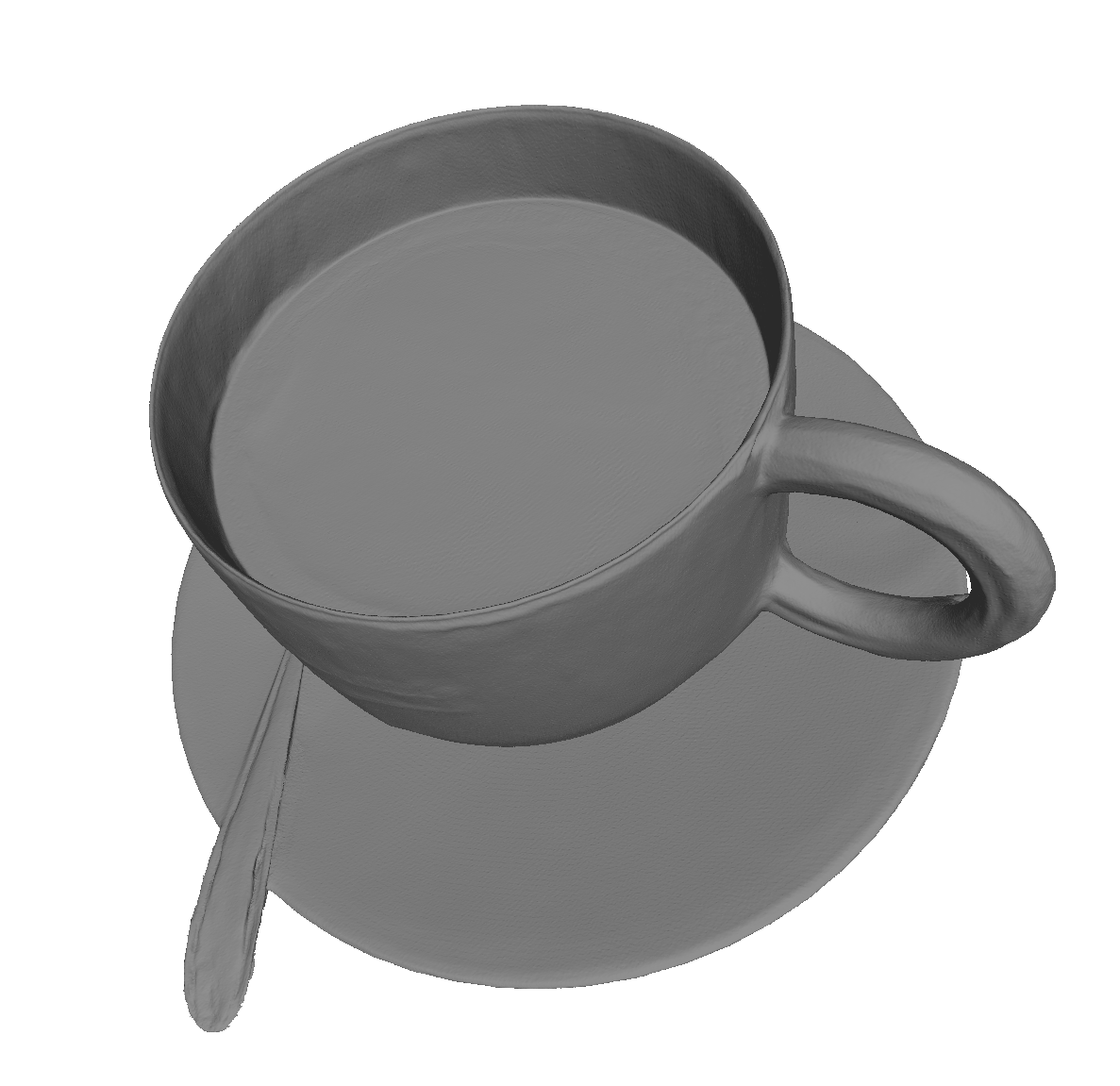} &
    \includegraphics[width=0.18\linewidth]{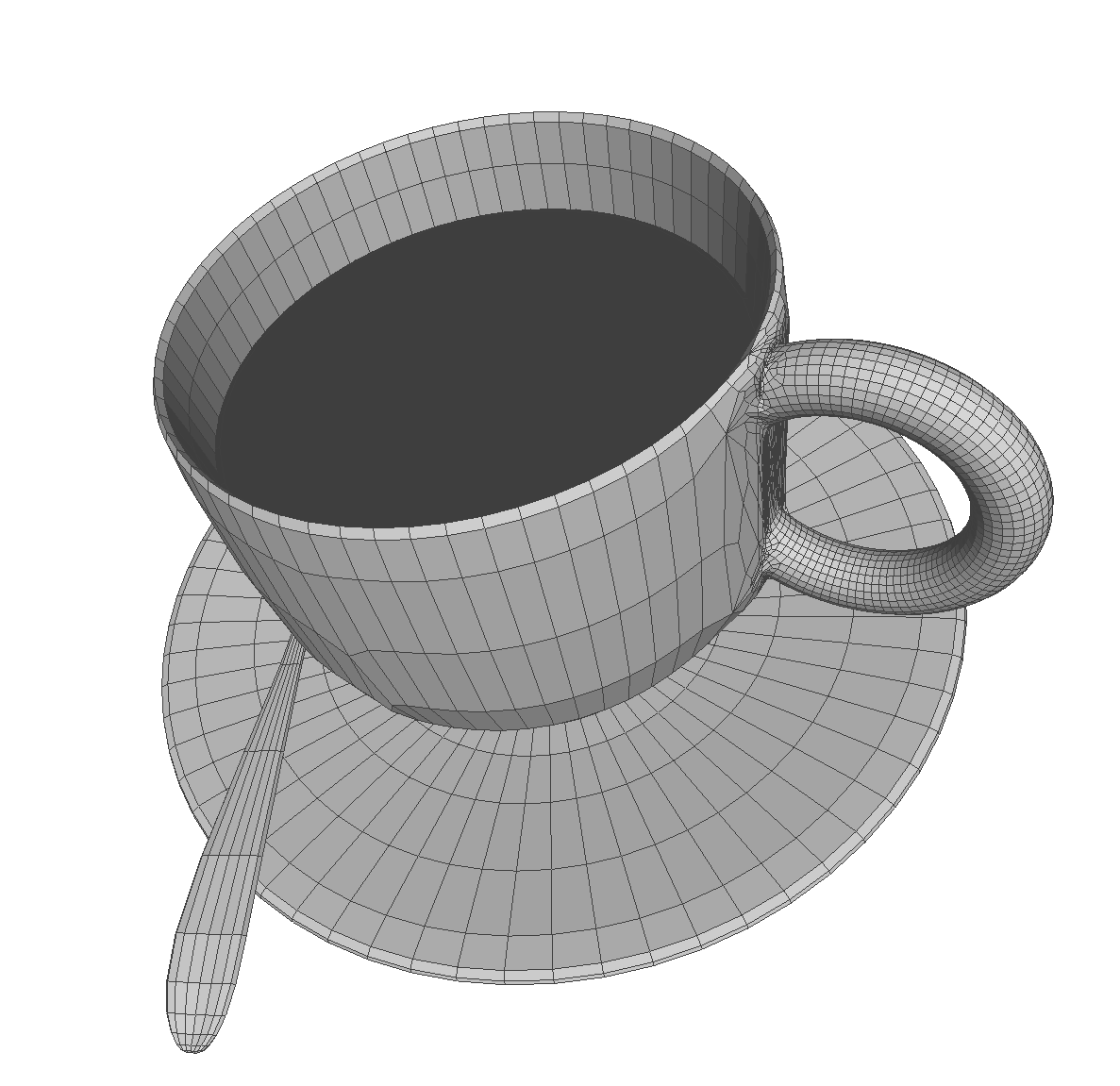} \\
    
    \includegraphics[width=0.18\linewidth]{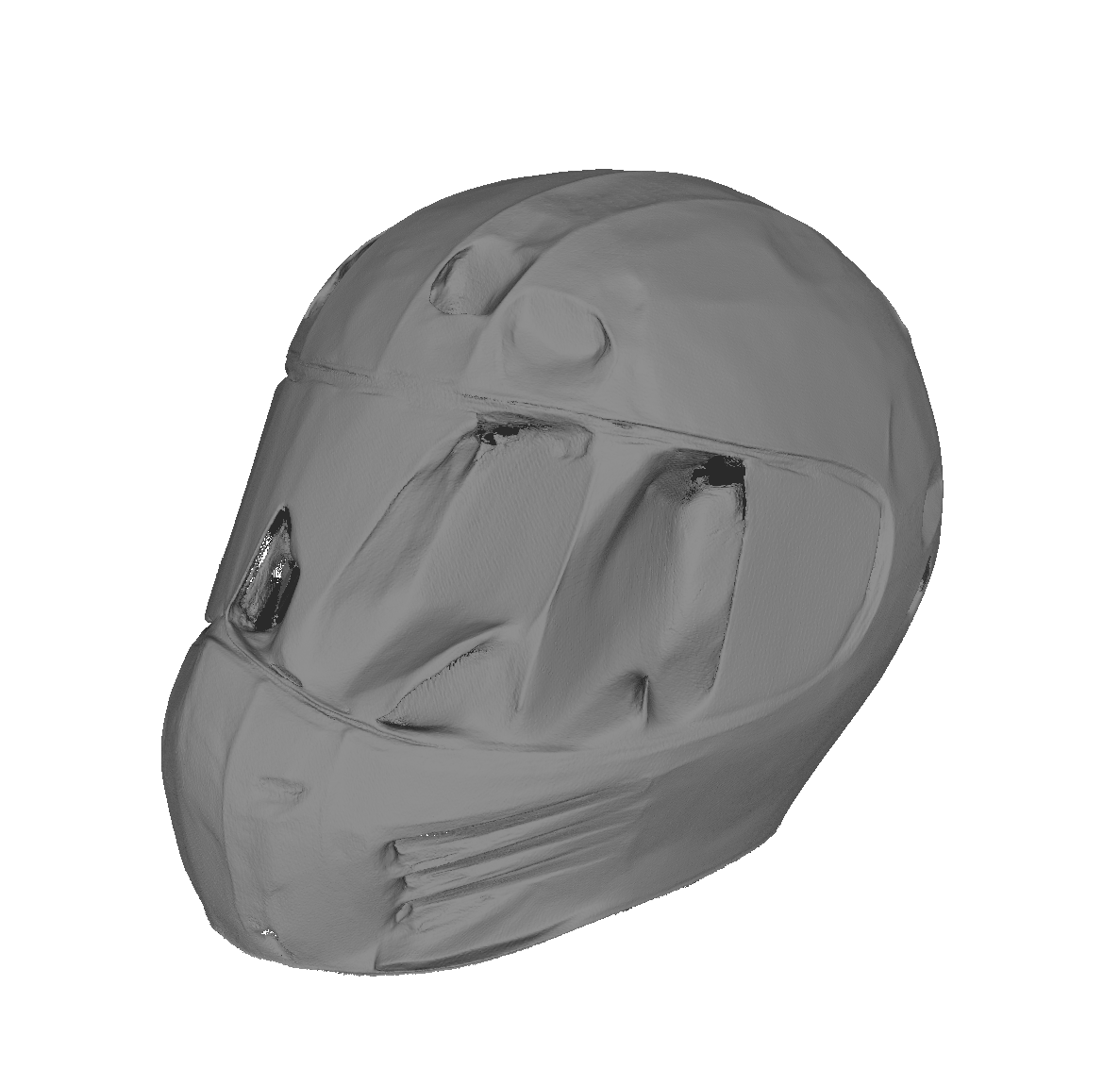} &
    \includegraphics[width=0.18\linewidth]{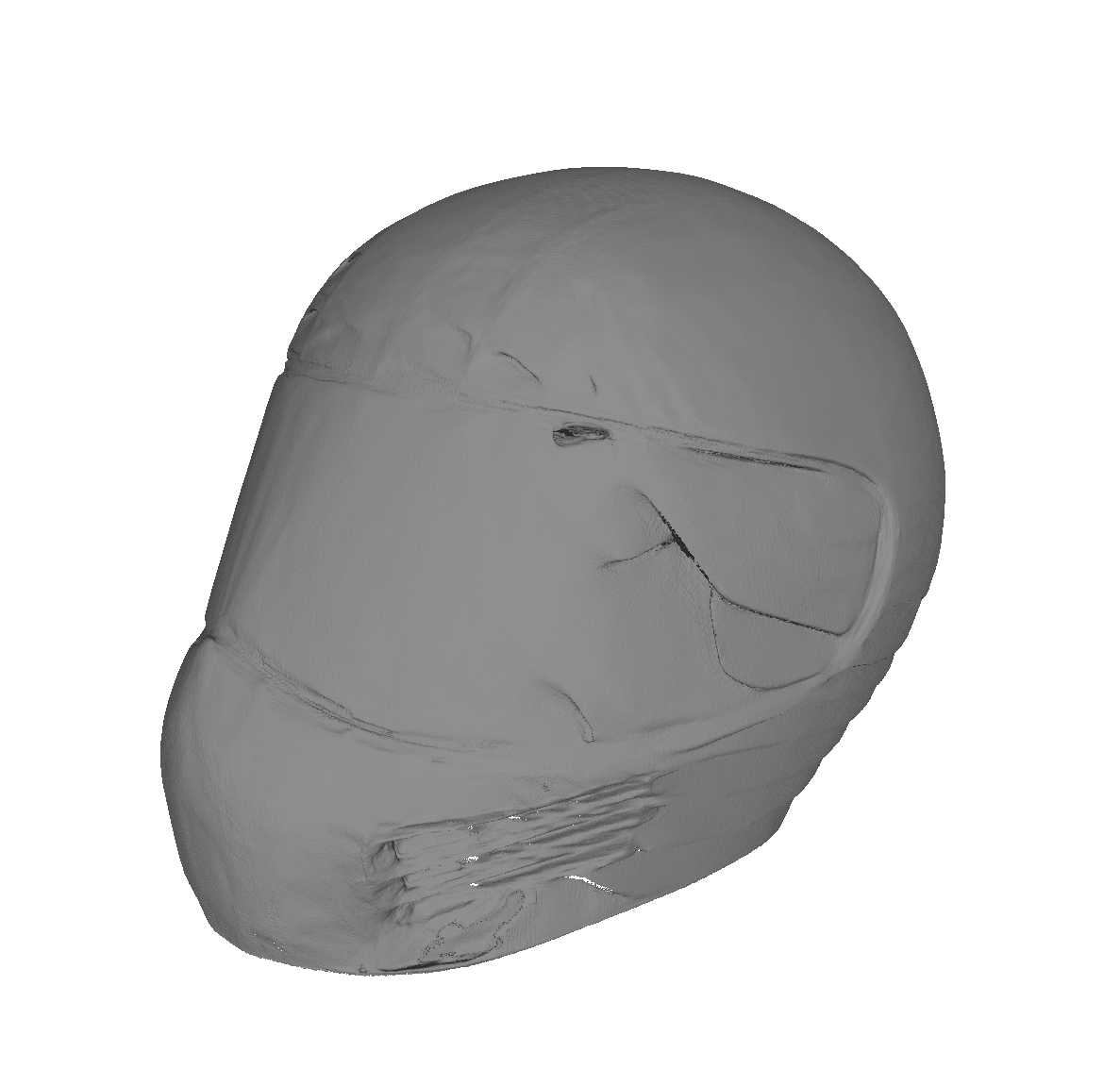} &
    
    \includegraphics[width=0.18\linewidth]{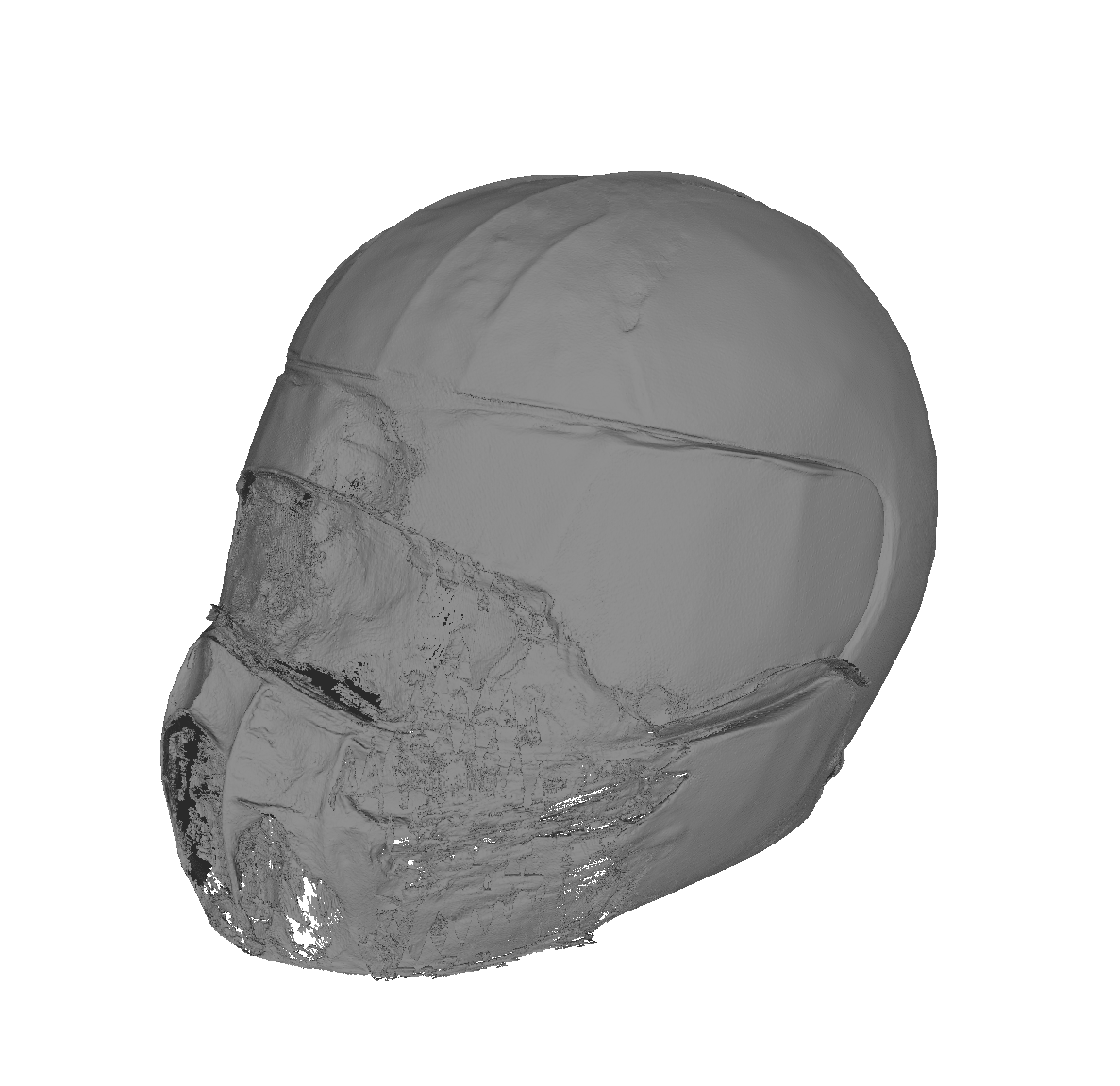} &
    
    \includegraphics[width=0.18\linewidth]{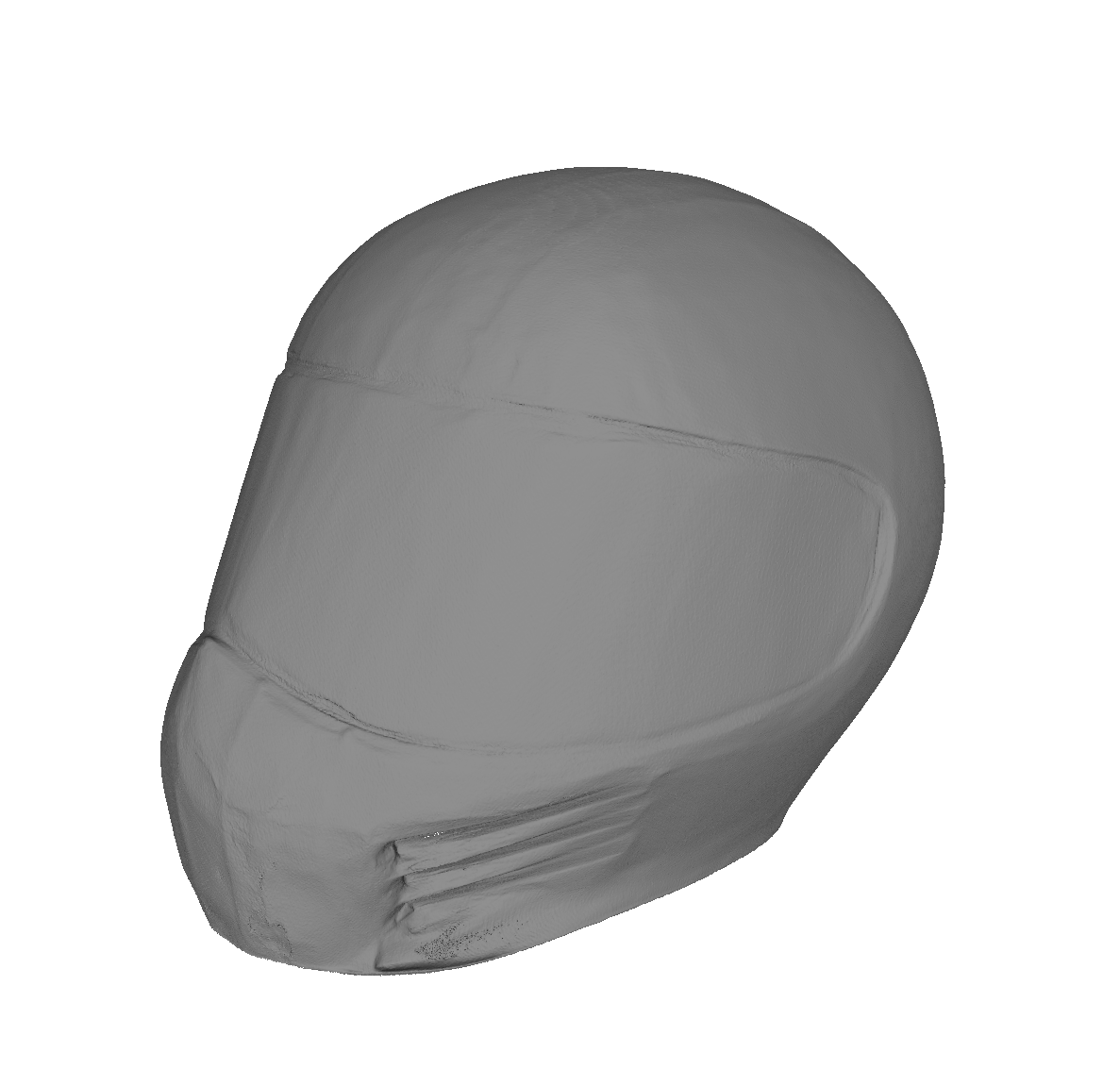} &
    \includegraphics[width=0.18\linewidth]{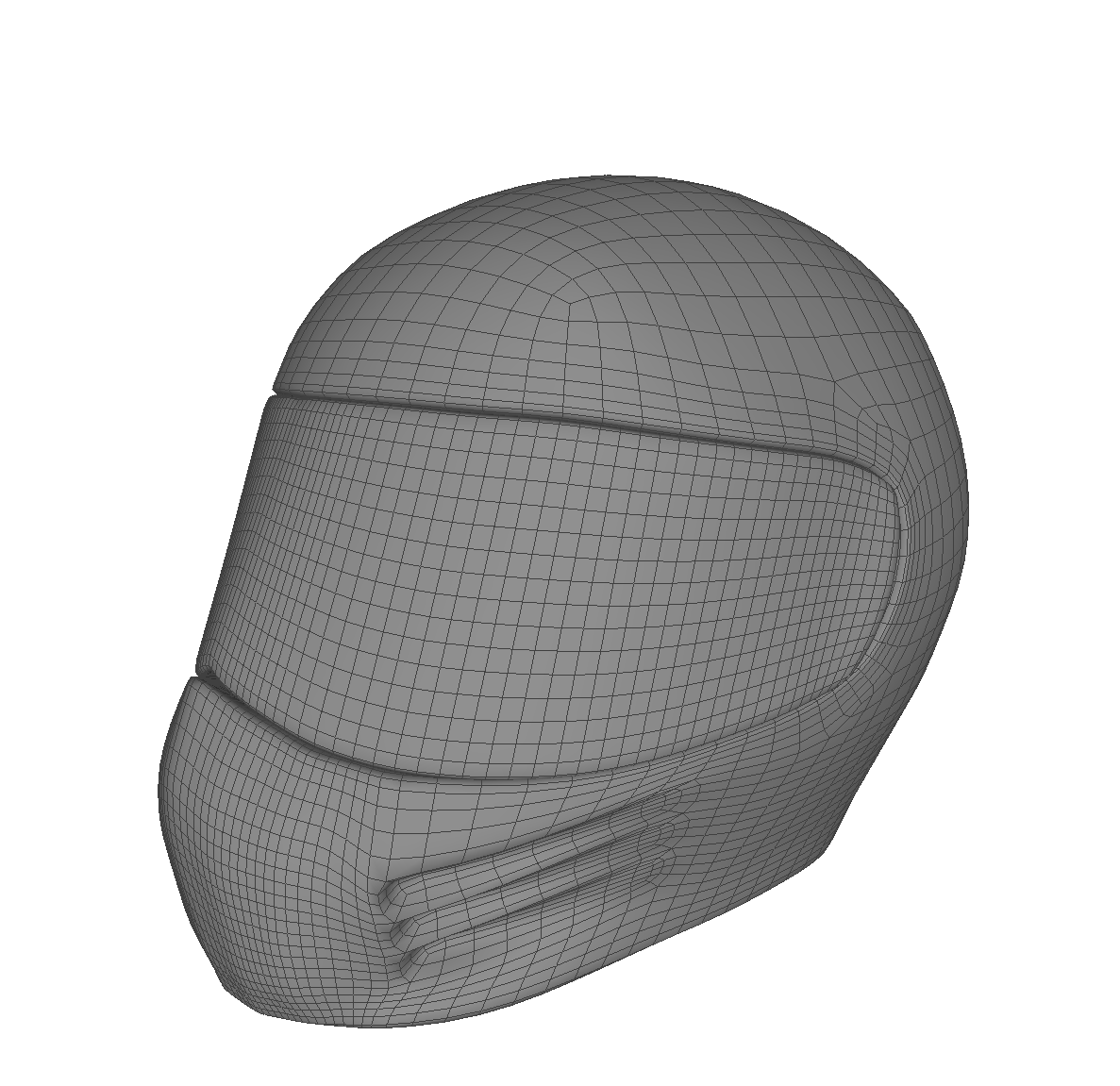} \\

    \includegraphics[width=0.18\linewidth]{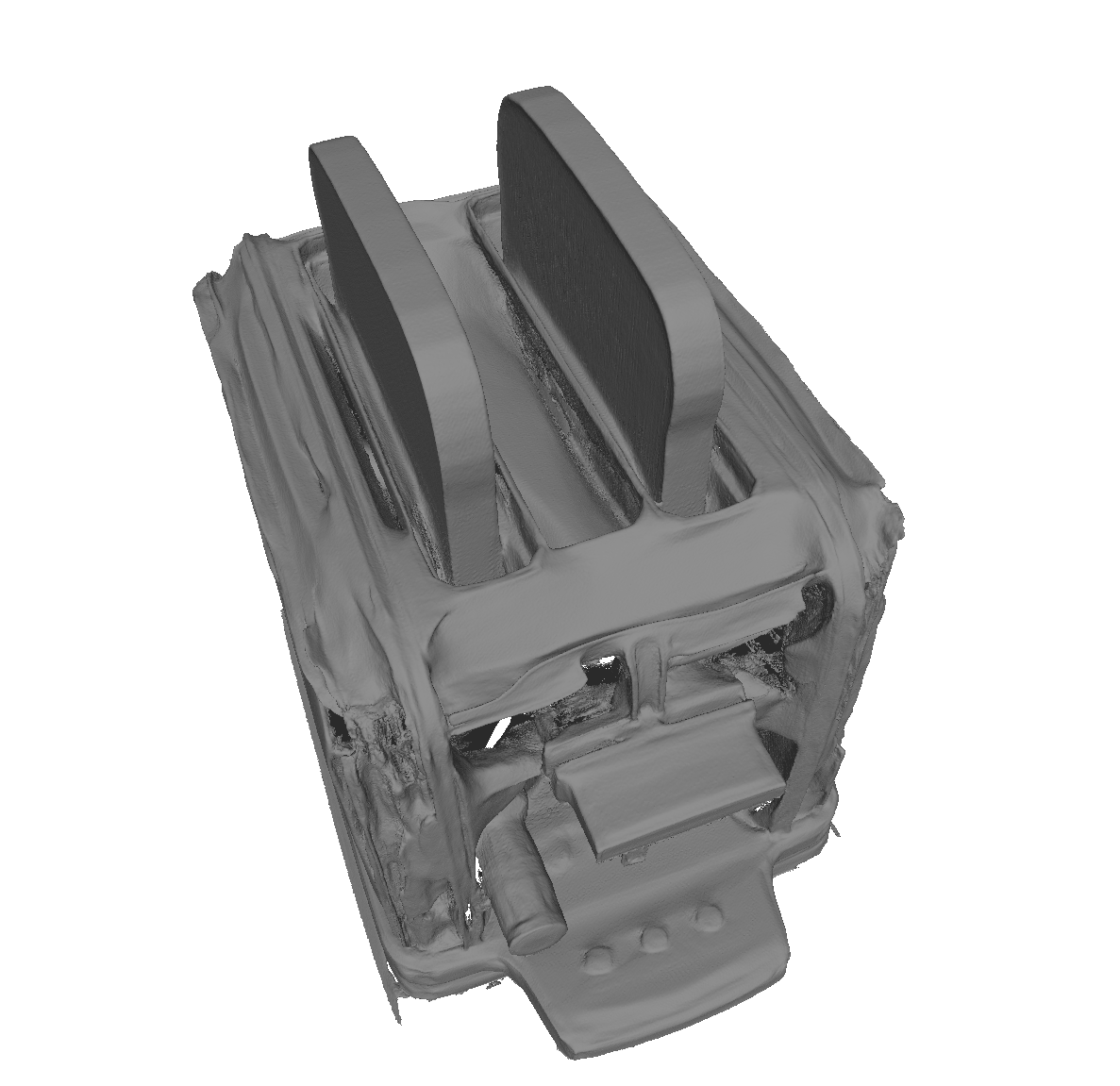} &
    \includegraphics[width=0.18\linewidth]{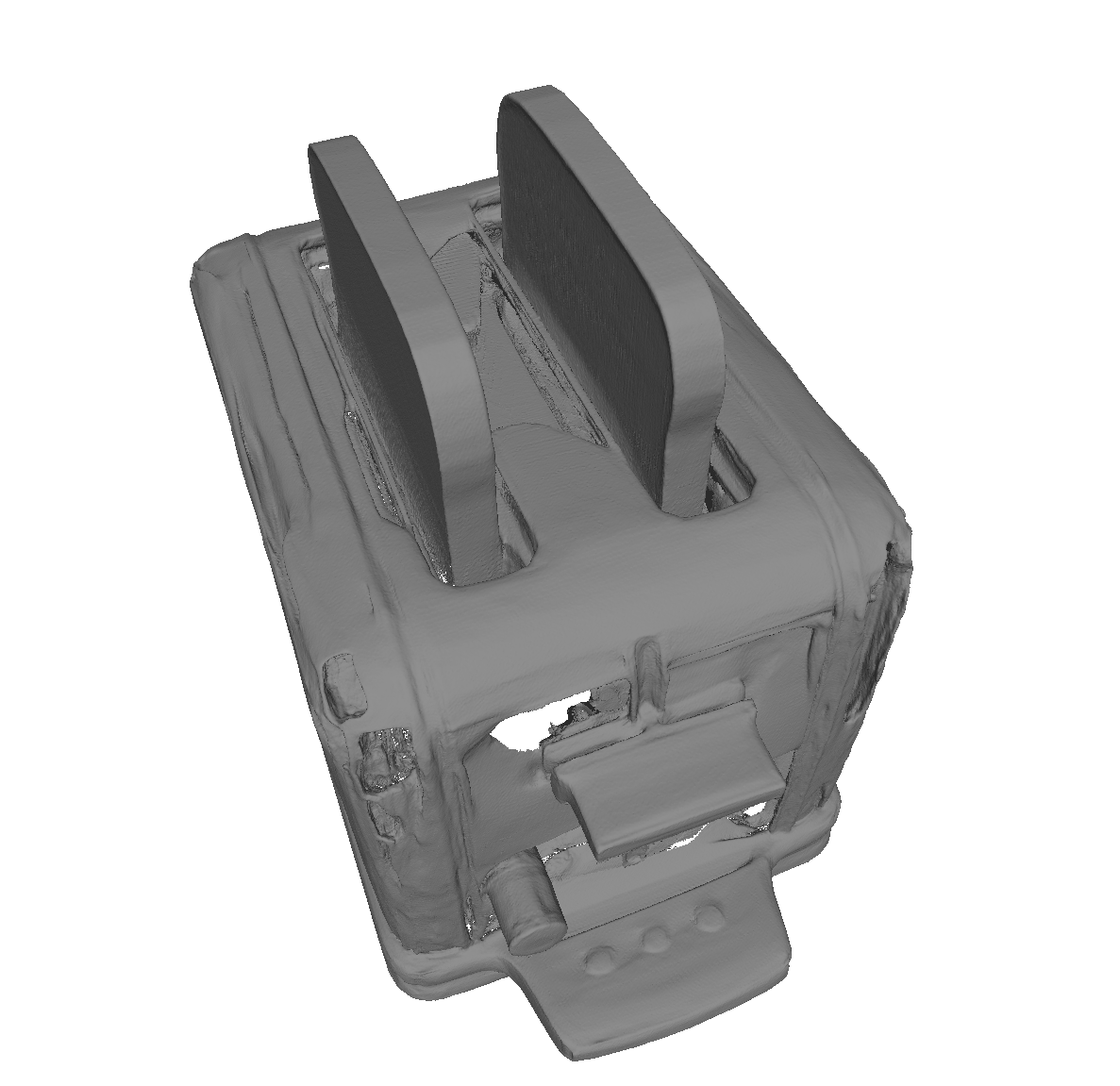} &
    
    \includegraphics[width=0.18\linewidth]{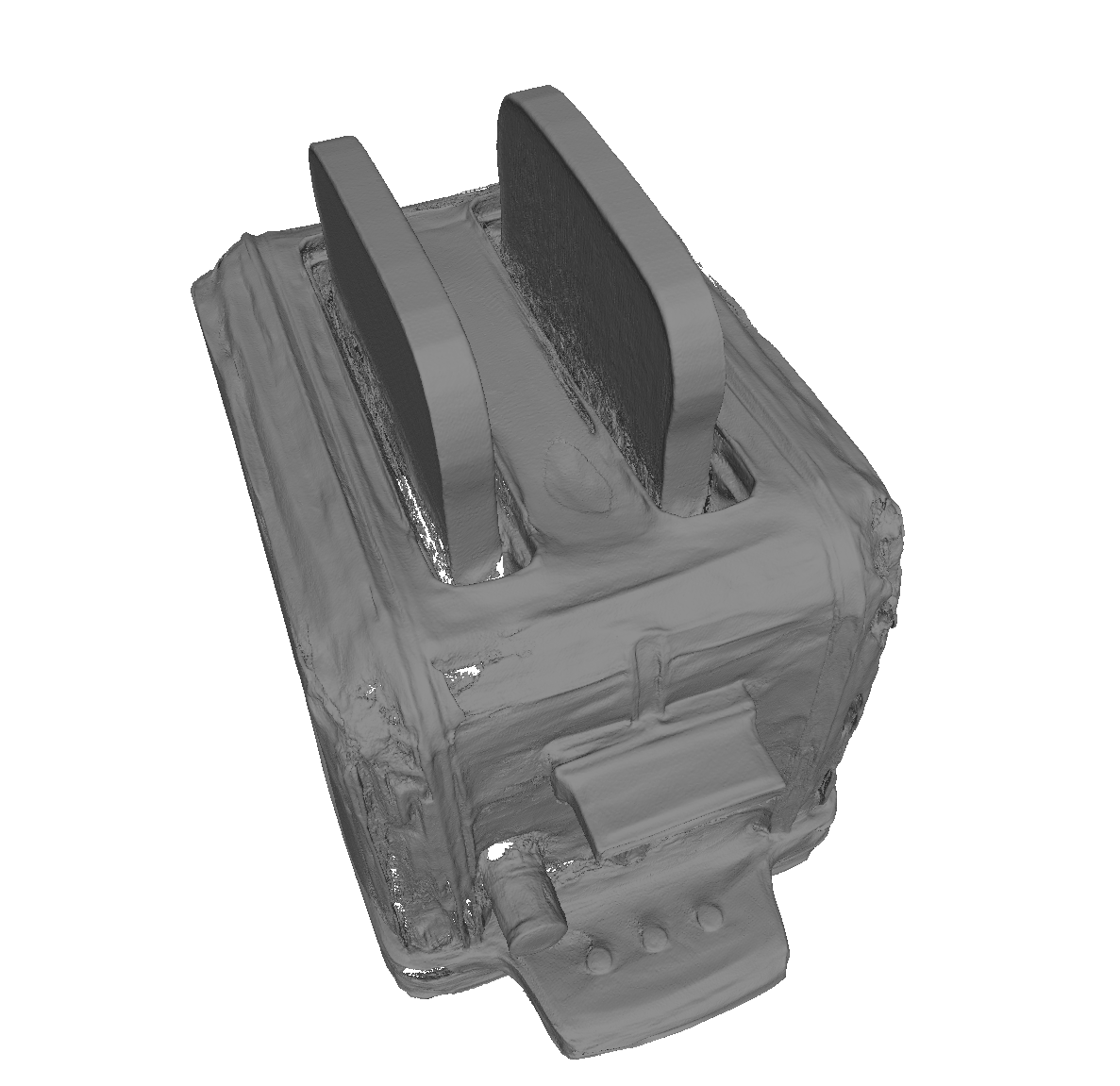} &
    
    \includegraphics[width=0.18\linewidth]{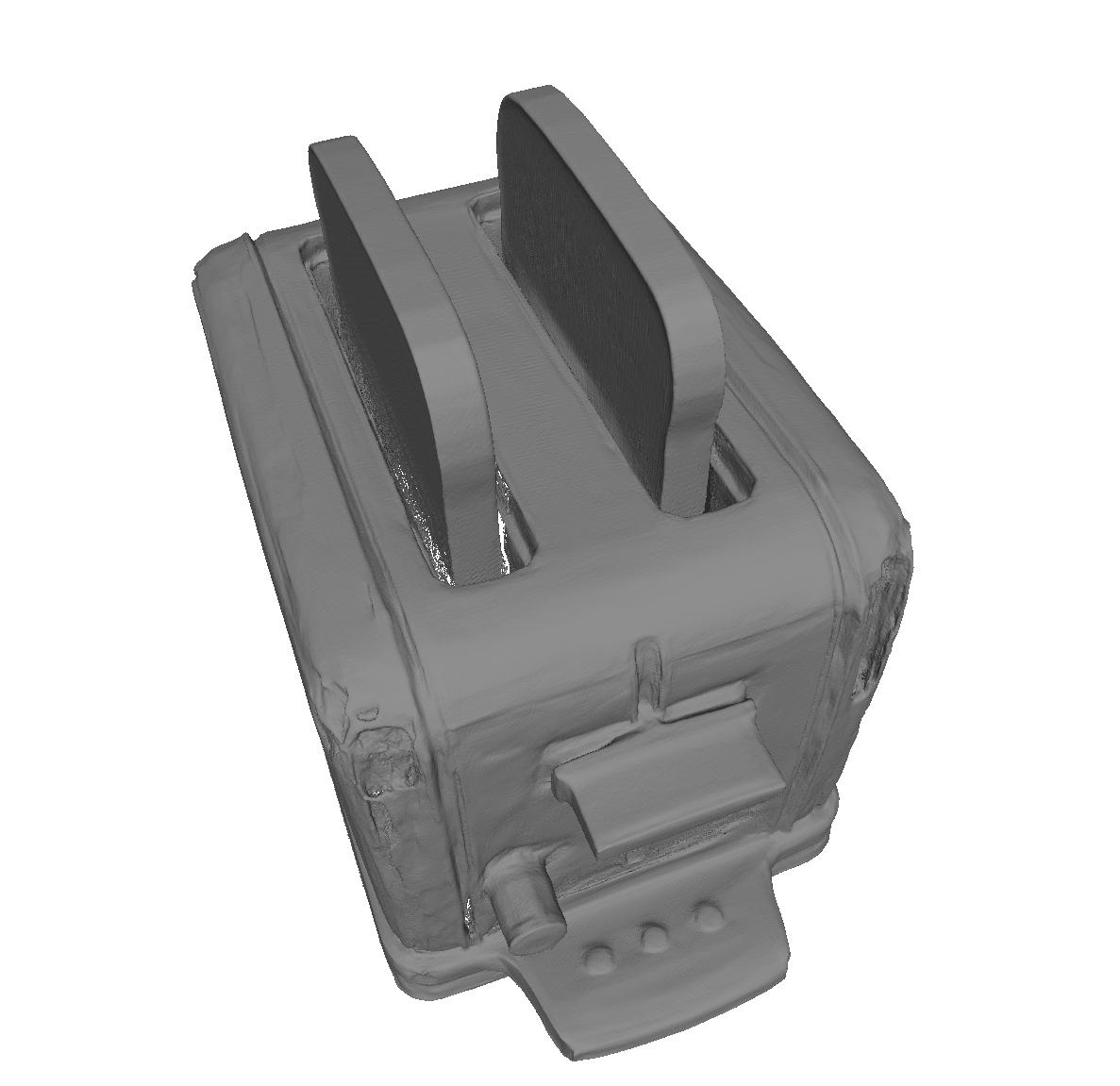} &
    \includegraphics[width=0.18\linewidth]{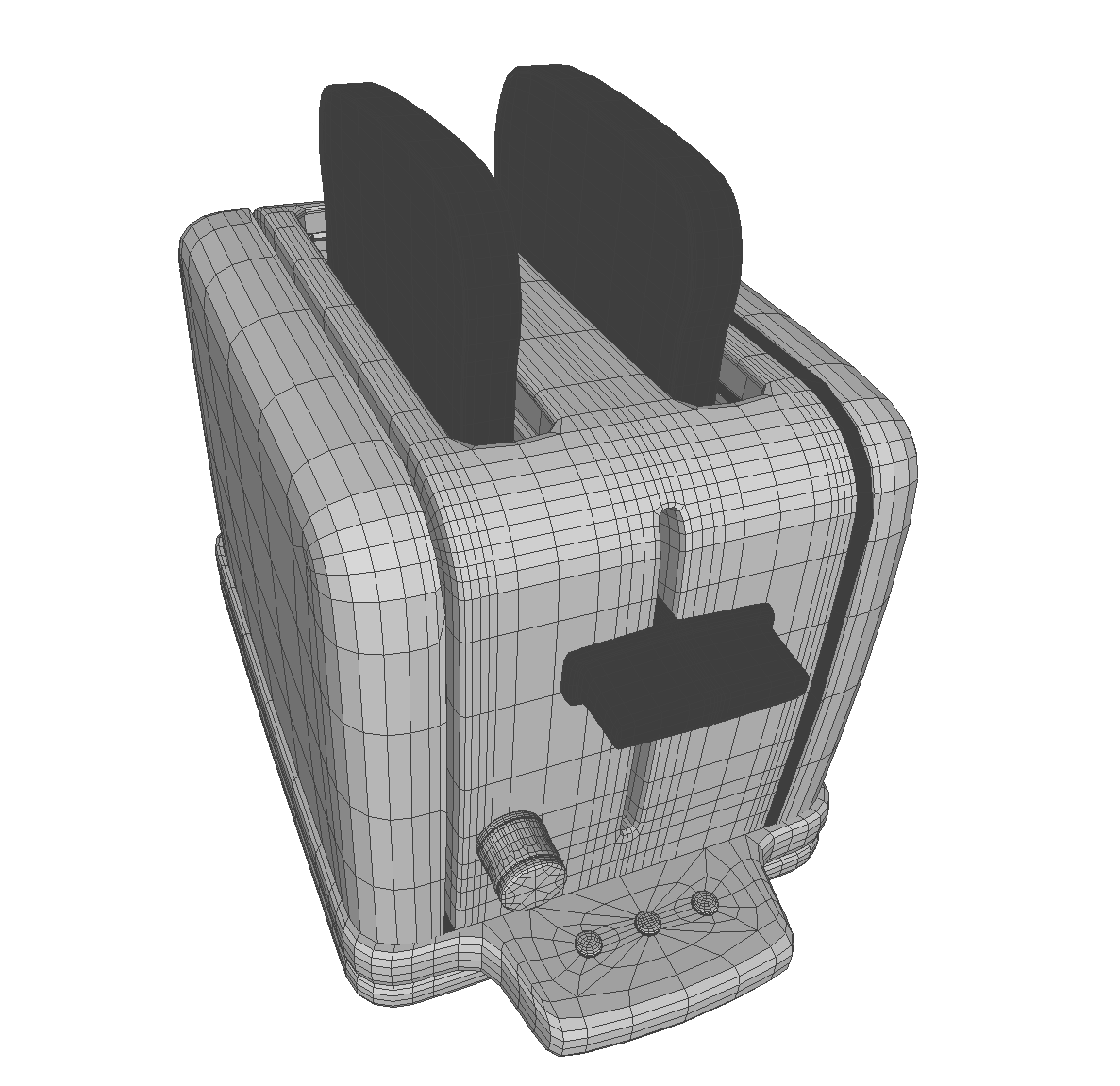} 
    \end{tabular}
    \caption{Ablation study on the Shiny Blender dataset. Multi-view geometric priors work much better than monocular priors for shiny objects.  }
    \label{ablation:qualtitative}
\end{figure*}

\paragraph{Implementation details. }
When using PGSR~\cite{chen2024pgsr}, the state of the art Gaussian-based reconstruction approach, we use their implementation with default values.
We set the function $f$ in \Cref{sub:method:reg} to 
\begin{equation}
    f(k) = \frac{k - 3000}{3000},
\end{equation}
$\lambda_{normal}=0.1$ and $ \lambda_{depth} = 0.1$ in \Cref{eq:lgeo}. We use geometric priors after the 3000th iteration for best convergence. PGSR-based experiments are done on a NVIDIA RTX4090 with 24GB.

\subsection{Impact of VGGT}
To study the effect of using VGGT, we evaluate the geometry estimation of VGGT, in the form of its depth maps, on the synthetic dataset Shiny Blender in~\Cref{exp:shinyblender}. 
Our experiments show that its geometry estimation is not as accurate as what state-of-the-art reconstruction methods can achieve; however, using it in geometric supervision can further improve 3DGS reconstruction (our results). 
The qualitative comparison in ~\Cref{fig:qualtitative} highlights that VGGT can capture the shape for shiny objects, but its geometry is not accurate, e.g., the normals at the side of the car and toaster. 
The erroneous inclination of the flat surface can be rectified when VGGT is integrated with Gaussian Splatting, but only when using the right regularization.

\subsection{Quantitative Evaluation. } \label{subsec:quantitative}

We evaluate our method with respect to geometry and rendering quality, but our strength is in geometric reconstruction which novel view synthesis methods often neglect.

\paragraph{Geometry.} We measure the quality of our results on the DTU, Shiny Blender and TnT datasets in \Cref{exp:dtu}. 
For geometric priors, the improvement in reconstruction is especially significant for shiny objects. 
Our method achieves the best performance among GS-based approaches and is competitive with SDF-based methods that are more time-consuming. 
However, our method shows little improvement on DTU or TnT. This is likely due to the less complex material properties in DTU and TnT, which can be handled well without priors.

\paragraph{Rendering.}
We evaluate the rendering quality of novel views on Shiny Blender on the official test views. Results are reported in \Cref{exp:rendering}. 
Even though the reconstruction quality can be greatly improved, the rendering quality is on par regardless of additional regularization.
This is expected, as the regularization is only of geometric nature, but proves that reconstruction is not improved at the price of rendering.

\begin{table*}
    \centering
     \setlength{\tabcolsep}{4.4pt}
    \begin{tabularx}{\textwidth}{X|c c c c c|c}
    \hline
        \textbf{Shiny Blender} &  Car & Coffee & Helmet & Teapot & Toaster & Mean\\ \hline
        PGSR & \textbf{26.43} & 30.86 & 25.72 & 37.08 & \textbf{20.24} & \textbf{28.07} \\ \hline
        Ours &26.31 & \textbf{30.95} & \textbf{25.73} & \textbf{37.58} & 19.69 & 28.05 \\ \hline
    \end{tabularx}

    \caption{Novel view rendering on the test views of the Shiny Blender. We report the results in PSNR (higher is better). The rendering quality when using geometric priors does not improve but also not degrade, even though the reconstruction accuracy (see other experiments) is enhanced significantly. }
    \label{exp:rendering}
\end{table*}

\subsection{Qualitative Evaluation} \label{subsec:qualitative}

We provide qualitative examples in \Cref{fig:qualtitative}. 
It is clearly visible that PGSR produces significant artifacts on reflective surfaces, which our proposed solution solves without degrading rendering quality in \Cref{exp:rendering}. 
This is consistent with the quantitative results on DTU (\Cref{exp:dtu}), where the improvements are minimal, but objects in this dataset are mostly Lambertian.

\begin{figure}[b]
\resizebox{1.0\linewidth}{!}{
    \centering
    \begin{tabular}{cccc}
  
    w/o confidence & Ours & w/o confidence & Ours \\
    \includegraphics[width=0.24\linewidth]{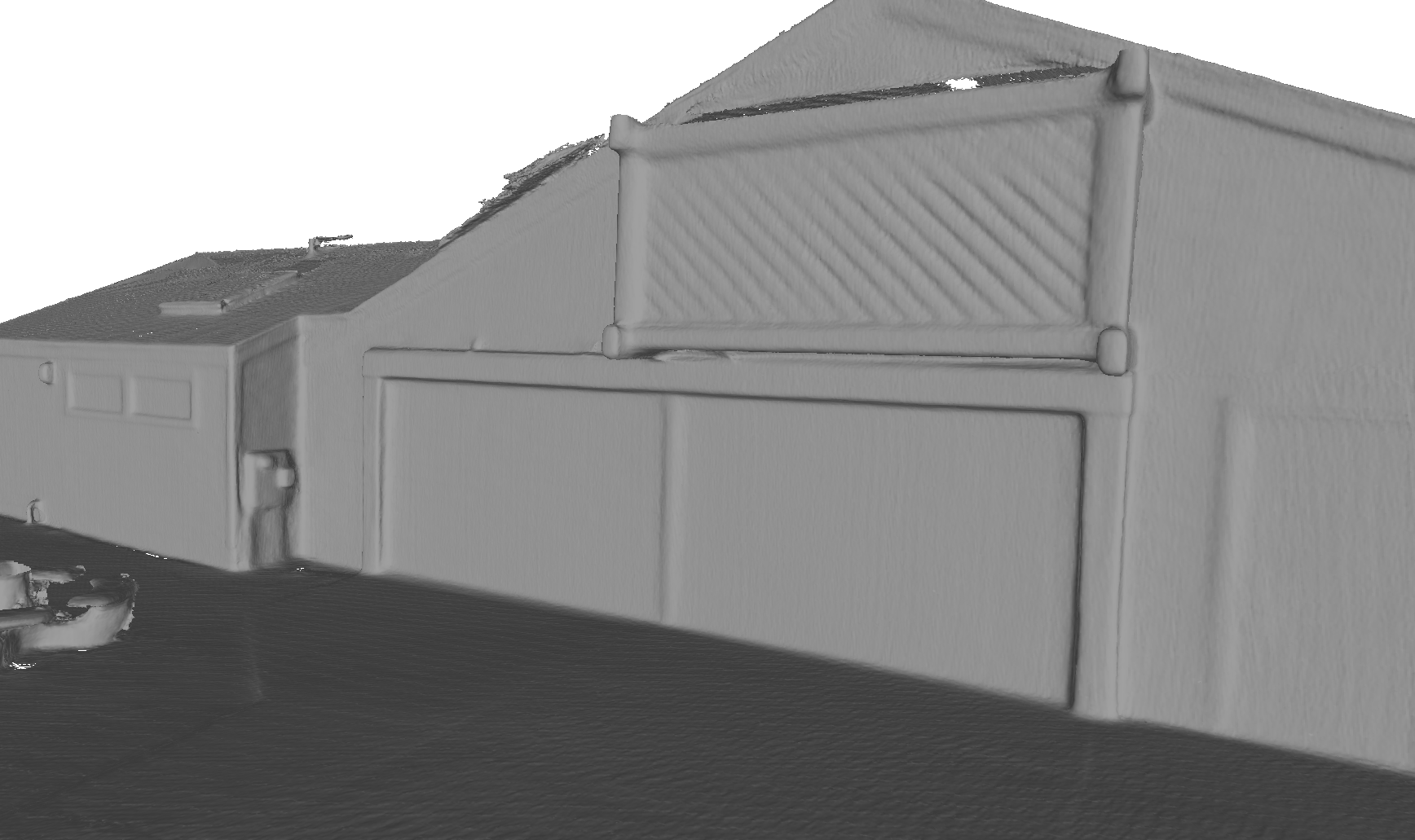} &
    \includegraphics[width=0.24\linewidth]{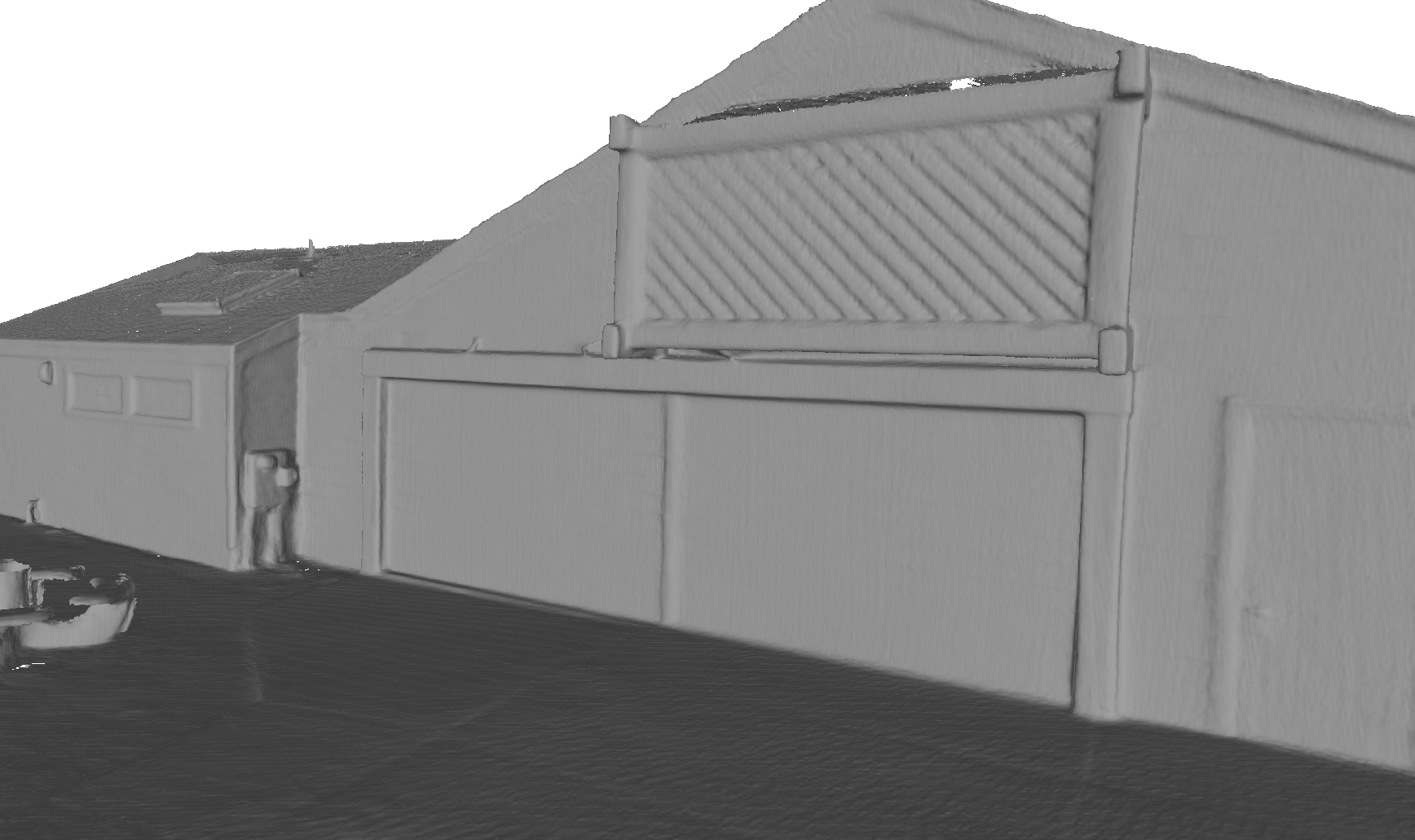} &
    
    \includegraphics[width=0.24\linewidth]{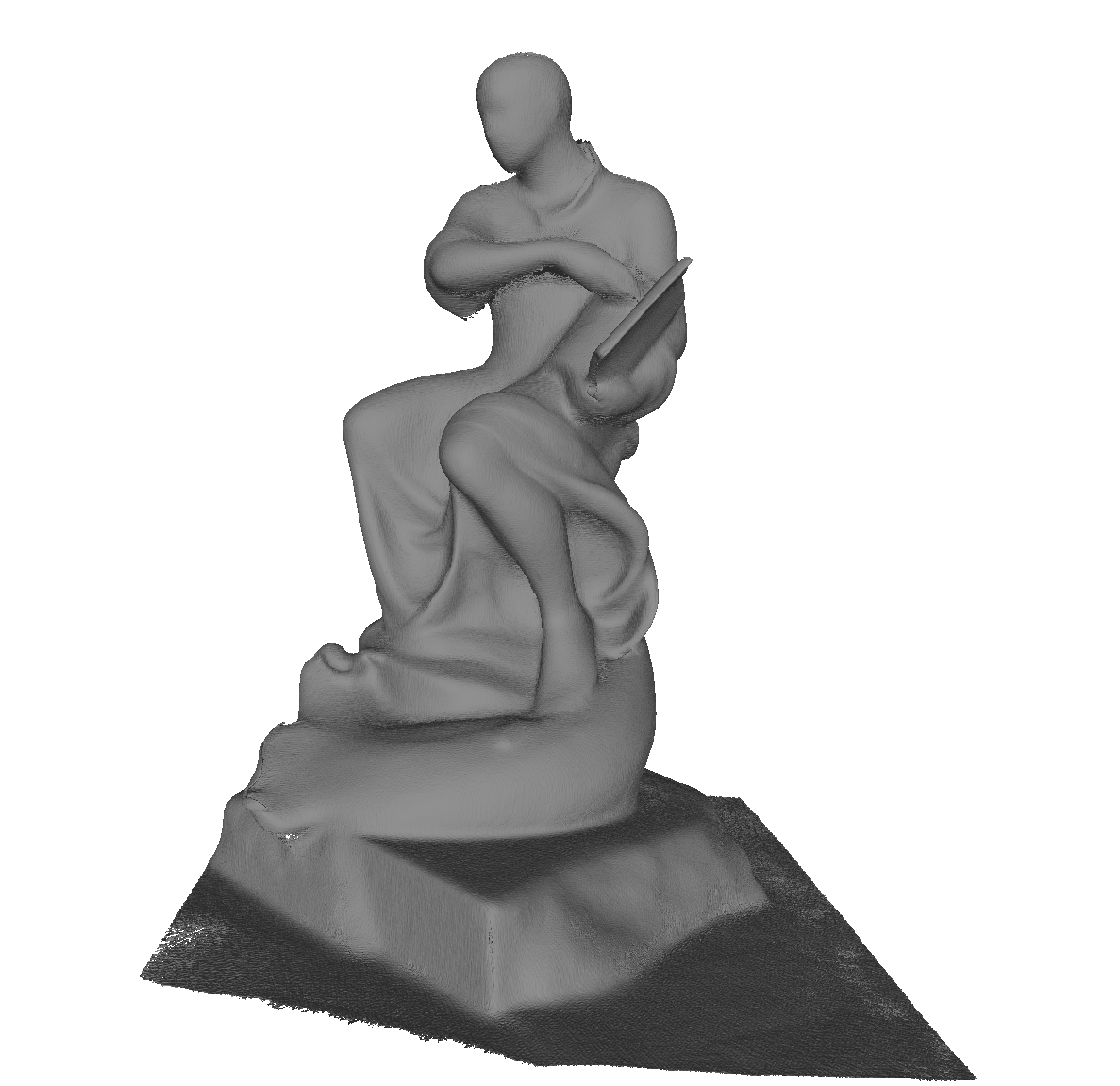} &
    \includegraphics[width=0.24\linewidth]{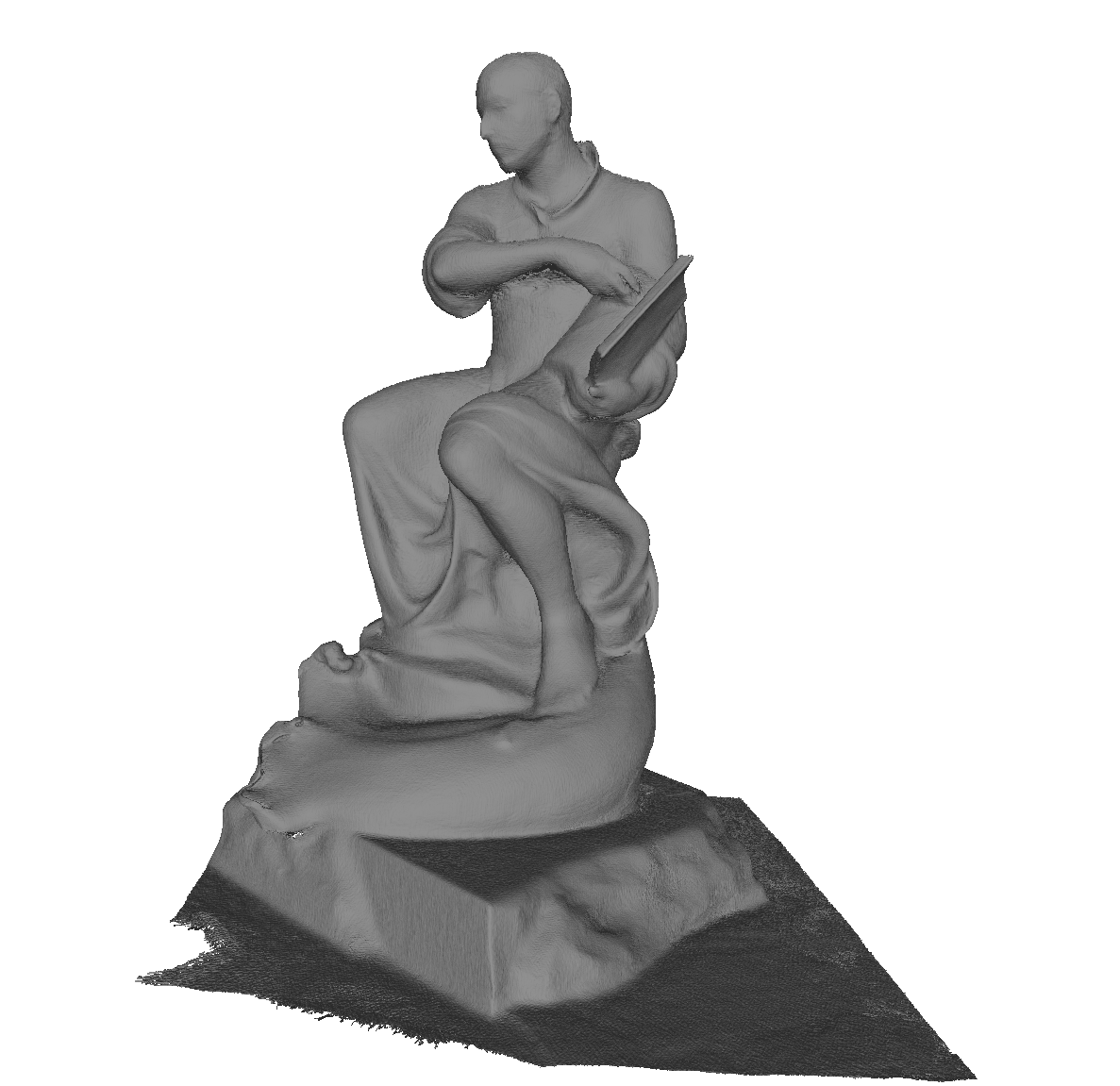}

    \end{tabular}
    }
    \caption{Ablation study on confidence weighting. Ours using confidence weighting reconstruct more details for the Barn and Ignatius scene in the TnT dataset.}
    \label{fig:ablation:weight}
\end{figure}

\begin{table}[t]
    \centering
     \setlength{\tabcolsep}{8.4pt}
    \begin{tabularx}{\textwidth}{X|c c c c c}
    \hline
        Vision models & VGGT & VGGT & SN & DA & {\centering /}  \\
        Confidence-based weighting & \ding{52} & \ding{55} &  {\centering /}  &  {\centering /} &  {\centering /}   \\\hline
        TnT~(F1 $\uparrow$) & \bf{0.40} & 0.37 & 0.35 & 0.03 & 0.38 \\ 
        DTU~(CD $\downarrow$) & \bf{0.52} & 0.57 & 0.57 & 0.78 & 0.53 \\ \hline
    \end{tabularx}
    \caption{Ablation study on vision model selection and confidence-based weighting. For notation, we take SN for StableNormal and DA for Depth Anything V2. VGGT improves reconstruction when combined with confidence weighting. Multi-view model as VGGT outperforms monocular estimation as StableNormal. However, without the confidence, both of them degrade the performance compared to the baseline. Depth Anything V2 can even fail the reconstruction completely on large scenes.}
    \label{ablation:tnt} \label{ablation:dtu}
\end{table}

\subsection{Ablation study} \label{subsec:ablation}
We study the effect of confidence-based weighting and the advantage of multi-view geometric priors over monocular priors.
We use StableNormal~\cite{ye2024stablenormal} and Depth Anything V2~\cite{yang2024depth} as monocular normal priors and depth priors, respectively. 
For multi-view priors, we use VGGT~\cite{wang2025vggt} for both normal and depth.
We conduct the ablation study on Barn, Courthouse and Meetingroom in the TnT dataset, and on all the scenes in the DTU dataset quantitatively. 
The mean score is reported in \Cref{ablation:tnt}. We show qualitative comparisons in \Cref{ablation:qualtitative} and \Cref{fig:ablation:weight} on Shiny Blender and TnT.

\paragraph{Confidence-based weighting.}
The confidence map from VGGT is a useful tool to distinguish errors. 
After using our weighting, the multi-view priors improve the reconstruction for almost all of the scenes. 
Without the confidence-based weighting, many details are lost because of the limited resolution of vision models, as shown in \Cref{fig:ablation:weight}. Adding confidence-based weighting solves this problem and even improves the reconstruction by a small margin.

\paragraph{Multi-view priors vs. monocular priors.}
Multi-view priors work better than monocular priors on all of the datasets. 
In addition, monocular priors do not provide a confidence score by design, so it is not possible to add confidence-based weighting.
For depth priors, monocular estimation and its scale ambiguity can mislead the reconstruction and reduce the quality significantly, which might be due to invalid assumptions during affine alignment. 
While the monocular depth gives a fair estimation of relative positions, the metric relationship between monocular depth estimation and ground truth depth is ambiguous. 
This ambiguity makes the monocular depth estimation difficult to use. 
However, this problem can be solved in multi-view settings as long as the object appears twice in different views. 
It is visible from our experiments that even if the multi-view estimation is obtained without strict multi-view stereo, as is the case for VGGT, it persistently increases the reconstruction quality. 
This phenomenon of misalignment for depth priors becomes severe for multi-object scenes such as the TnT dataset, as shown in ~\Cref{exp:tnt}. 

\section{Conclusion} \label{sec:conclusion}

We presented an in-depth analysis and experimental evaluation of the effects of multi-view geometric priors in the form of normal and depth maps in 3D Gaussian splatting methods, with a special focus on the geometric reconstruction quality. 
Our results show that in challenging cases, especially on highly reflective surfaces, the multi-view geometric priors have a significant positive effect. 
The confidence map produced as a by-product of multi-view models actually plays a central role in reducing the impact of false predictions and adds robustness in our framework, especially in complex scenes with multiple objects.
In conclusion, we found that the multi-view geometric priors provide important signals for reconstruction from 3DGS and can likely be used as plug-in regularization in existing methods. 
We hope this will provide meaningful information for future reconstruction work regarding how to incorporate geometric priors.

\noindent\textbf{Acknowledgements. }
The project was supported by the DFG grant LA 5191/2-1. 
We gratefully acknowledge access to the Marvin cluster of the University of Bonn.

%
%
%
%
\bibliographystyle{splncs04}
\bibliography{egbib}

\newpage
{\centering\Large\textbf{Confidence matters: Leveraging Multi-view Geometric Priors for GS-based Reconstruction}\\
\vspace{0.5em}Supplementary Material \\
\vspace{1.0em}}

\renewcommand{\thesection}{\Alph{section}}
\setcounter{figure}{0}
\setcounter{section}{0}
\setcounter{table}{0}
\renewcommand\thefigure{A.\arabic{figure}}  

\section{Multi-view \& monocular geometric priors}\label{supp:sec:priors}
\begin{figure*}
    \centering
    \resizebox{\textwidth}{!}{
    \begin{tabular}{cccccc}
         Image & SN & DA & VGGT(N) & VGGT(D) & Confidence  \\
         \includegraphics[width=0.15\textwidth]{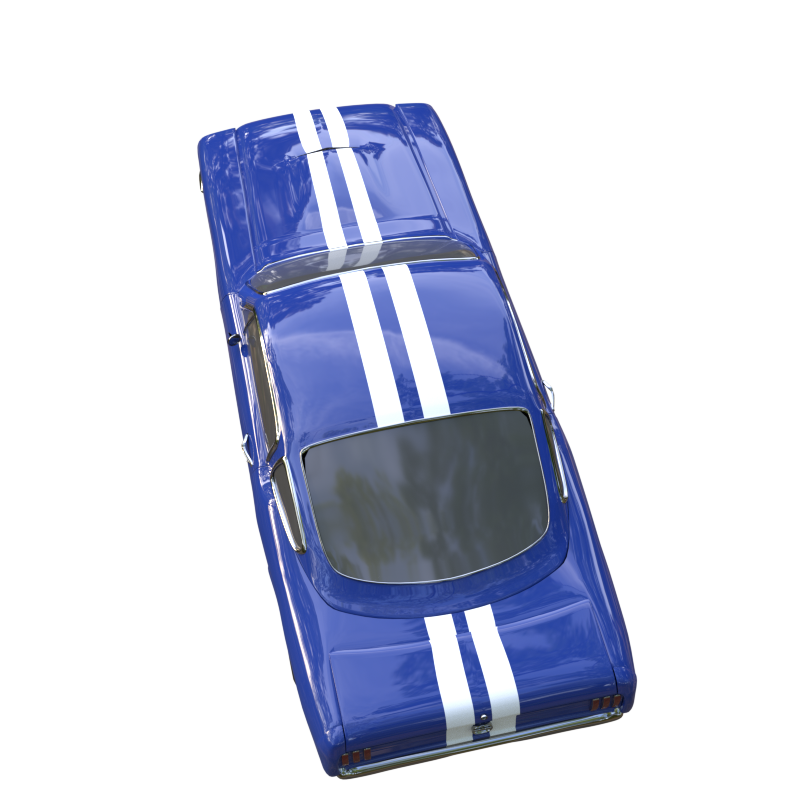} &
         \includegraphics[width=0.15\textwidth]{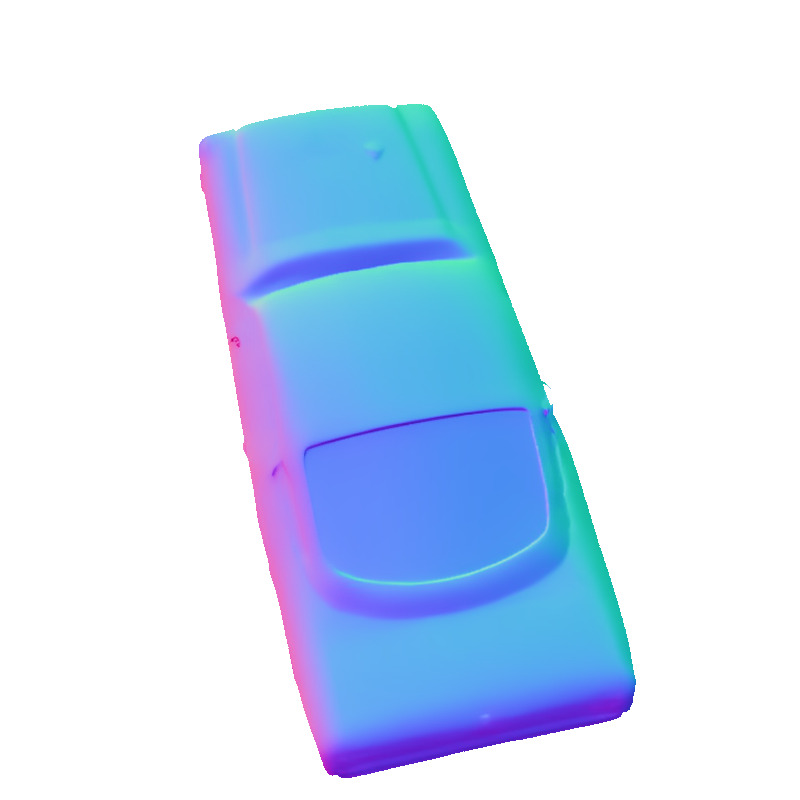} &
         \includegraphics[width=0.15\textwidth]{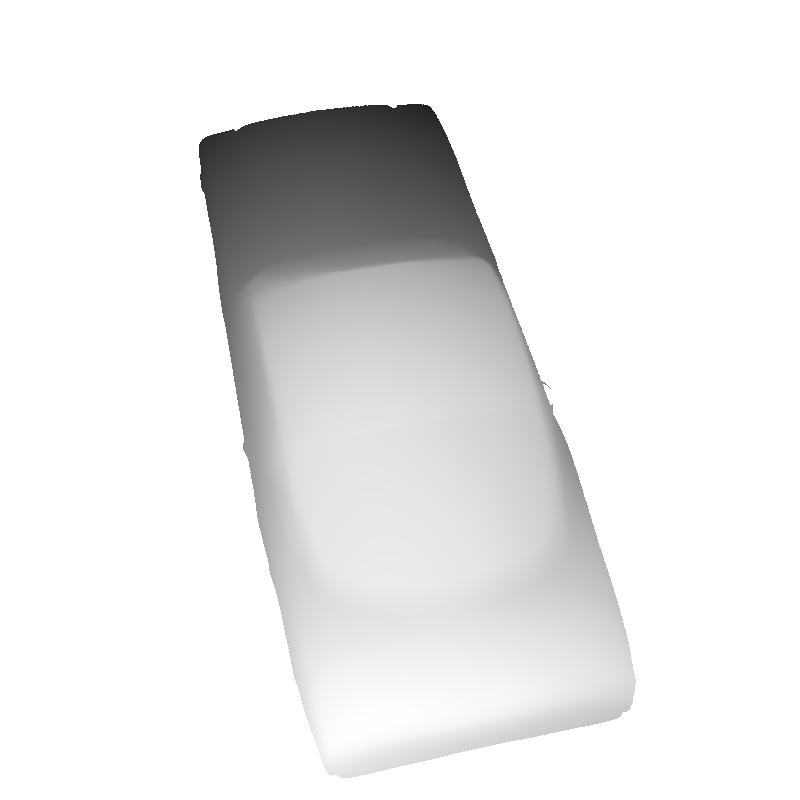} &
         \includegraphics[width=0.15\textwidth]{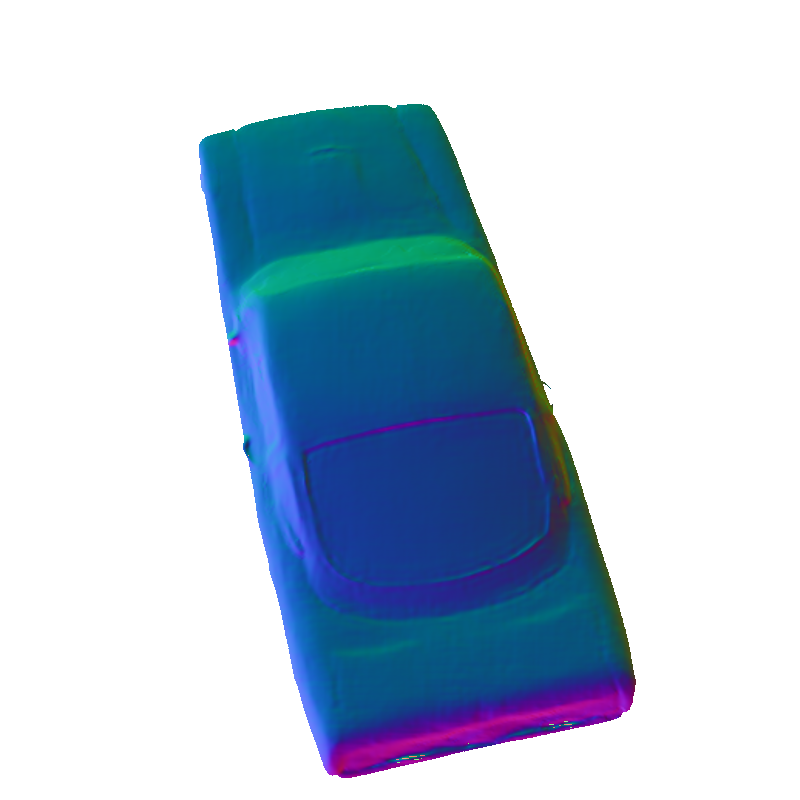} &
         \includegraphics[width=0.15\textwidth]{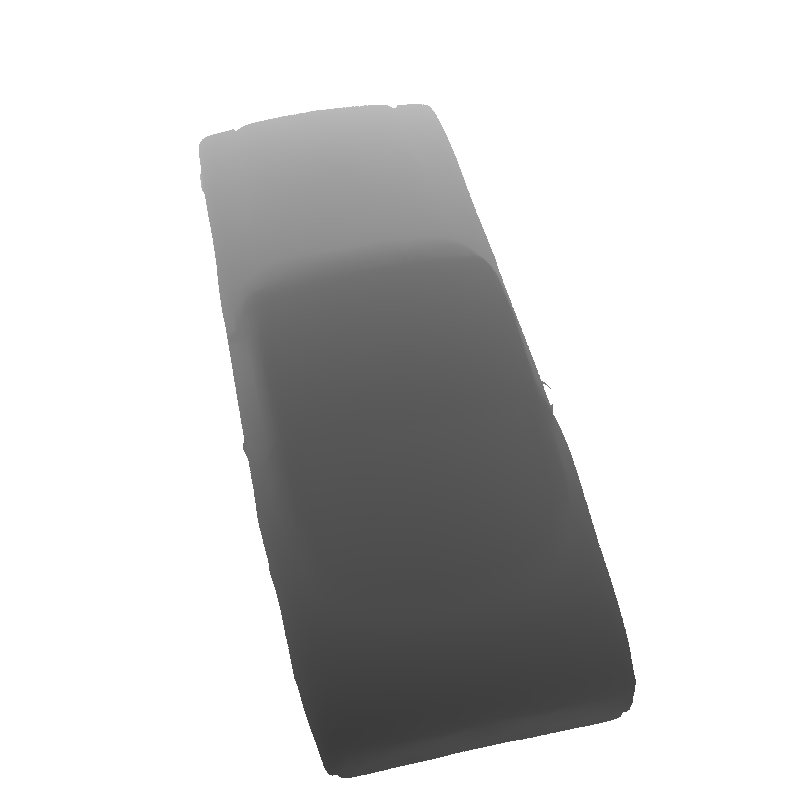} &
         \includegraphics[width=0.15\textwidth]{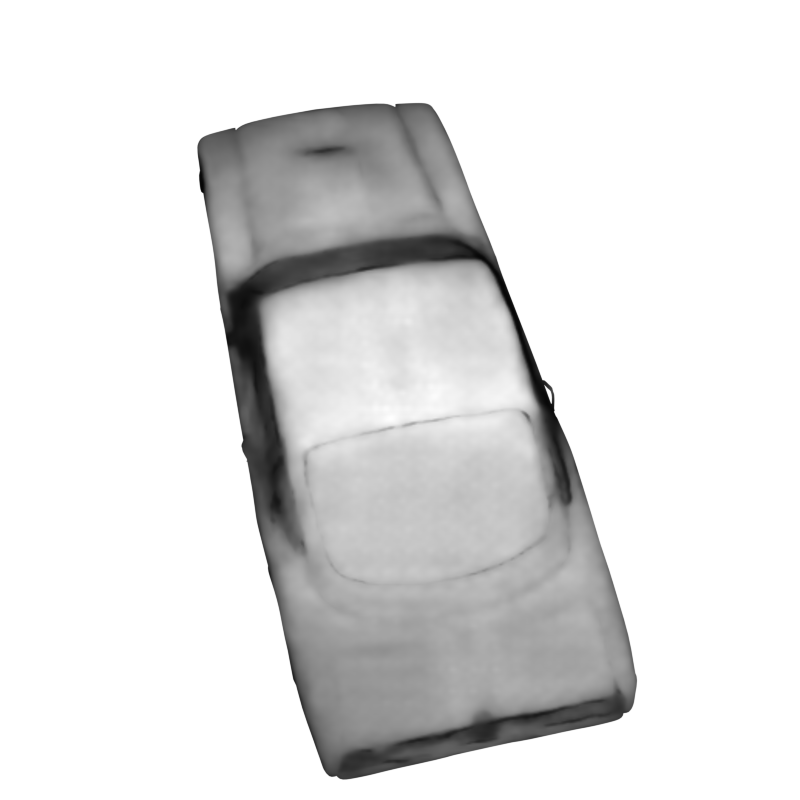} \\

         \includegraphics[width=0.15\textwidth]{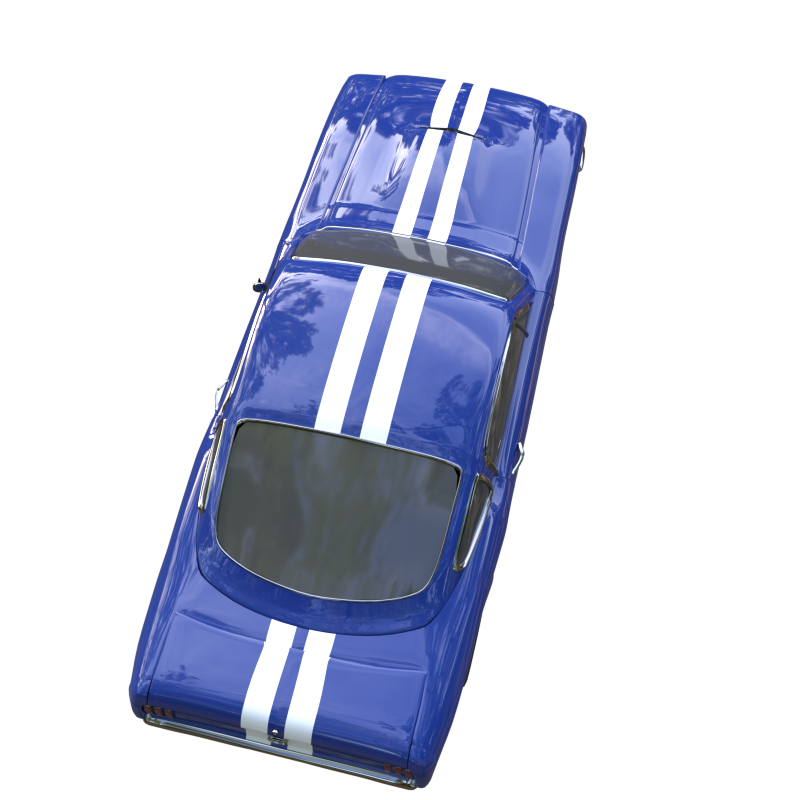} &
         \includegraphics[width=0.15\textwidth]{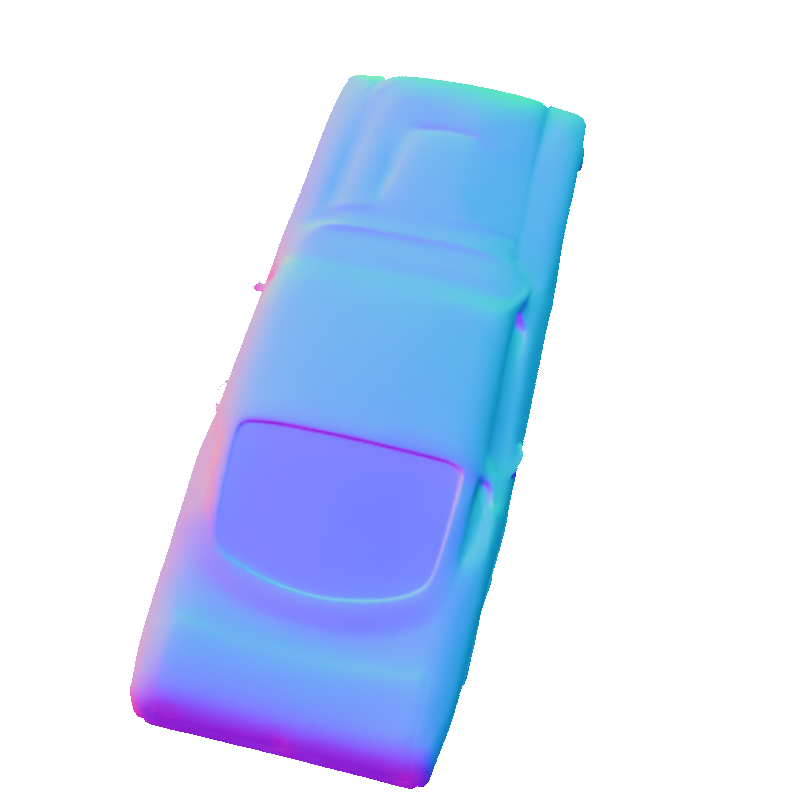} &
         \includegraphics[width=0.15\textwidth]{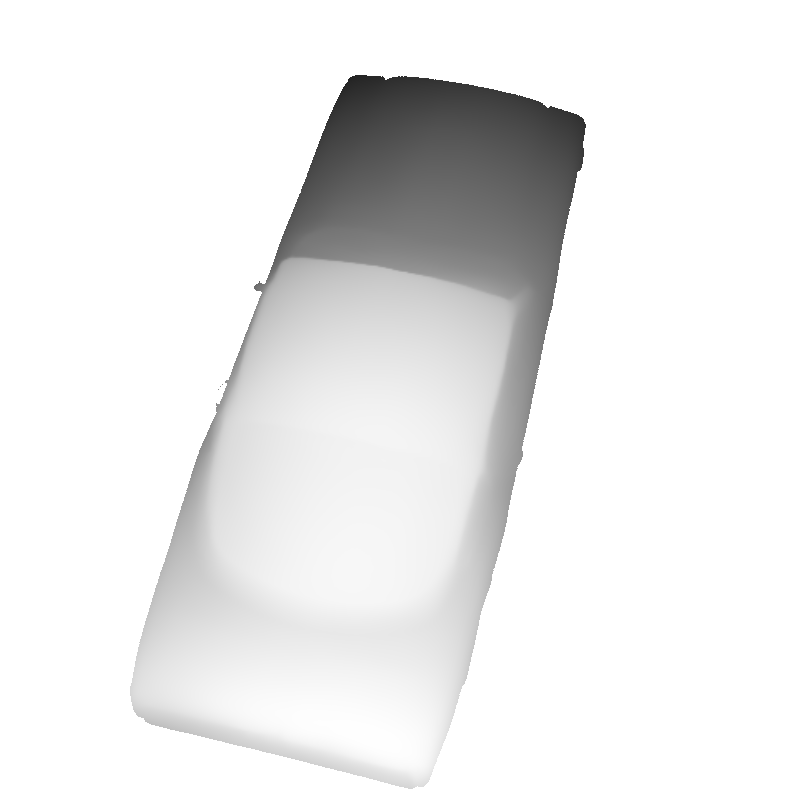} &
         \includegraphics[width=0.15\textwidth]{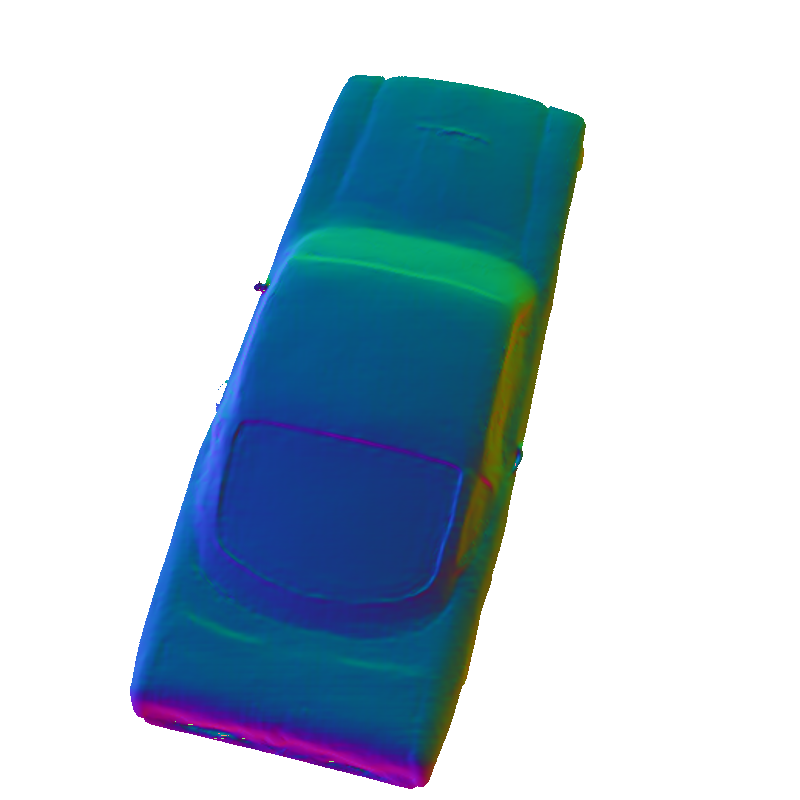} &
         \includegraphics[width=0.15\textwidth]{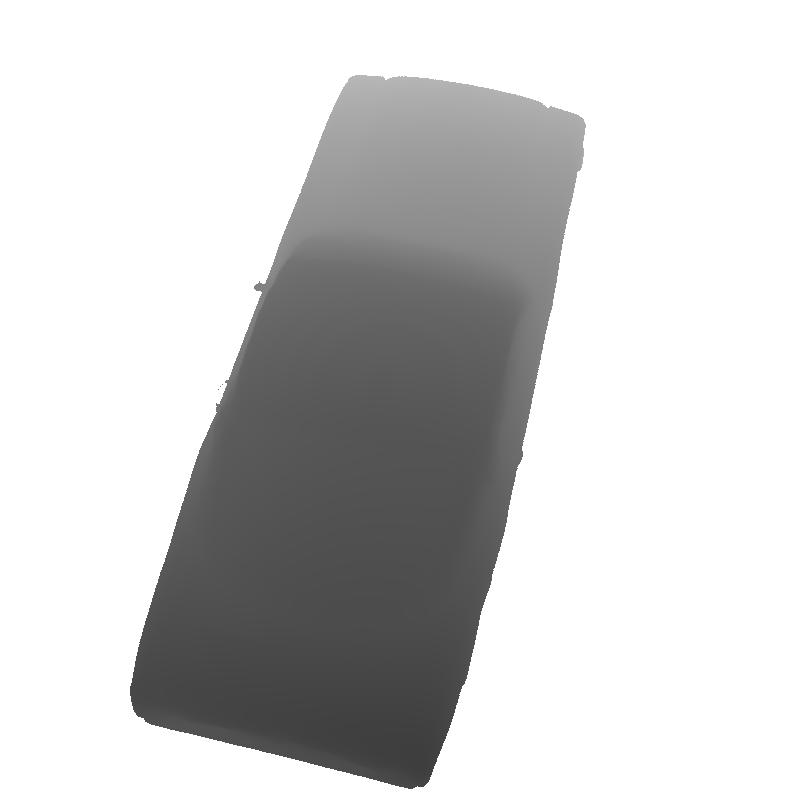} &
         \includegraphics[width=0.15\textwidth]{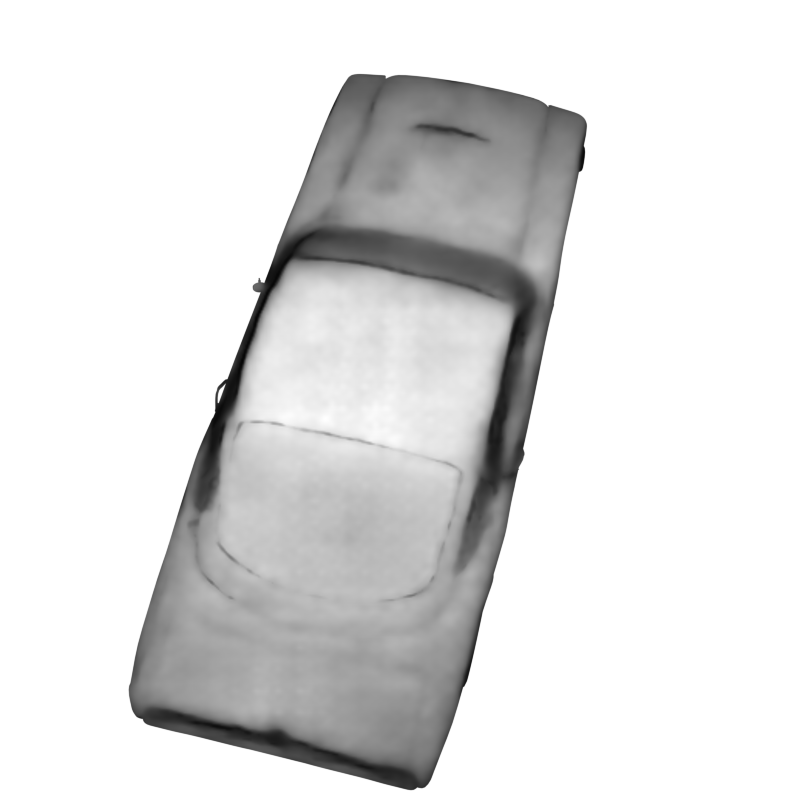} \\

         \includegraphics[width=0.15\textwidth]{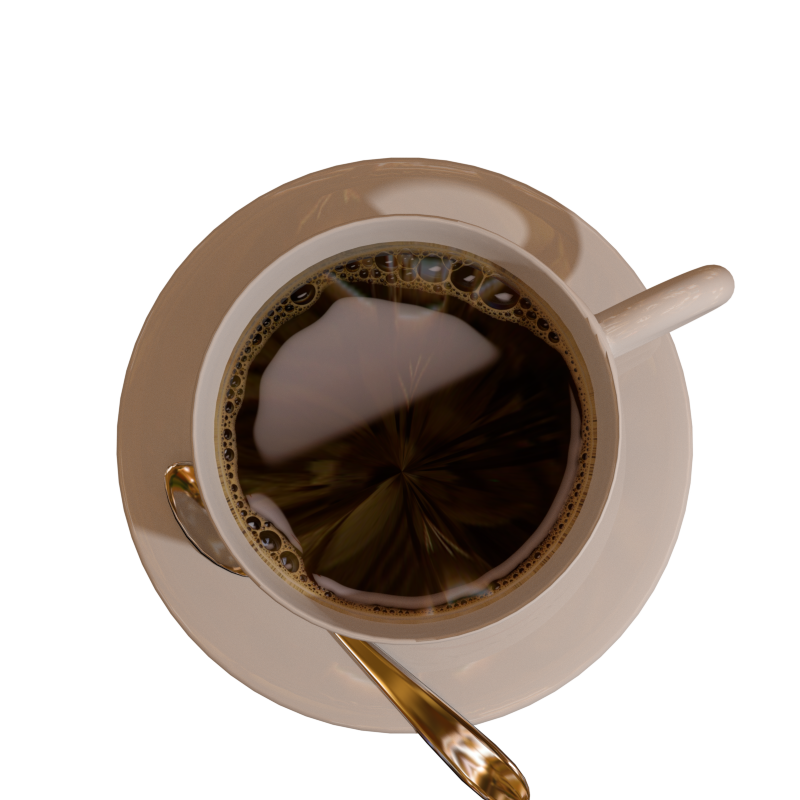} &
         \includegraphics[width=0.15\textwidth]{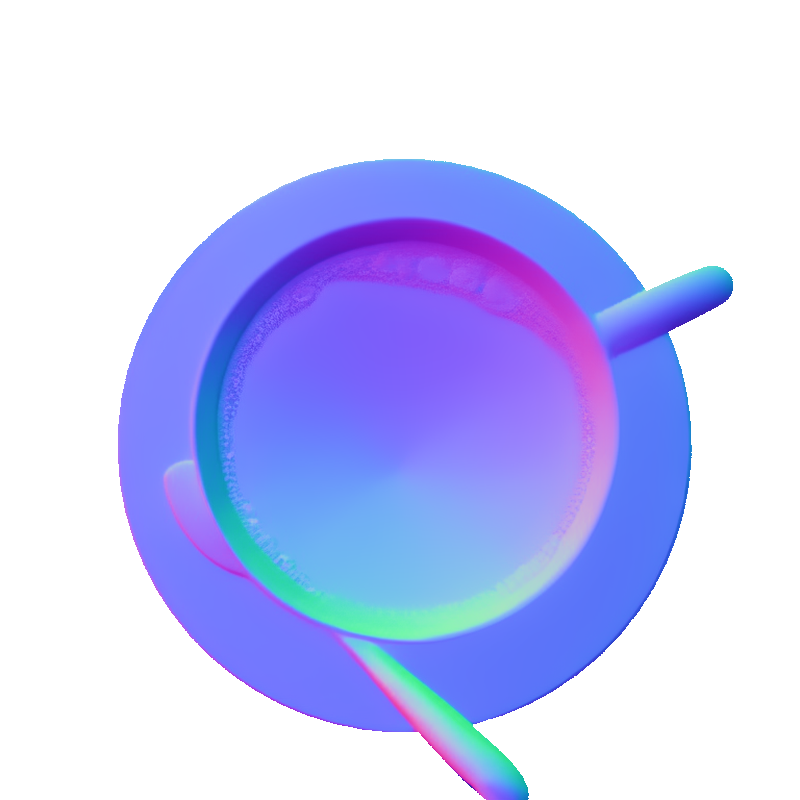} &
         \includegraphics[width=0.15\textwidth]{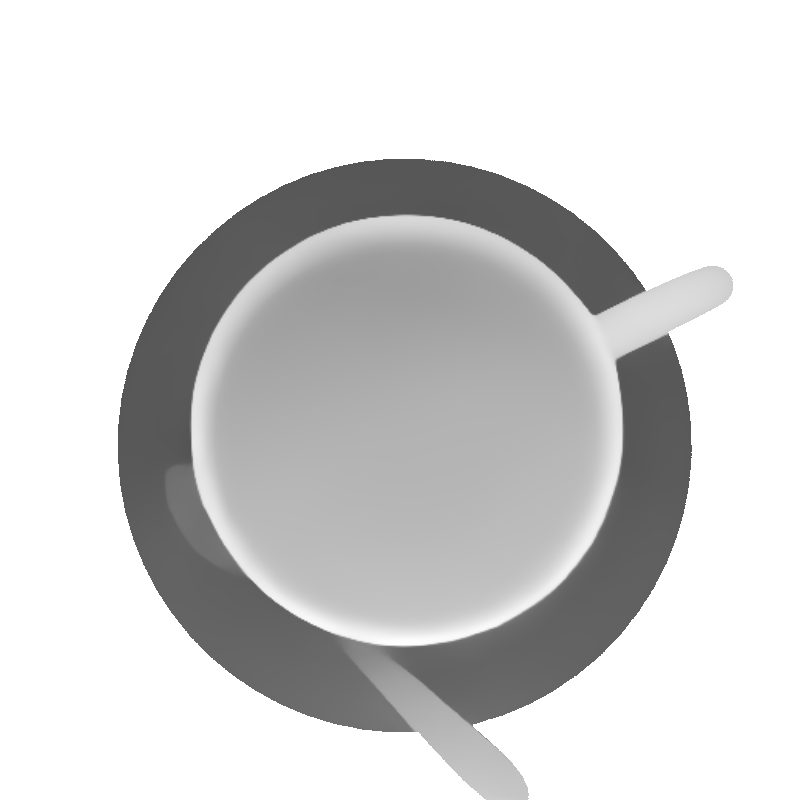} &
         \includegraphics[width=0.15\textwidth]{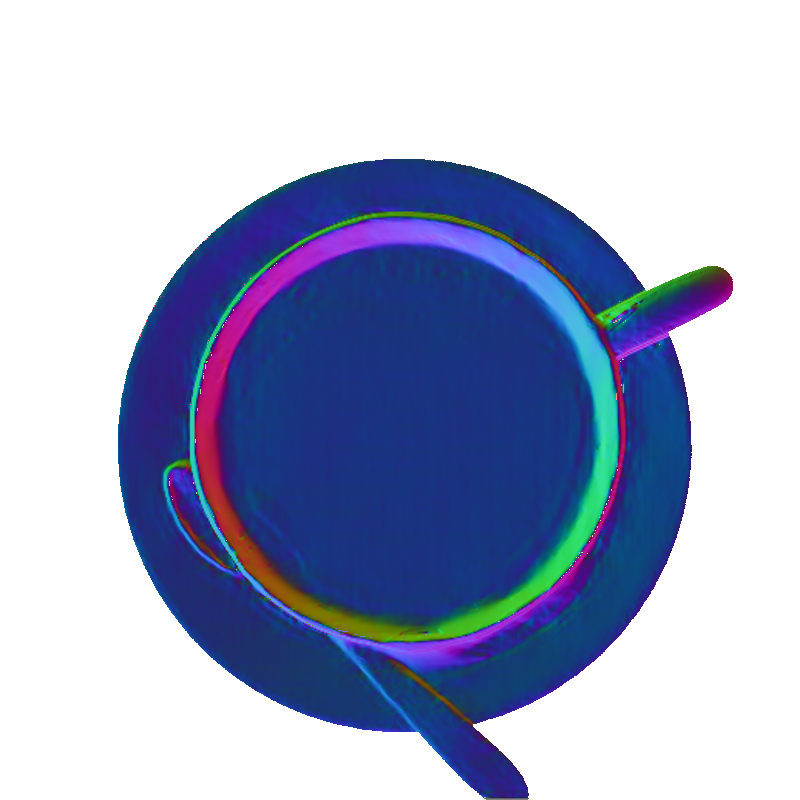} &
         \includegraphics[width=0.15\textwidth]{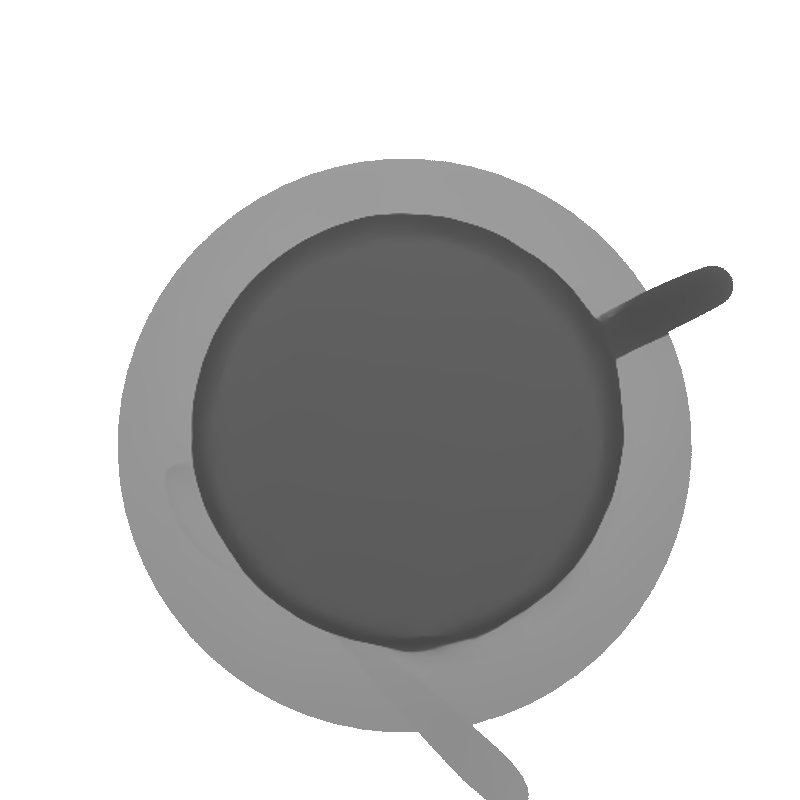} &
         \includegraphics[width=0.15\textwidth]{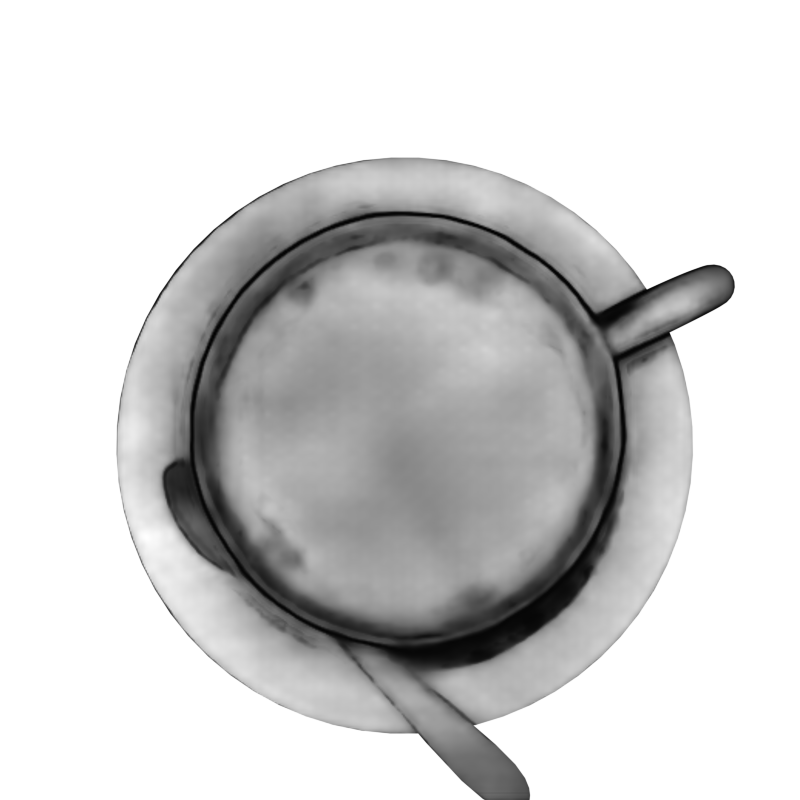} \\

         \includegraphics[width=0.15\textwidth]{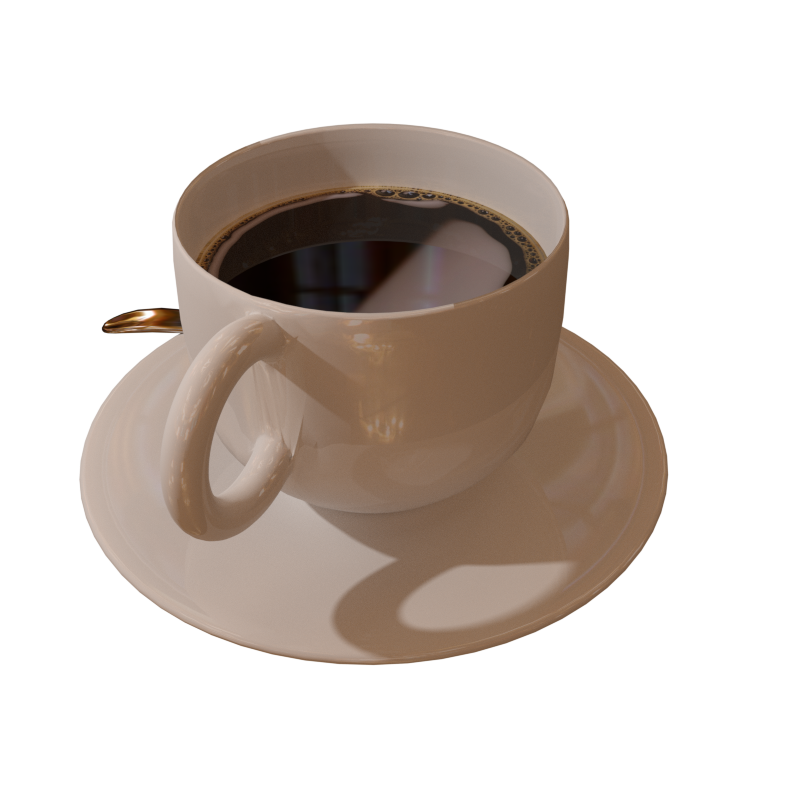} &
         \includegraphics[width=0.15\textwidth]{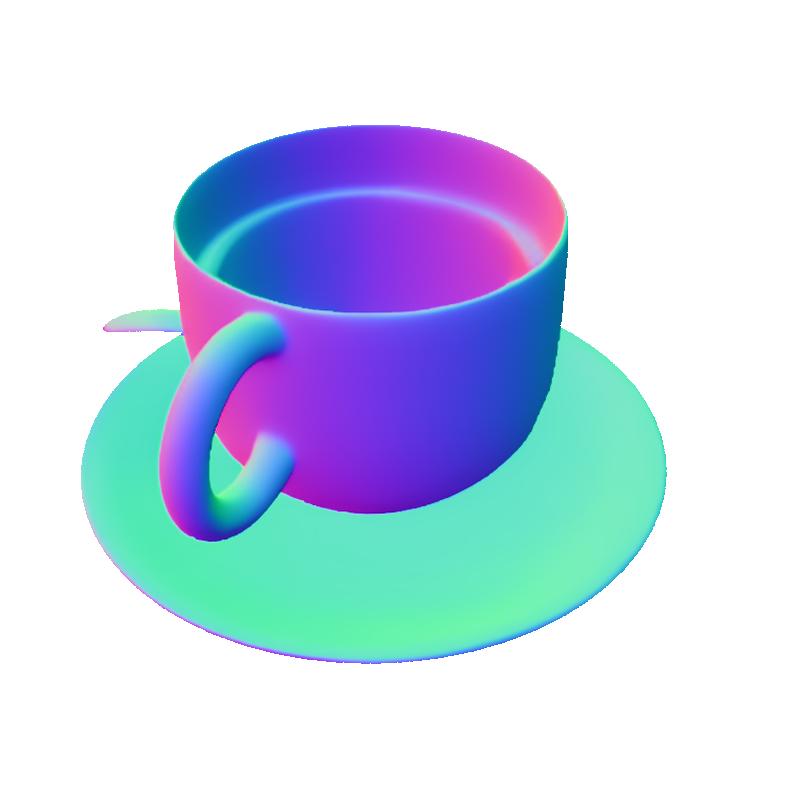} &
         \includegraphics[width=0.15\textwidth]{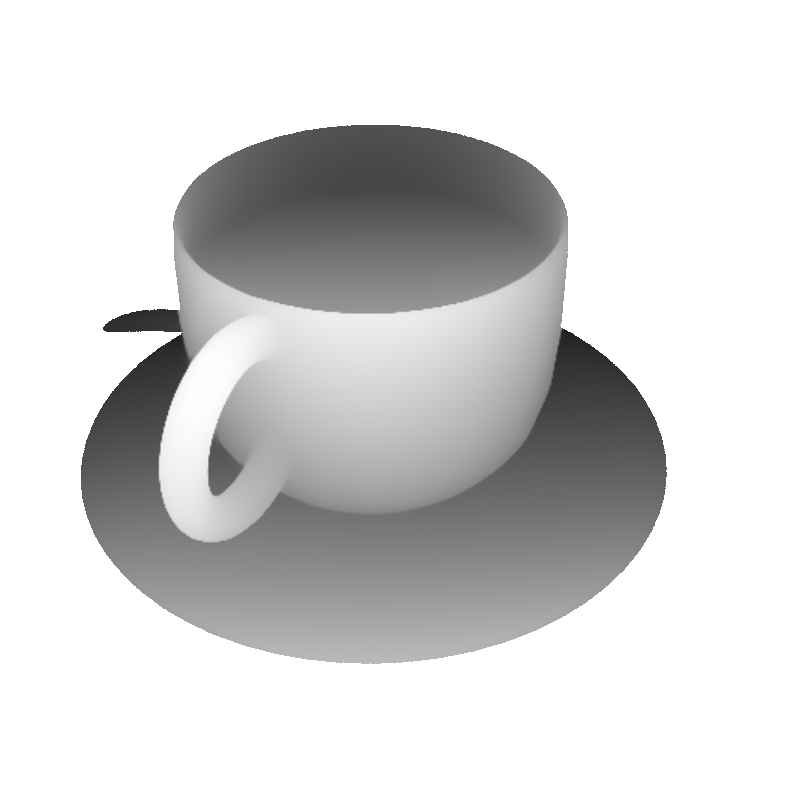} &
         \includegraphics[width=0.15\textwidth]{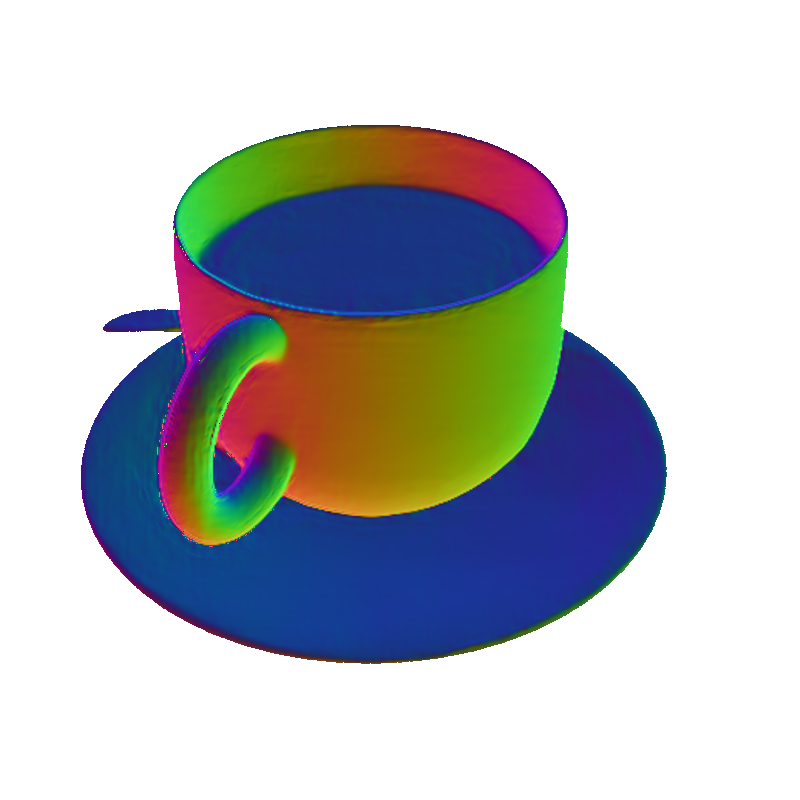} &
         \includegraphics[width=0.15\textwidth]{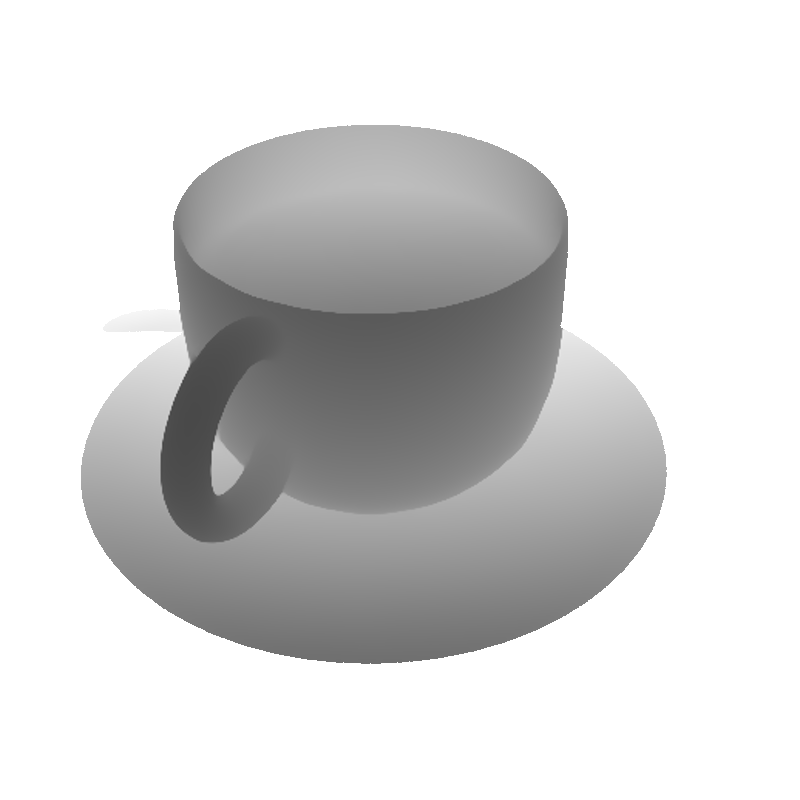} &
         \includegraphics[width=0.15\textwidth]{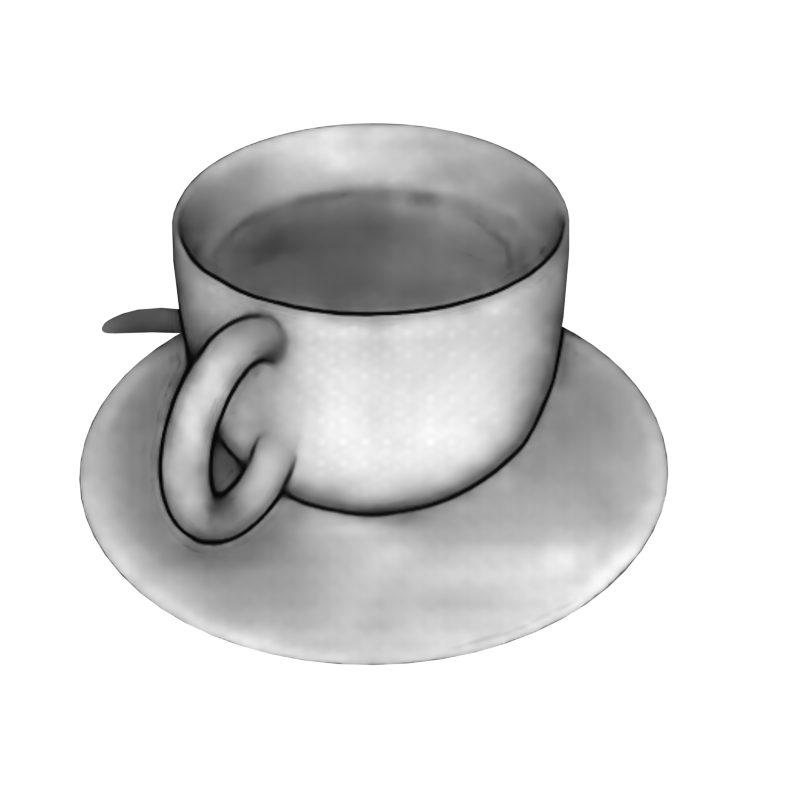} \\
    \end{tabular}
     }
     \caption{Multi-view geometric priors and monocular geometric priors on Shiny Blender. For the confidence, the darker, the less confident. Monocular normal estimation has no consistency as the stable normal of the car's front window changes drastically when the rotation is very small. The monocular estimation is also more error-prone, which is visualized in the coffee images.}
     \label{supp:fig:priors}
\end{figure*}

We visualize the multi-view geometric priors against monocular estimations in \cref{supp:fig:priors}. We show the geometric estimation of StableNormal~(SN)~\cite{ye2024stablenormal}, Depth-anything V2~(DA)~\cite{yang2024depth}, the VGGT~(N)~\cite{wang2025vggt} and the VGGT~(D) on Car and Coffee, two objects from Shiny Blender. The monocular normal estimation lacks consistency even when the car rotates by a small angle, while the derived multi-view normal estimation keeps consistency. Moreover, the monocular normal estimation is more error-prone. It completely fails to estimate the normal of the coffee surface, while the multi-view estimation is relatively correct and robust. The confidence map with the multi-view estimation gives a low value to surfaces at the side and gives a high value to the surfaces up-front. 

\section{Details on Ablation study}\label{supp:ablation}
We provide quantitative details in \cref{supp:ablation:DTU} and \cref{supp:ablation:tnt}. Note that we only use the results on Barn, Courthouse and Meetingroom in the TnT dataset in the ablation study in the main paper because DA breaks down on other scenes. In \cref{supp:ablation:tnt}, we also report the results on Caterpillar and Ignatius.

 \begin{table*}[h]
    \centering
    \resizebox{\textwidth}{!}{
    \begin{tabular}{l|c c c c c c c c c c c c c c c|c}
    \hline
          \textbf{DTU} & 24 & 37 & 40 & 55 & 63 & 65 & 69 & 83 & 97 & 105 & 106 & 110 & 114 & 118 & 122 & Mean\\ \hline
          PGSR & 0.34 & 0.58 & 0.29 & 0.29 & 0.78 & 0.58 & 0.54 & 1.01 & 0.73 & 0.51 & 0.49 &  0.69 & 0.31 & 0.37 & 0.38 & 0.53 \\
              +SN & 0.42	&0.51&	0.35&	0.34&	0.77&	0.67&	0.57&	1.17&	0.83&	0.64&	0.48&	0.61&	0.37&	0.52&	0.41&	0.58 \\ 
              +DA & 0.38	&0.54	&0.38&	0.34&	1.06&	1.67&	0.87&	1.71&	1.28&	0.61&	0.90&	0.750&	0.33&	0.44&	0.43&	0.78 \\ 
              +VGGT~(w/o conf) & 0.36&	0.55&	0.33&	0.35&	0.88&	0.59&	0.52&	1.31&	0.71&	0.69&	0.41&	0.59&	0.35&	0.50&	0.41&	0.57 \\
              Ours & 0.33&	0.53&	0.33&	0.34&	0.83&	0.54&	0.48&	1.14&	0.65&	0.62&	0.38&	0.59&	0.31&	0.37&	0.36&	0.52 \\ \hline
    \end{tabular}
    }
    \caption{Ablation study on DTU.}
    \label{supp:ablation:DTU}
    
\end{table*}

\begin{table*}[h]
    \centering
    \begin{tabular}{l|c c c c c}
    \hline
        & Barn & Courthouse & Meetingroom & Caterpillar & Ignatius  \\ \hline
        PGSR & 0.66 & 0.21 & 0.29 & 0.41 & 0.80\\
        +SN & 0.62 & 0.10 & 0.34 & 0.38 & 0.67\\
        +DA & 0.02 & 0.05 & 0.03 & {\centering /} & {\centering /}\\
        +VGGT~(w/o conf)& 0.63 & 0.14 & 0.33 & 0.27 & 0.61\\
        Ours & 0.65 & 0.19 & 0.37 & 0.44 & 0.74\\ \hline
    \end{tabular}
    \caption{Ablation study on TnT.}
    \label{supp:ablation:tnt}
\end{table*}

\section{More qualitative results}\label{supp:qualitative:geometry}
\begin{figure*}
\setlength{\tabcolsep}{1pt} 
    \renewcommand{\arraystretch}{1} 
    \centering
    \begin{tabular}{cccccc}
    PGSR & +SN & +DA & \makecell[tc]{+VGGT\\(w/o conf)} & Ours & GT image\\
    \includegraphics[width=0.15\textwidth]{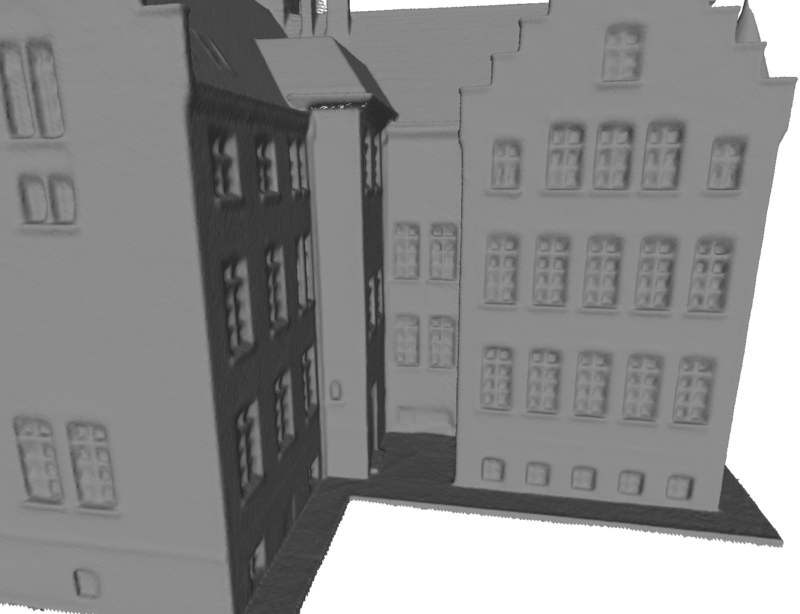} &
    \includegraphics[width=0.15\textwidth]{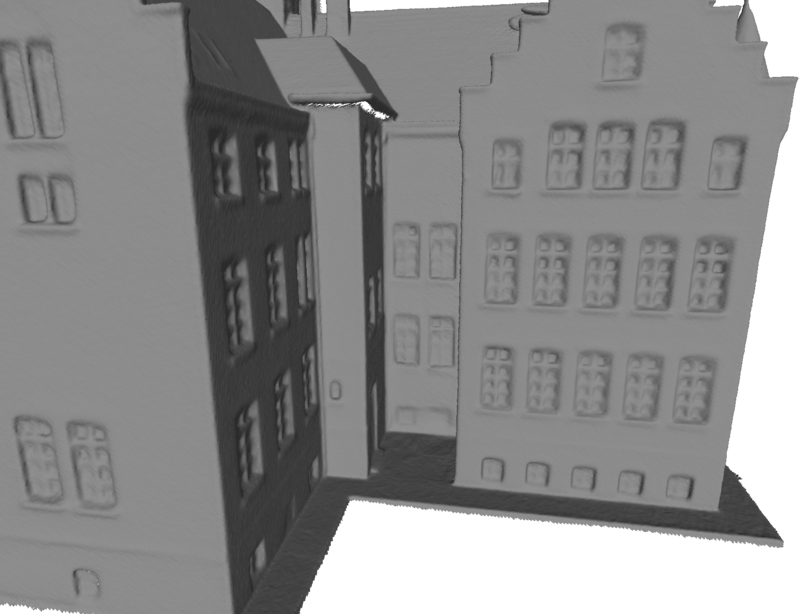} &
    
    \includegraphics[width=0.15\textwidth]{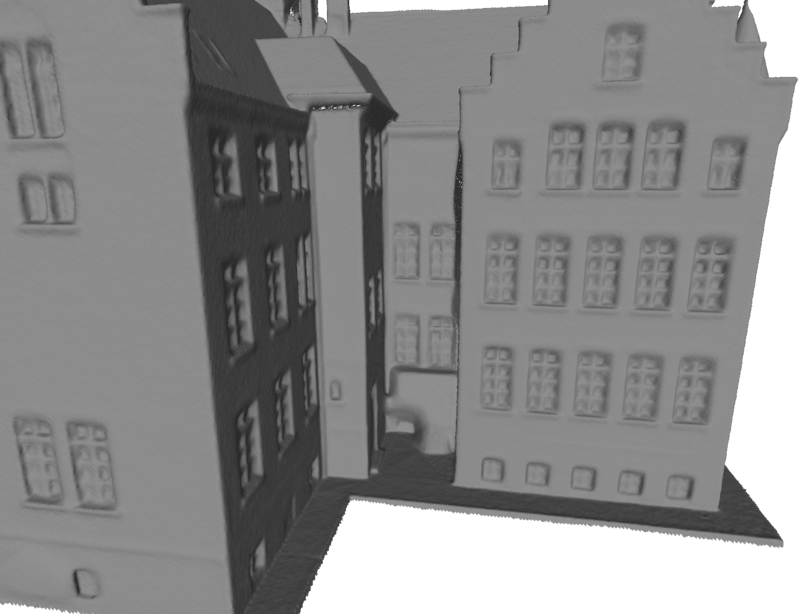} &
    
    \includegraphics[width=0.15\textwidth]{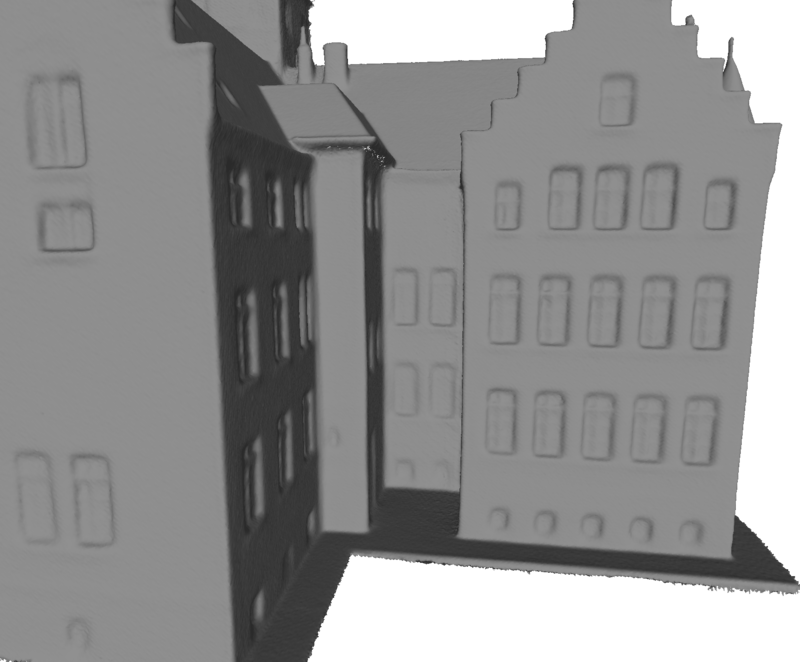} &
    \includegraphics[width=0.15\textwidth]{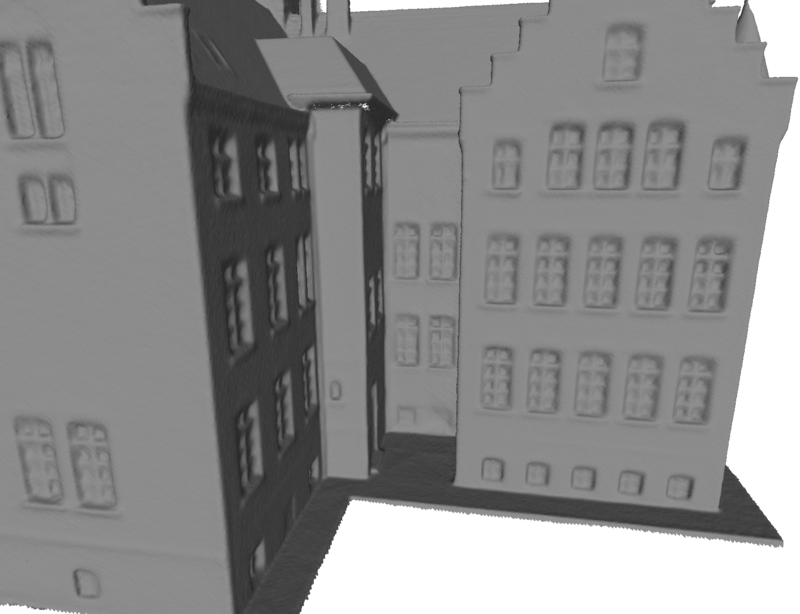} &
    \includegraphics[width=0.15\textwidth]{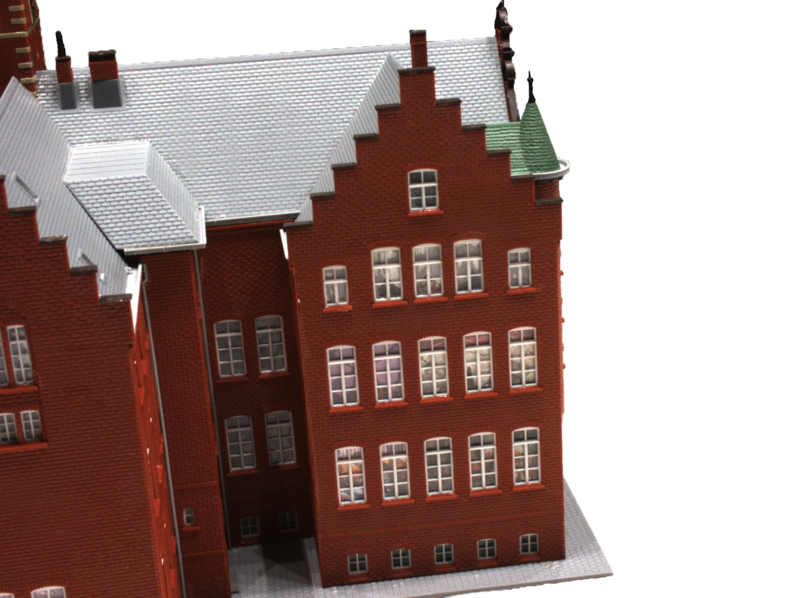} \\

    \includegraphics[width=0.15\textwidth]{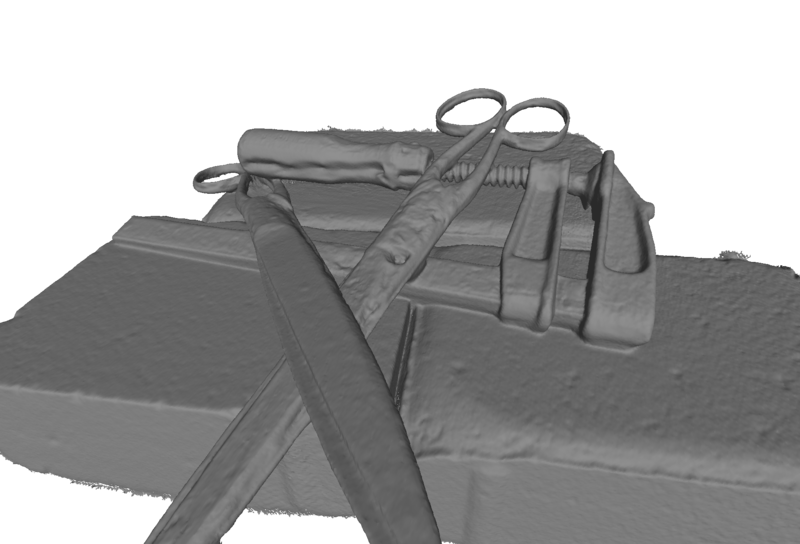} &
    \includegraphics[width=0.15\textwidth]{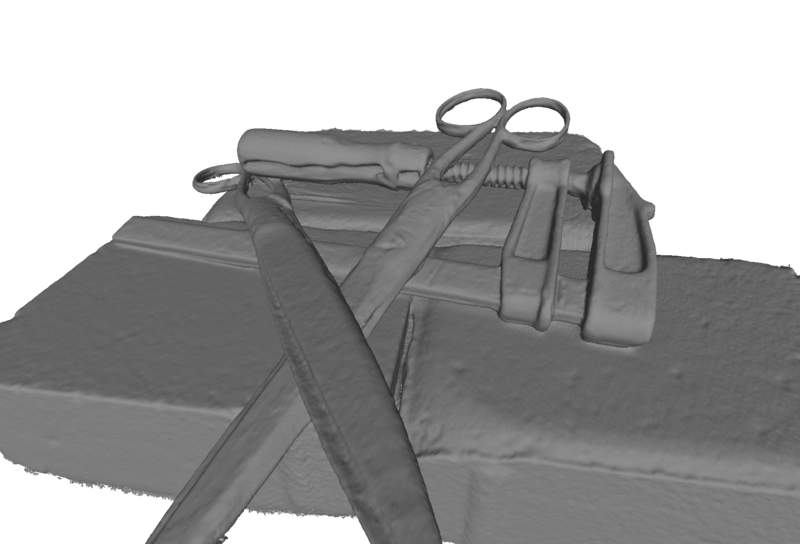} &
    
    \includegraphics[width=0.15\textwidth]{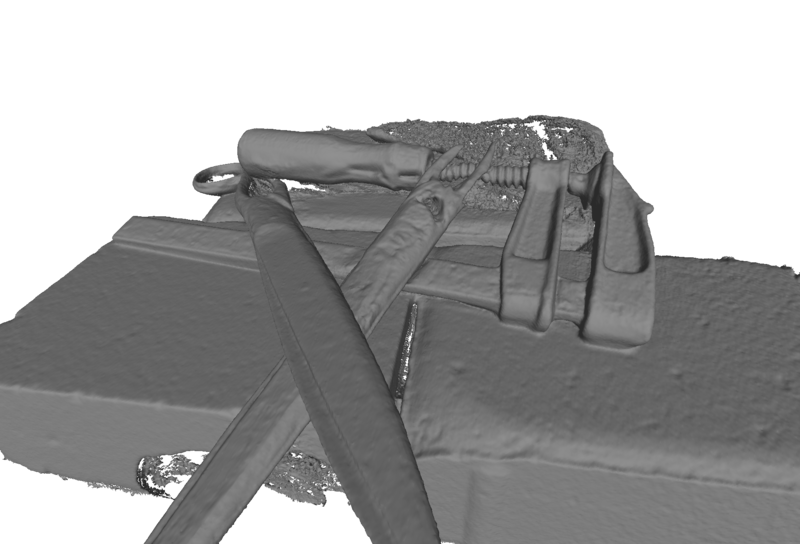} &
    
    \includegraphics[width=0.15\textwidth]{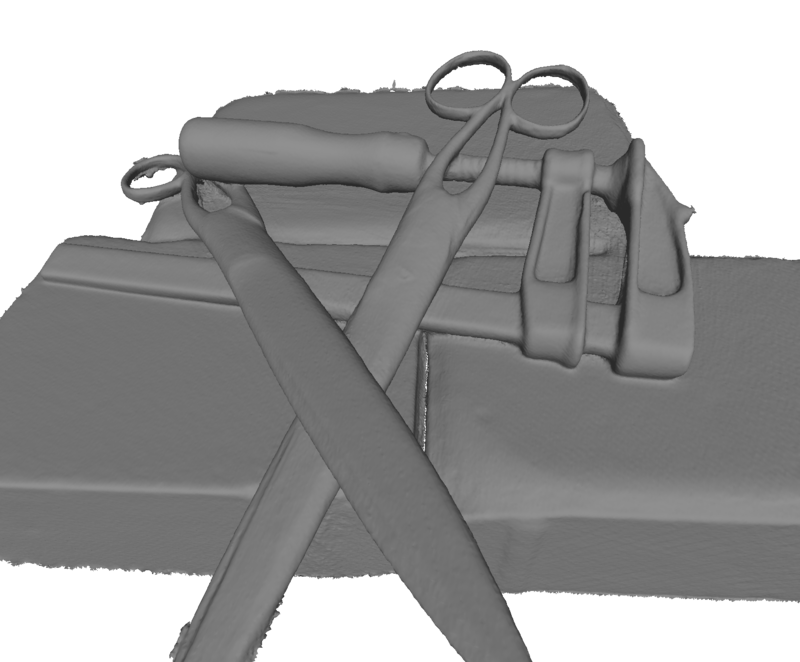} &
    \includegraphics[width=0.15\textwidth]{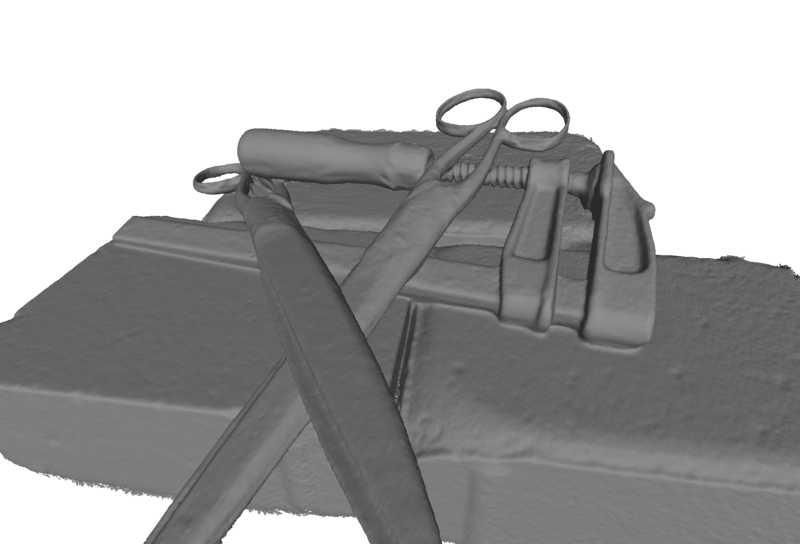} &
    \includegraphics[width=0.15\textwidth]{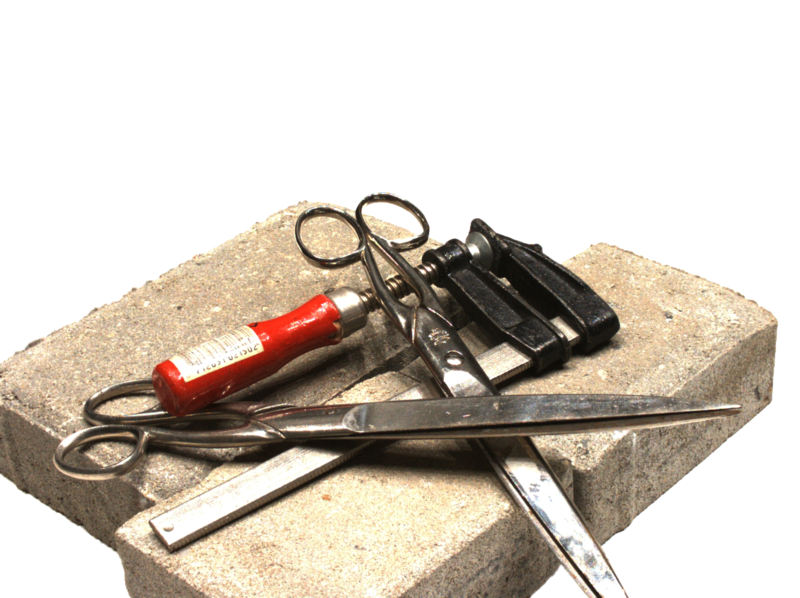} \\

    \includegraphics[width=0.15\textwidth]{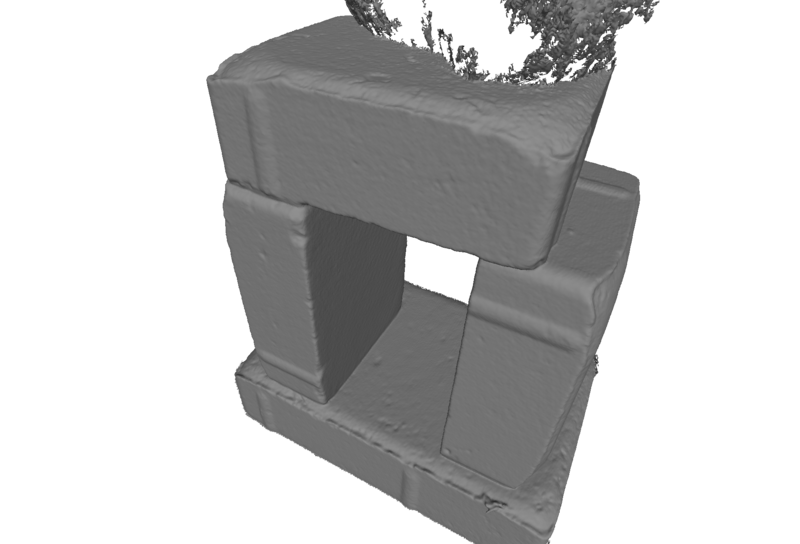} &
    \includegraphics[width=0.15\textwidth]{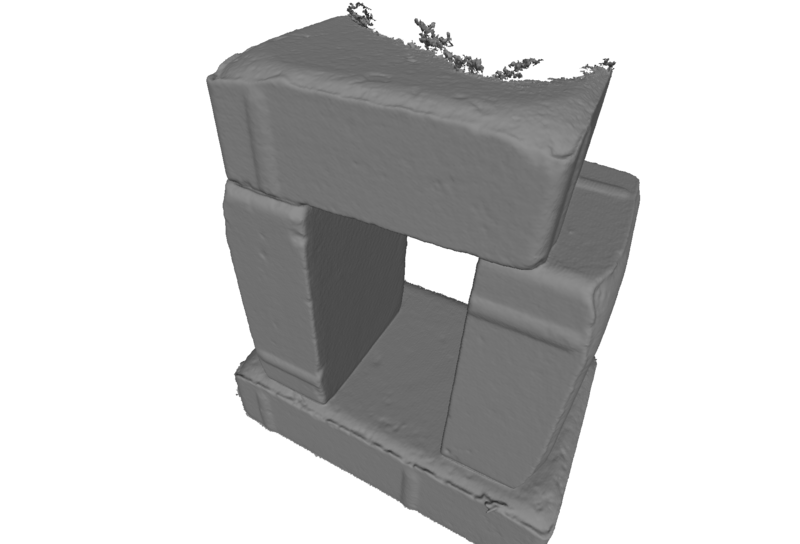} &
    
    \includegraphics[width=0.15\textwidth]{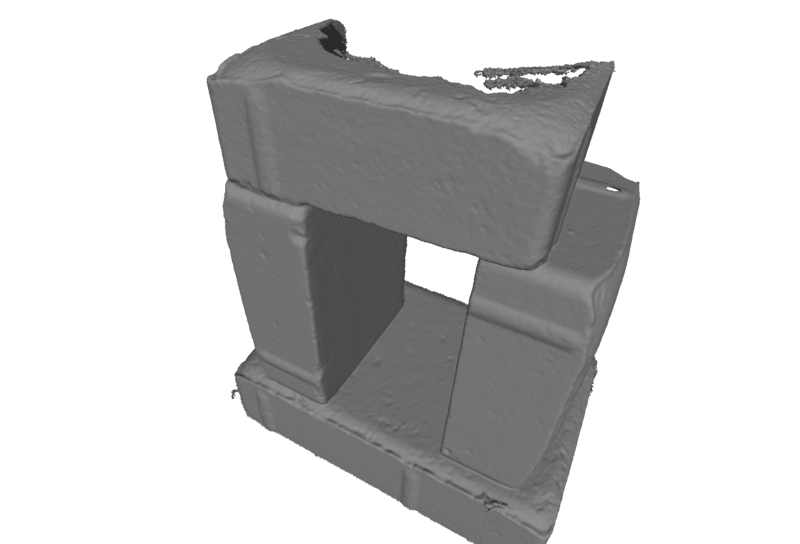} &
    
    \includegraphics[width=0.15\textwidth]{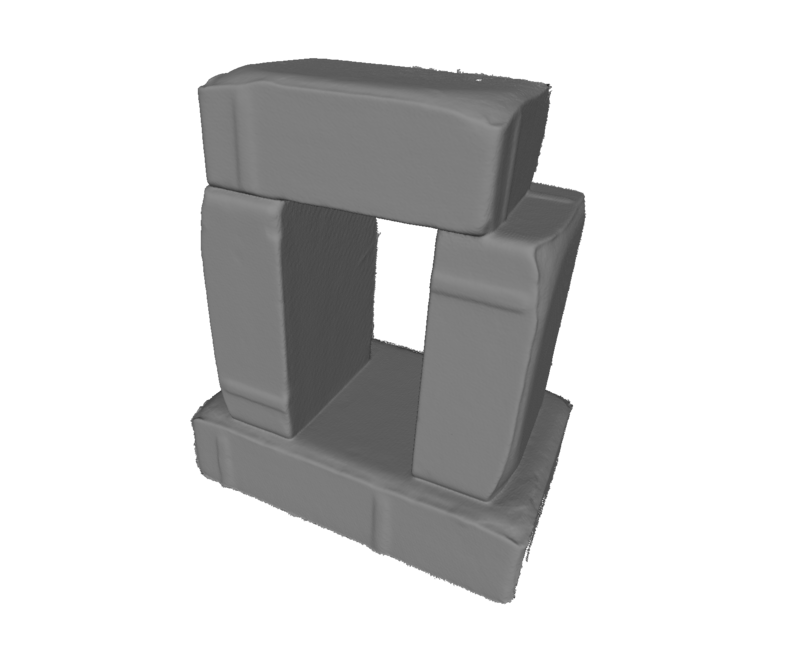} &
    \includegraphics[width=0.15\textwidth]{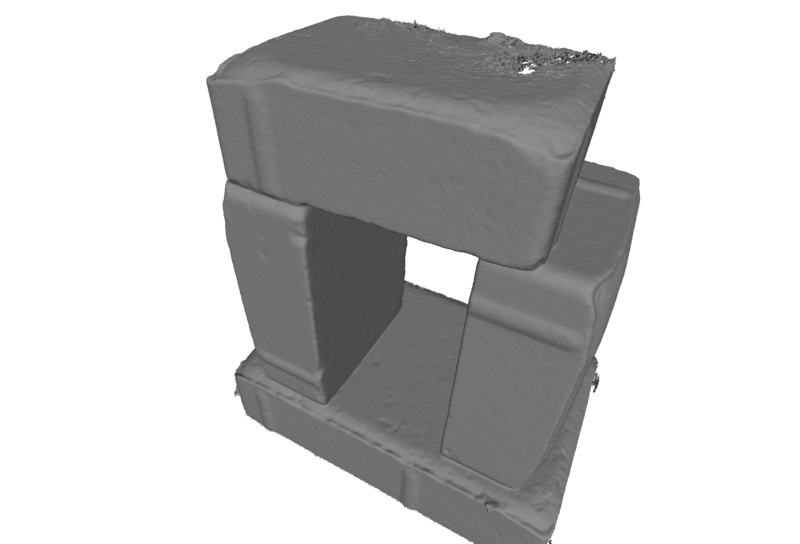} &
    \includegraphics[width=0.15\textwidth]{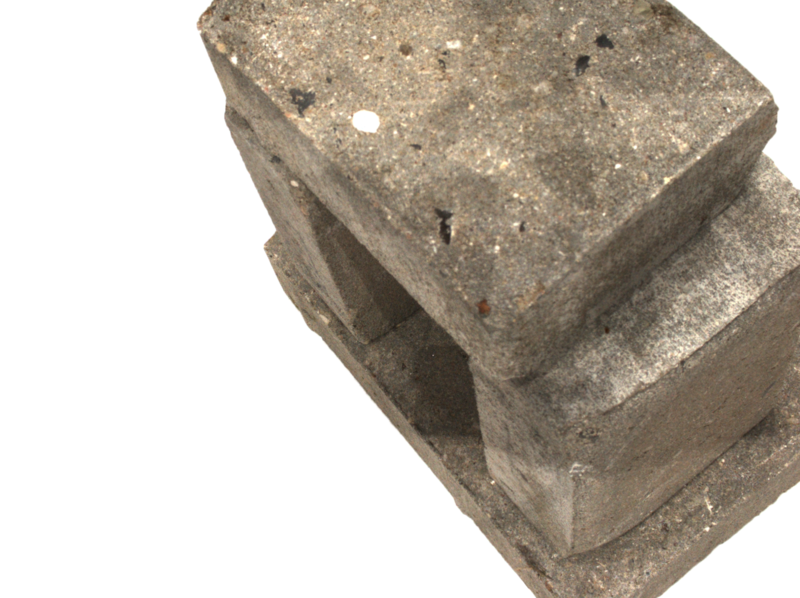} \\

    \includegraphics[width=0.15\textwidth]{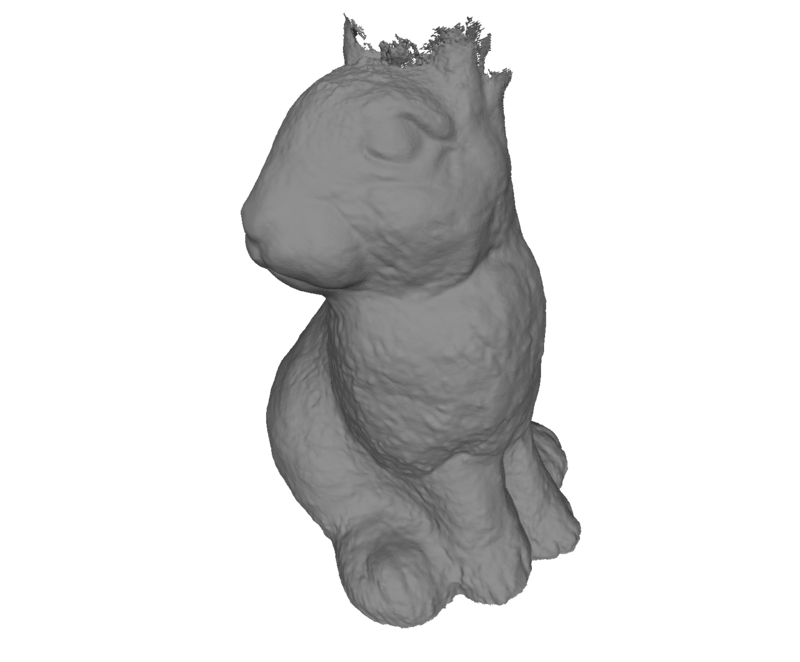} &
    \includegraphics[width=0.15\textwidth]{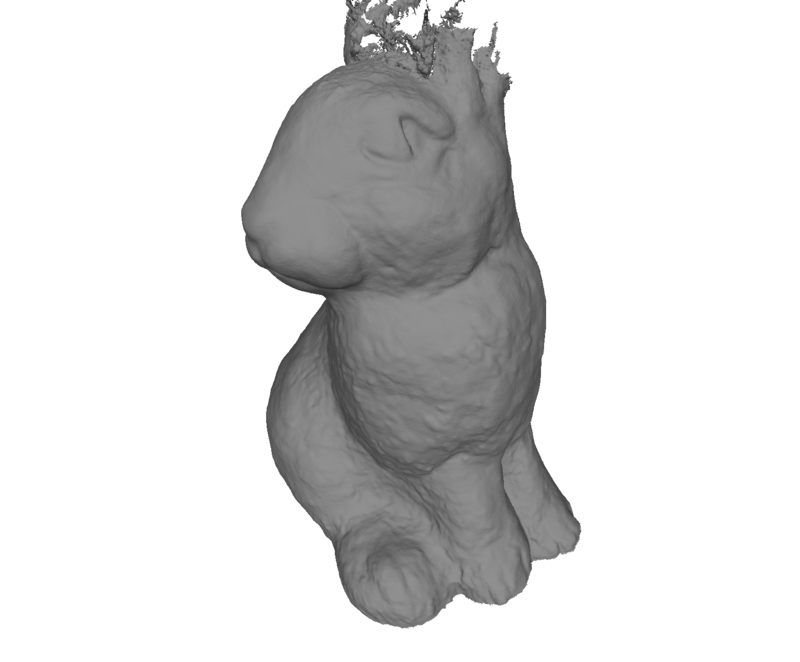} &
    
    \includegraphics[width=0.15\textwidth]{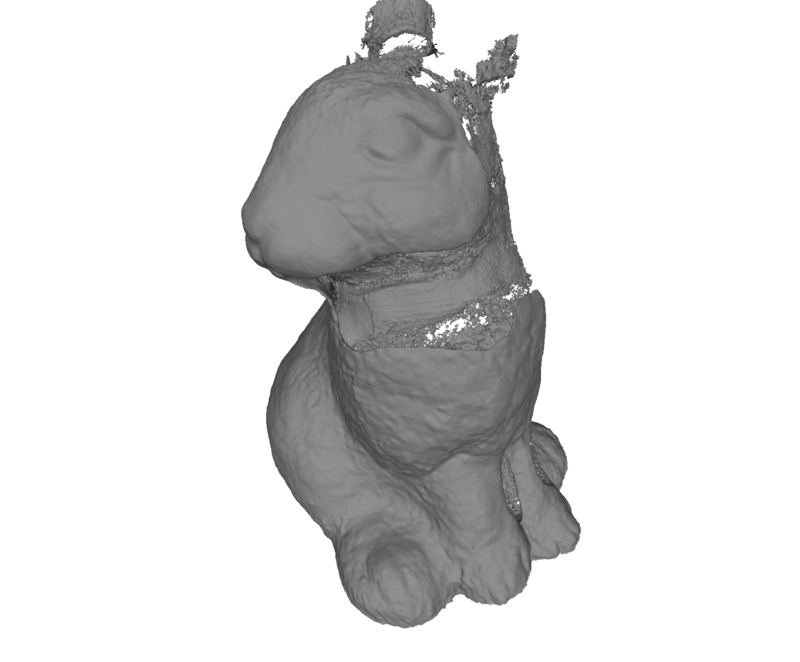} &
    
    \includegraphics[width=0.15\textwidth]{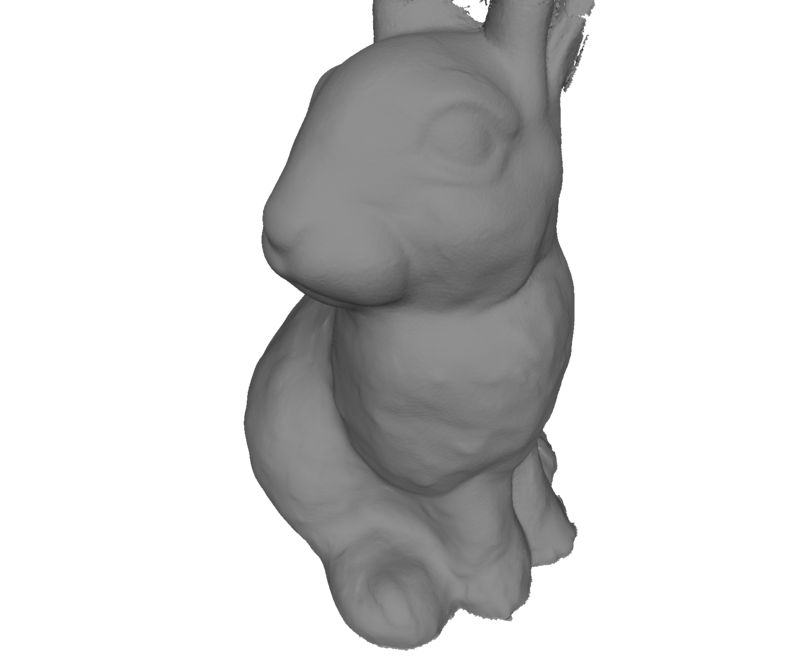} &
    \includegraphics[width=0.15\textwidth]{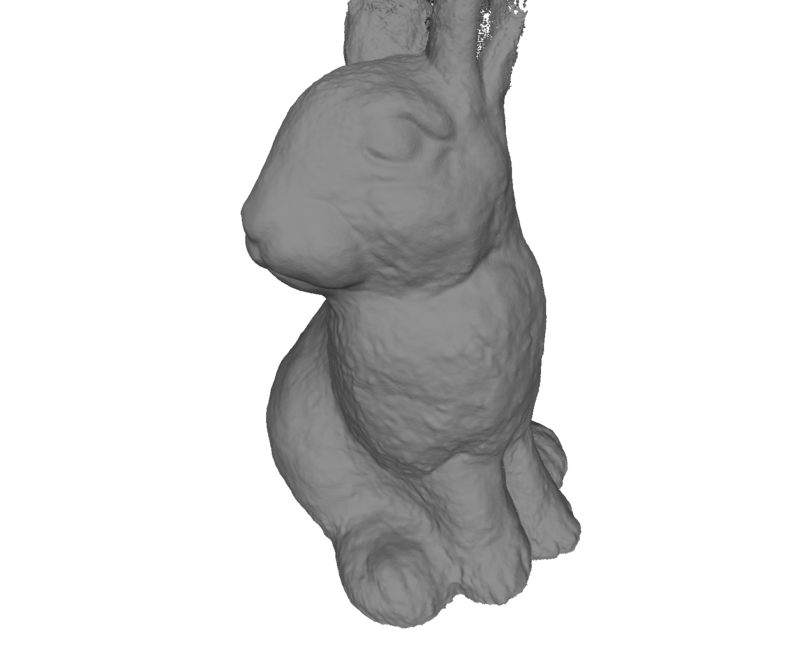} &
    \includegraphics[width=0.15\textwidth]{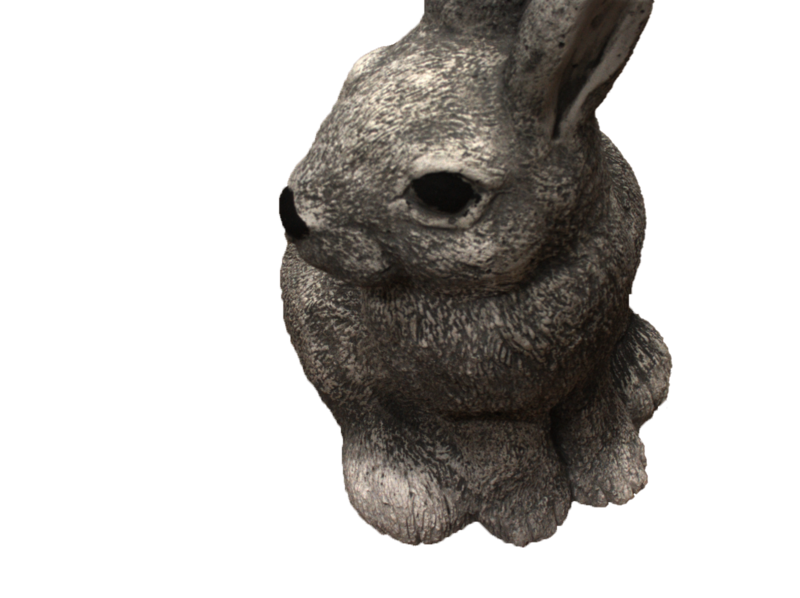} \\

    \includegraphics[width=0.15\textwidth]{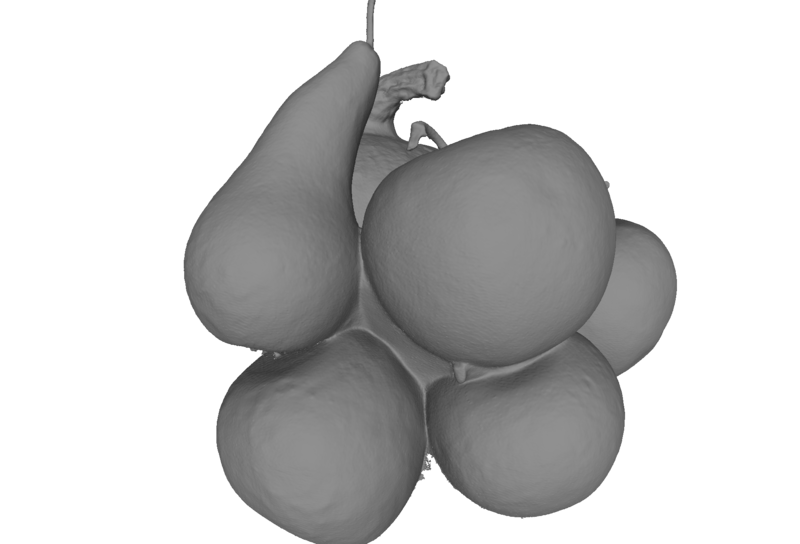} &
    \includegraphics[width=0.15\textwidth]{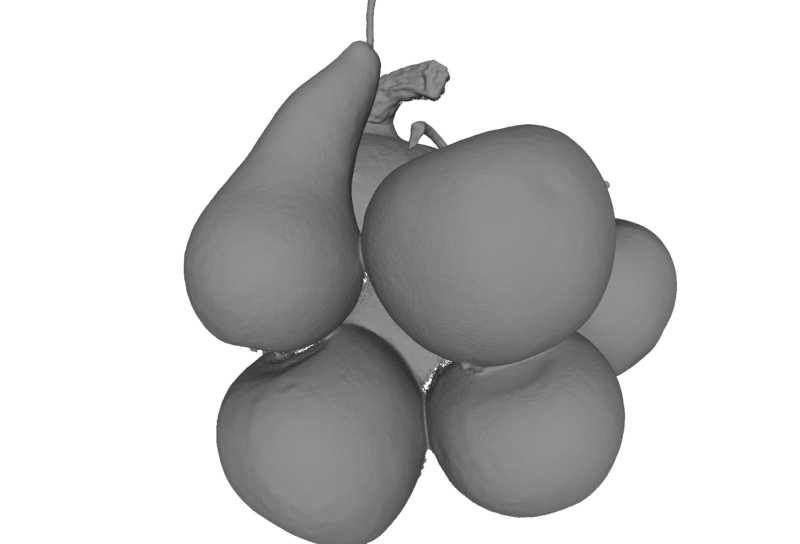} &
    
    \includegraphics[width=0.15\textwidth]{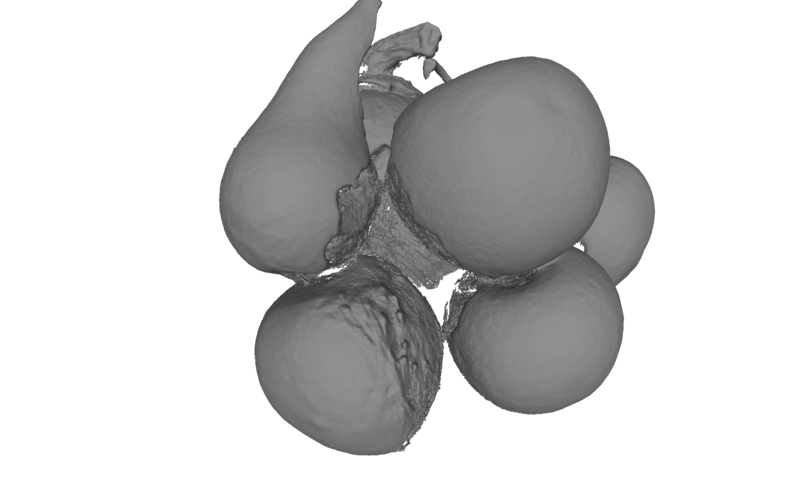} &
    
    \includegraphics[width=0.15\textwidth]{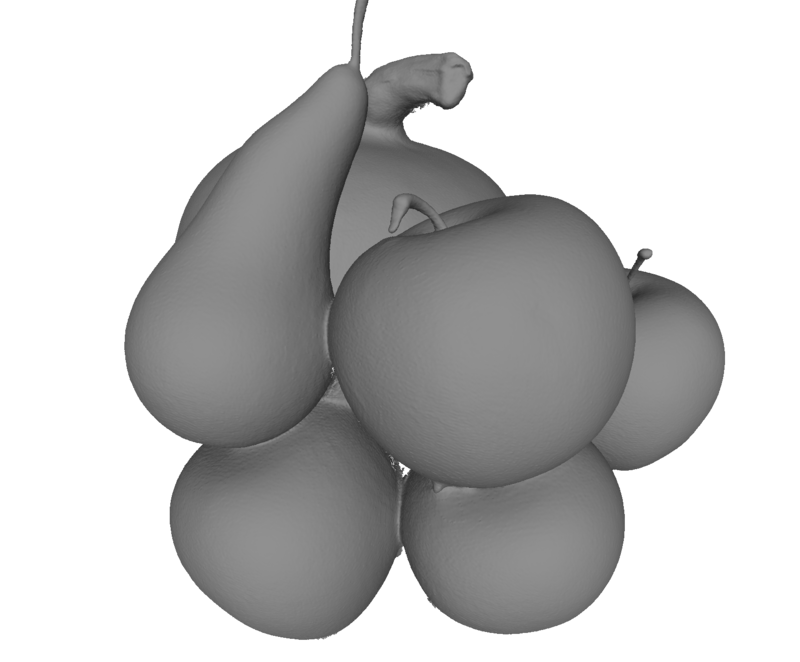} &
    \includegraphics[width=0.15\textwidth]{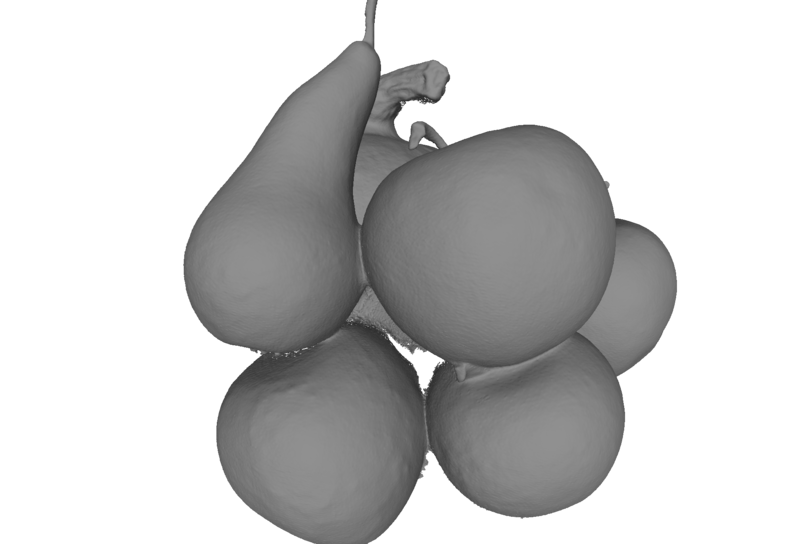} &
    \includegraphics[width=0.15\textwidth]{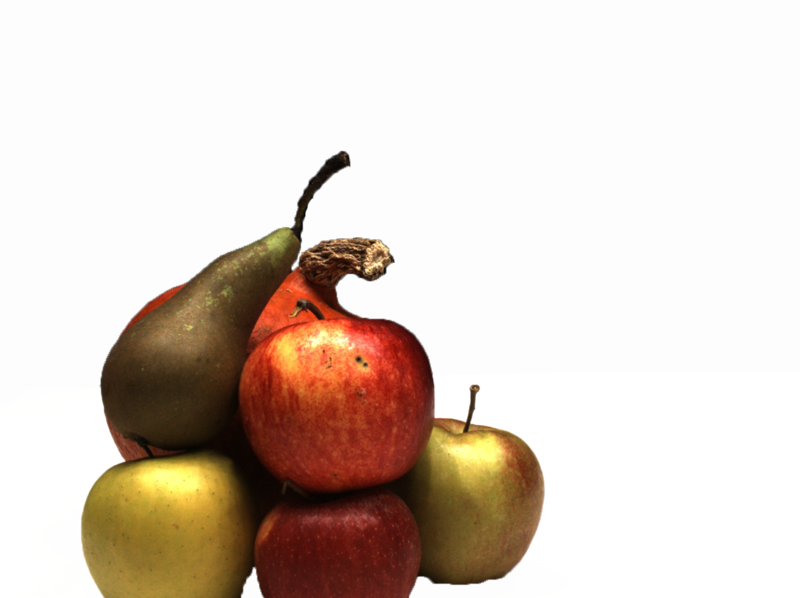} \\

    \includegraphics[width=0.15\textwidth]{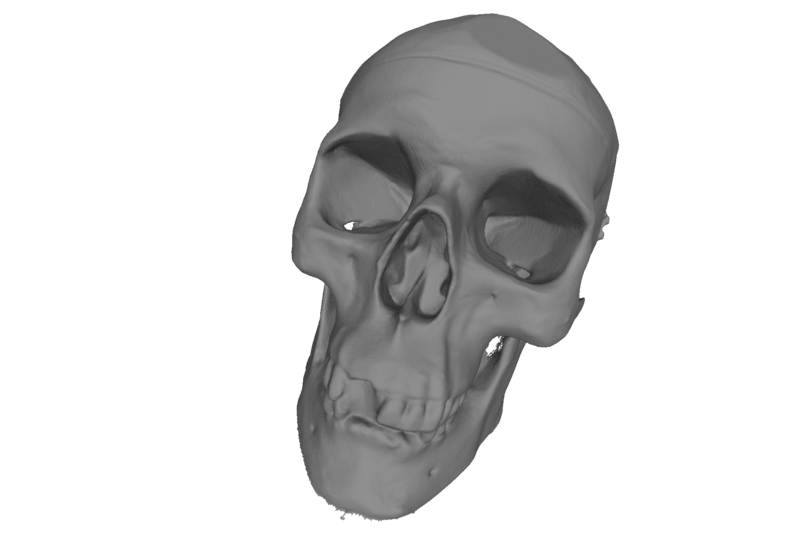} &
    \includegraphics[width=0.15\textwidth]{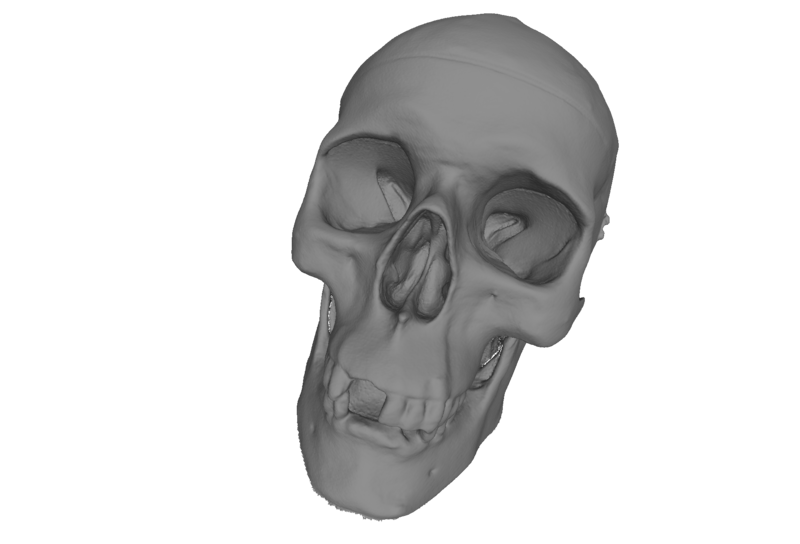} &
    
    \includegraphics[width=0.15\textwidth]{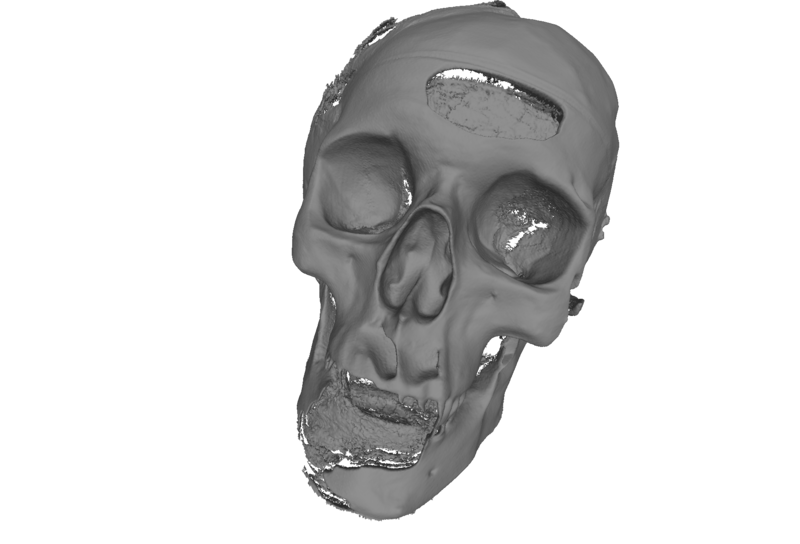} &
    
    \includegraphics[width=0.15\textwidth]{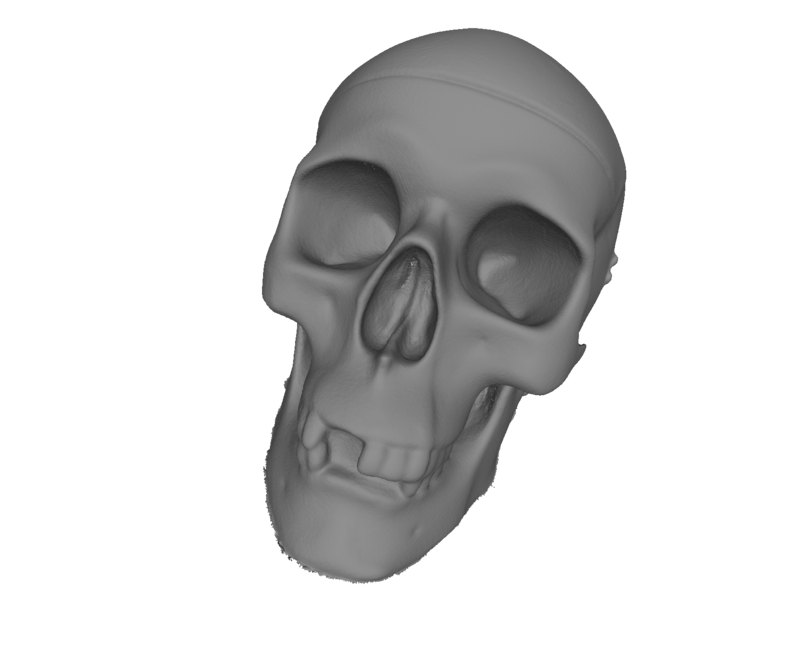} &
    \includegraphics[width=0.15\textwidth]{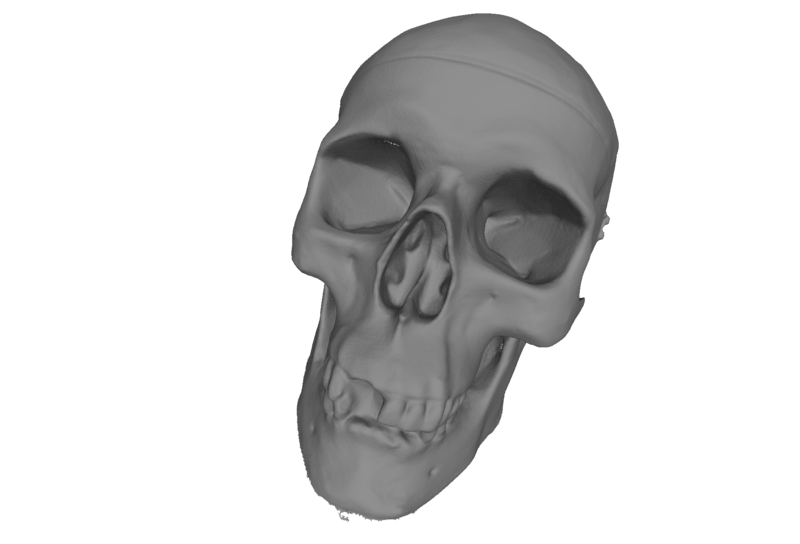} &
    \includegraphics[width=0.15\textwidth]{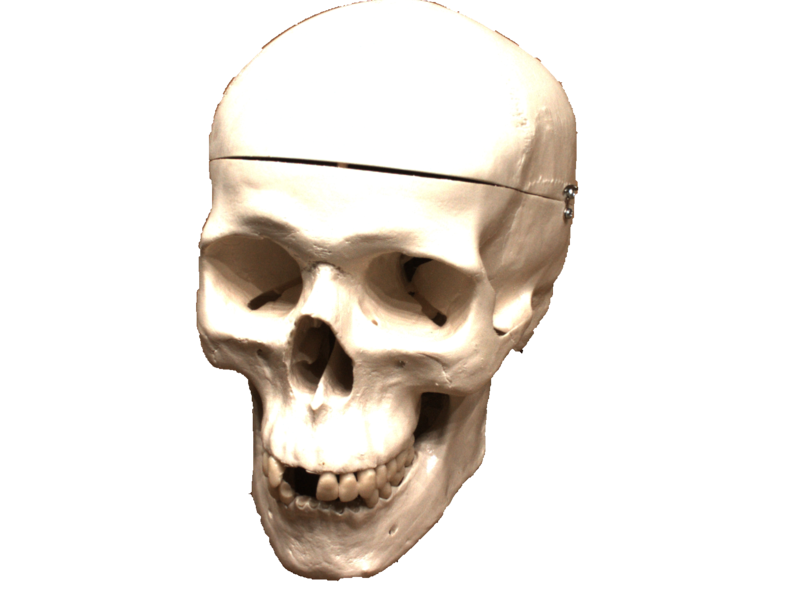} \\

    \includegraphics[width=0.15\textwidth]{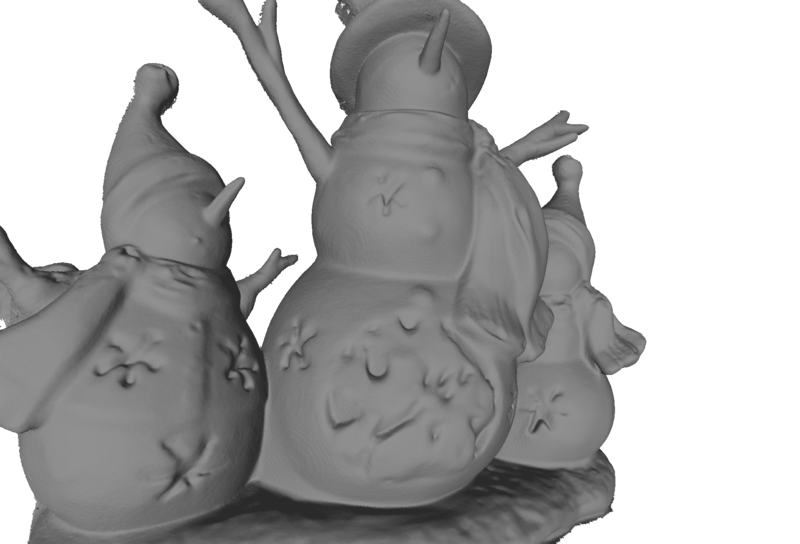} &
    \includegraphics[width=0.15\textwidth]{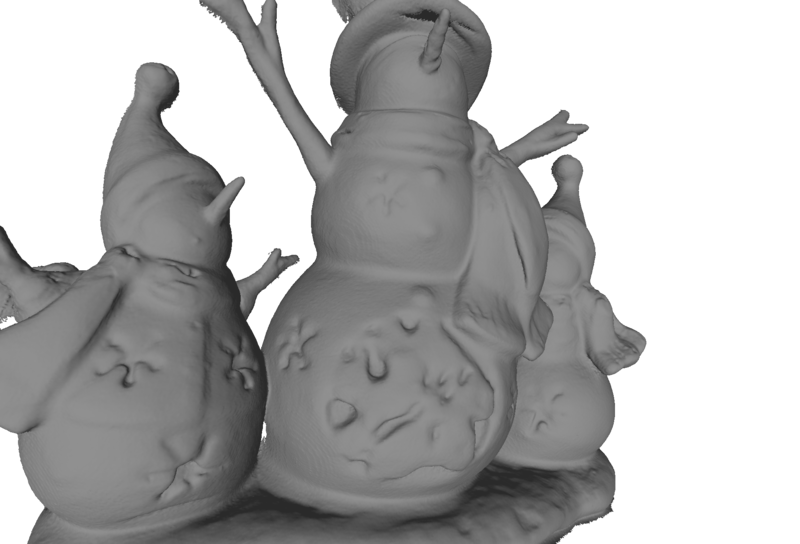} &
    
    \includegraphics[width=0.15\textwidth]{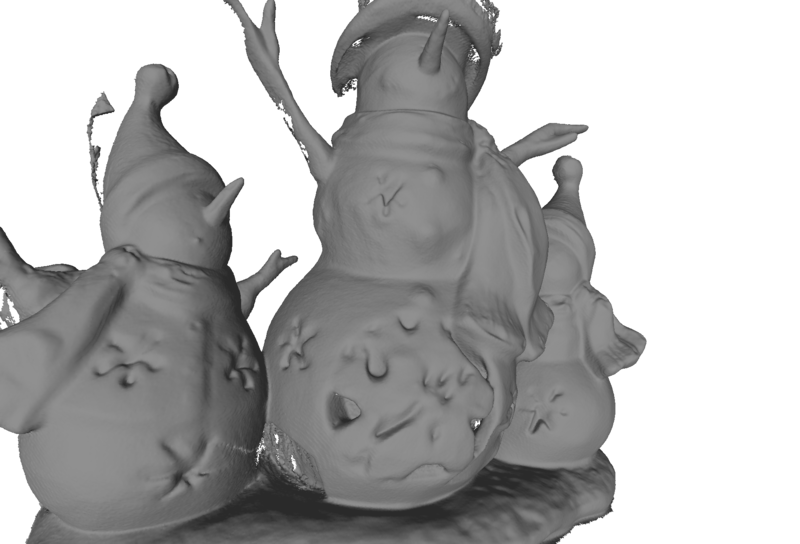} &
    
    \includegraphics[width=0.15\textwidth]{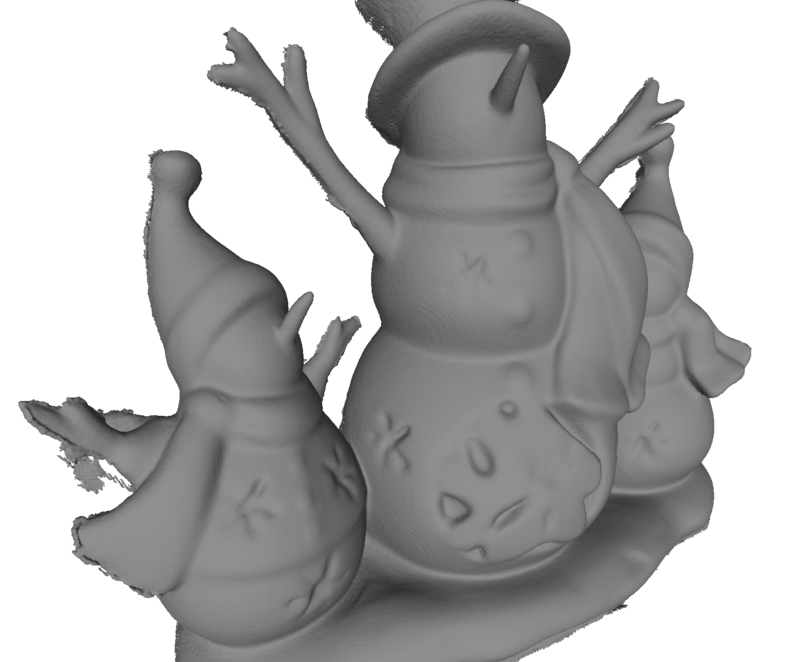} &
    \includegraphics[width=0.15\textwidth]{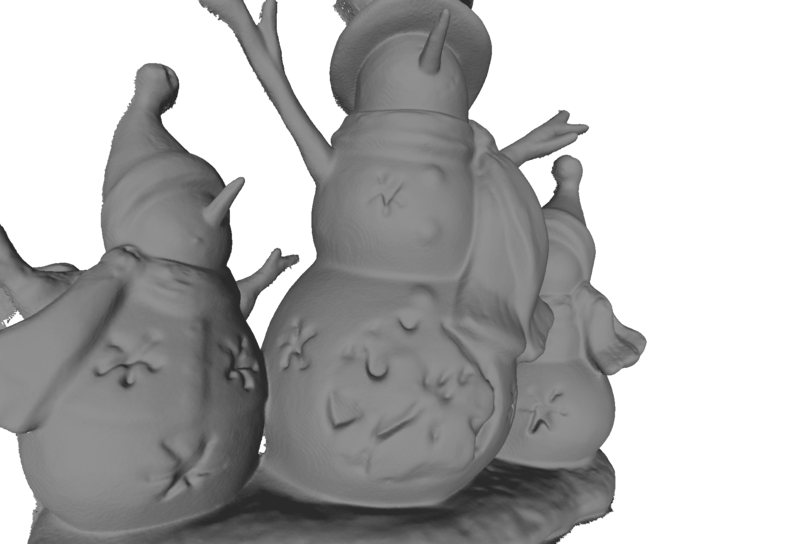} &
    \includegraphics[width=0.15\textwidth]{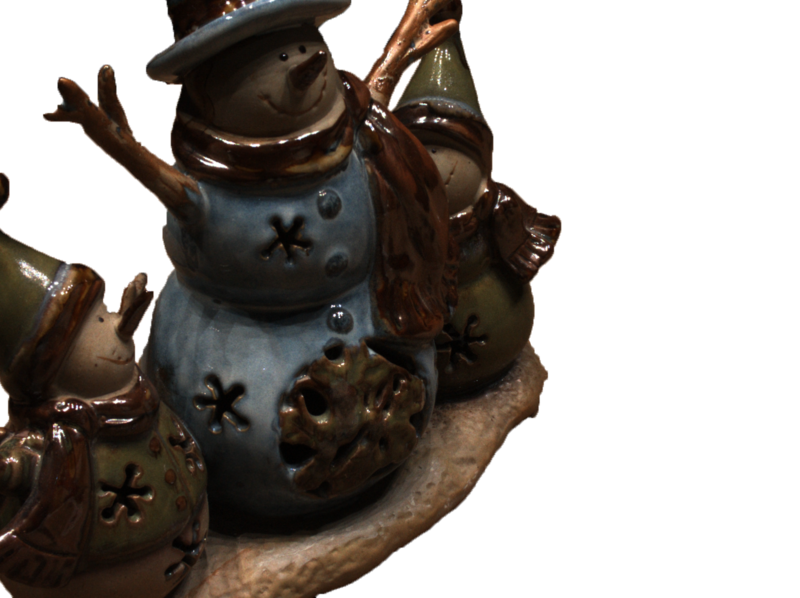} \\

    \includegraphics[width=0.15\textwidth]{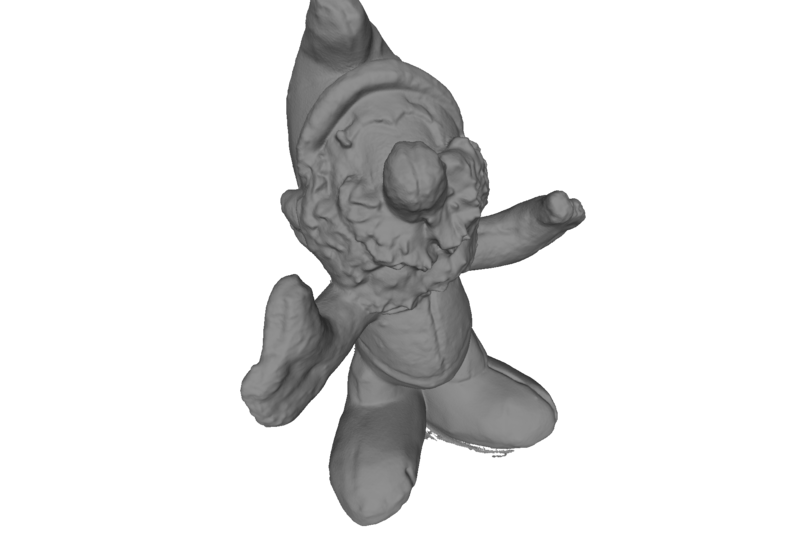} &
    \includegraphics[width=0.15\textwidth]{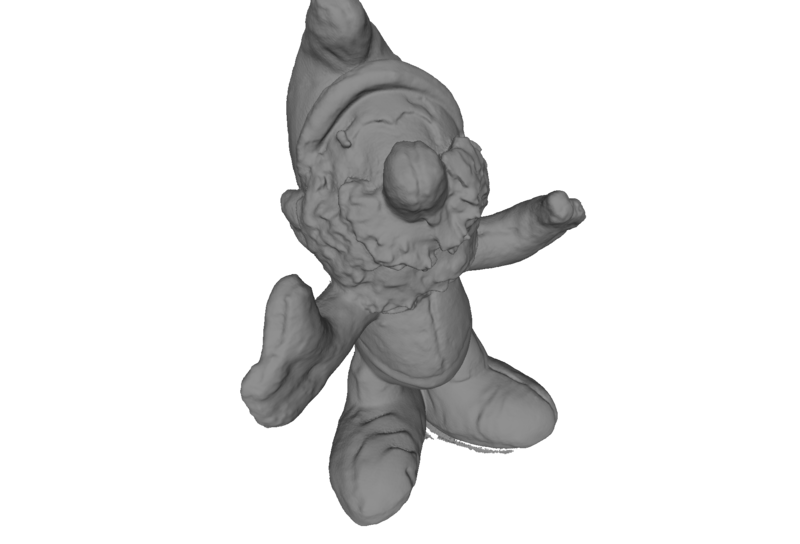} &
    \includegraphics[width=0.15\textwidth]{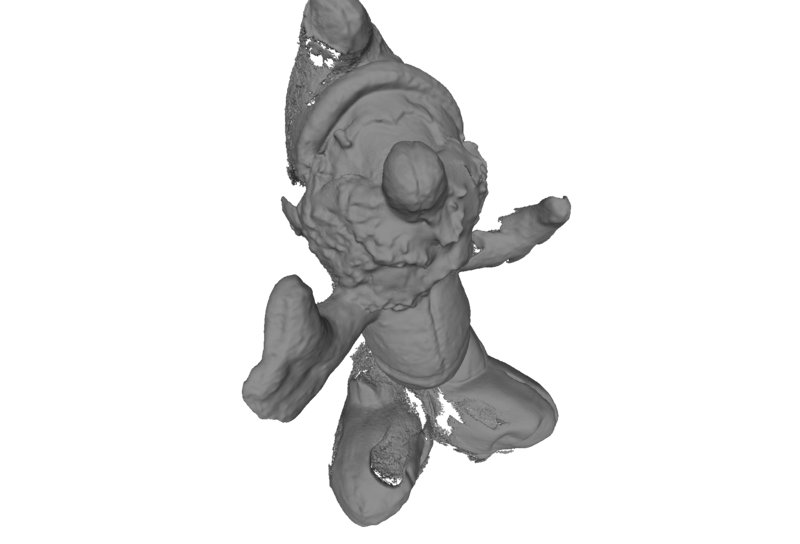} &
    \includegraphics[width=0.15\textwidth]{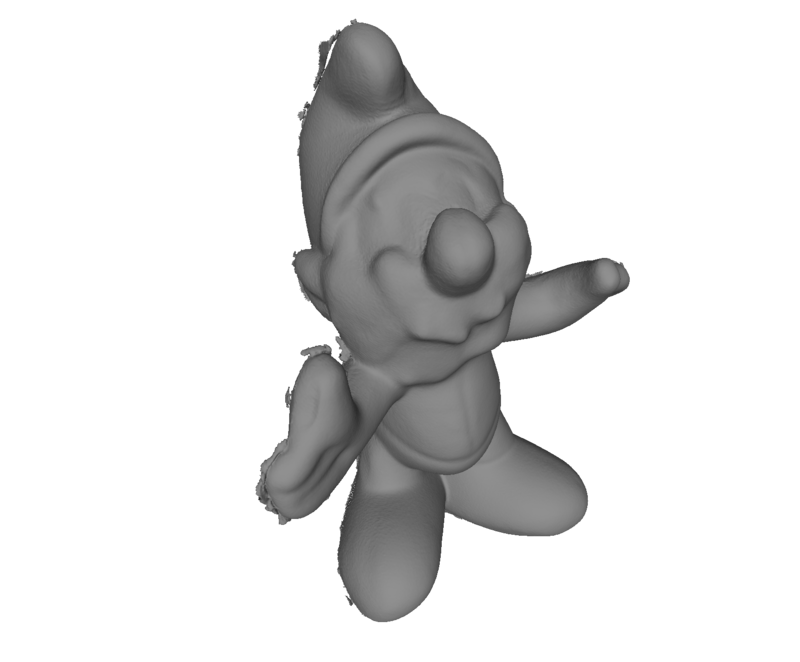} &
    \includegraphics[width=0.15\textwidth]{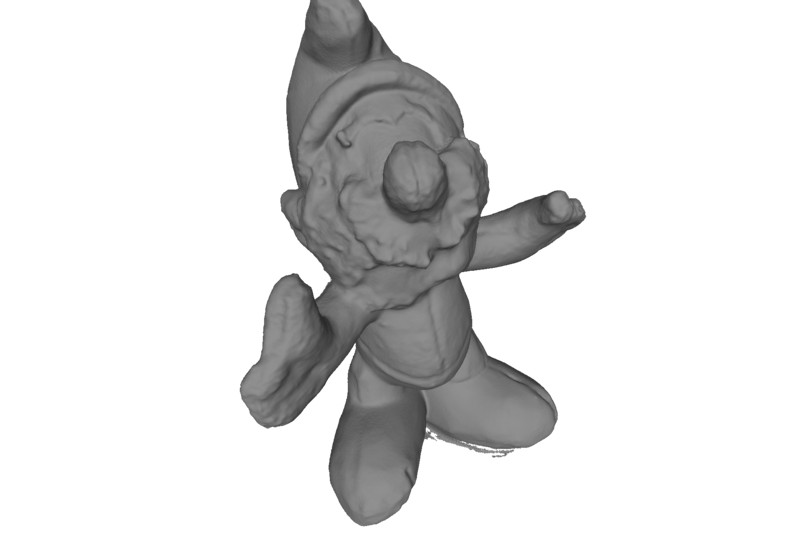} &
    \includegraphics[width=0.15\textwidth]{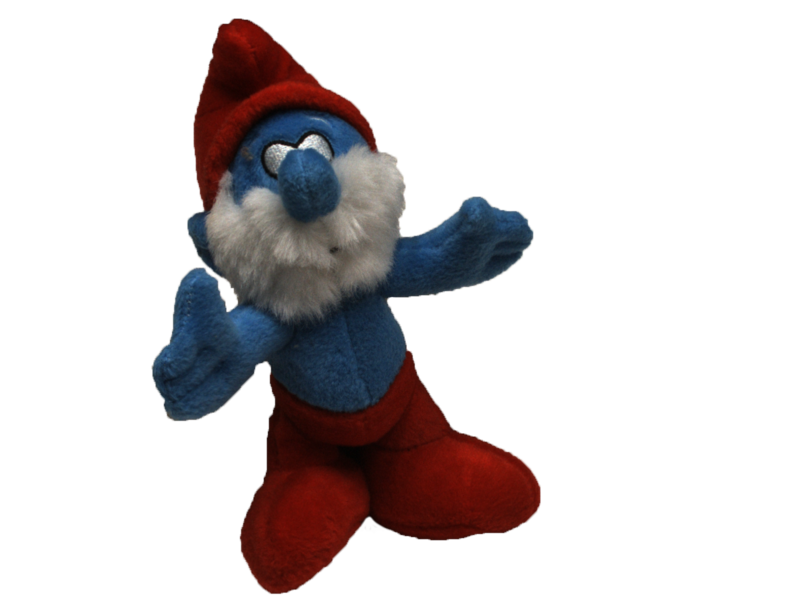} \\

    \includegraphics[width=0.15\textwidth]{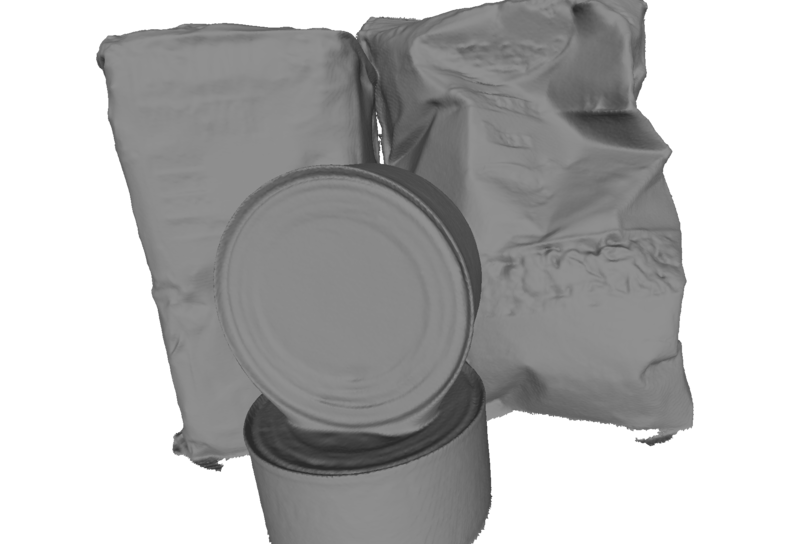} &
    \includegraphics[width=0.15\textwidth]{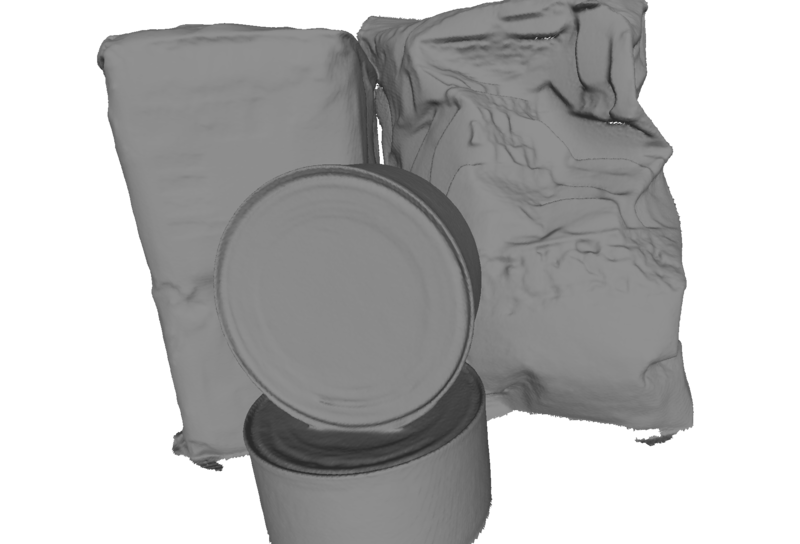} &
    \includegraphics[width=0.15\textwidth]{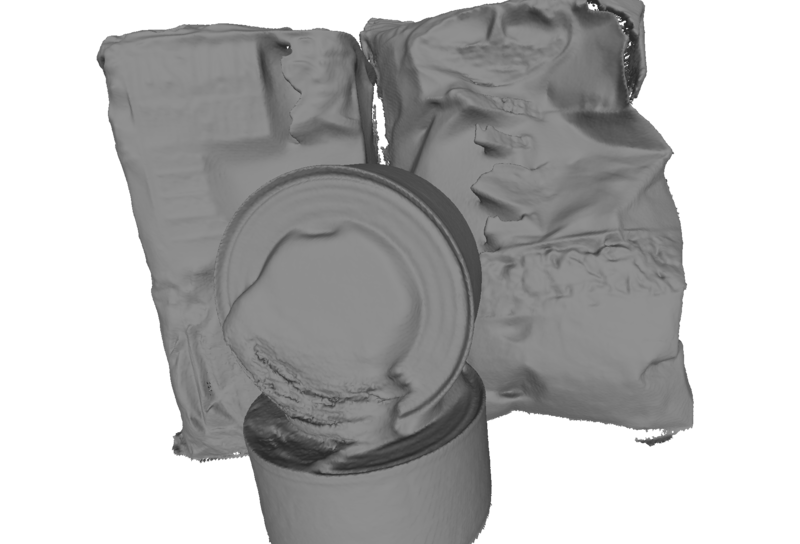} &
    \includegraphics[width=0.15\textwidth]{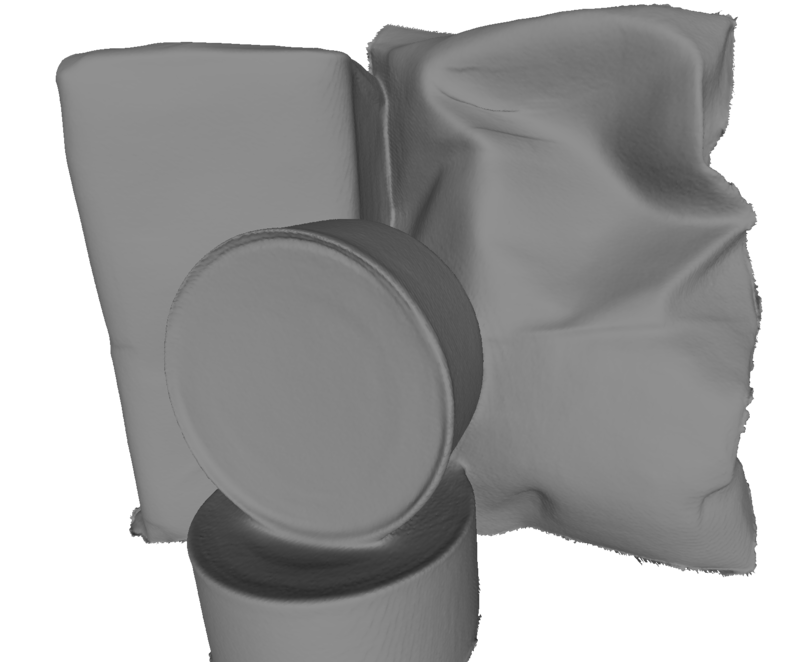} &
    \includegraphics[width=0.15\textwidth]{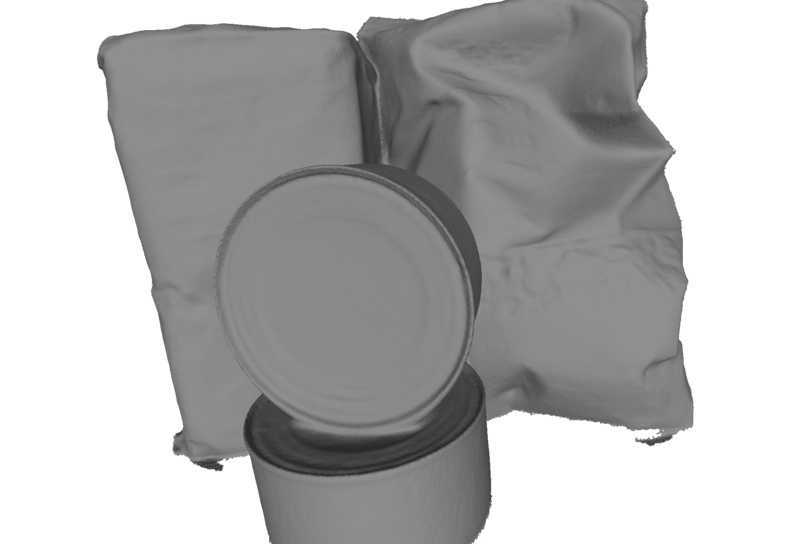} &
    \includegraphics[width=0.15\textwidth]{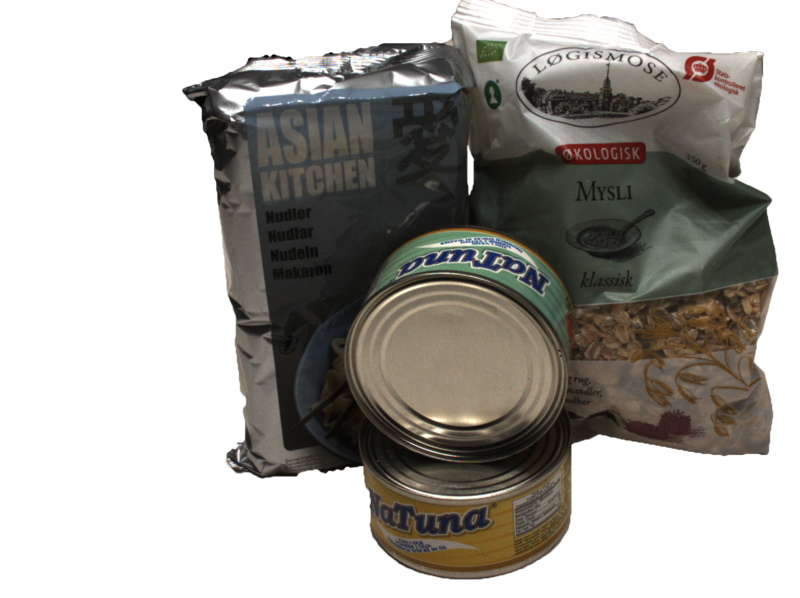} \\

    \includegraphics[width=0.15\textwidth]{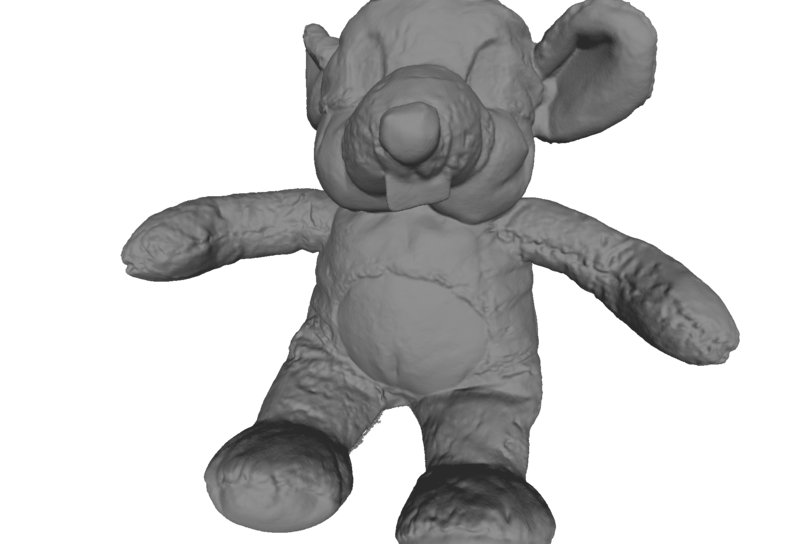} &
    \includegraphics[width=0.15\textwidth]{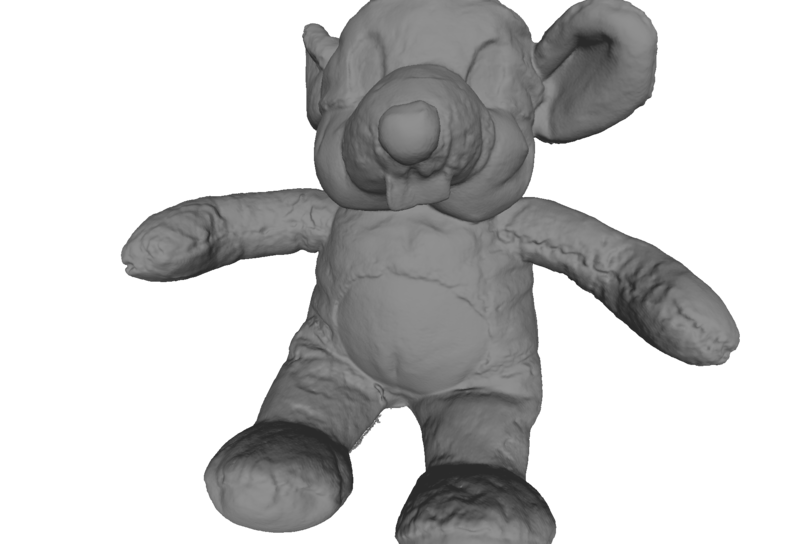} &
    \includegraphics[width=0.15\textwidth]{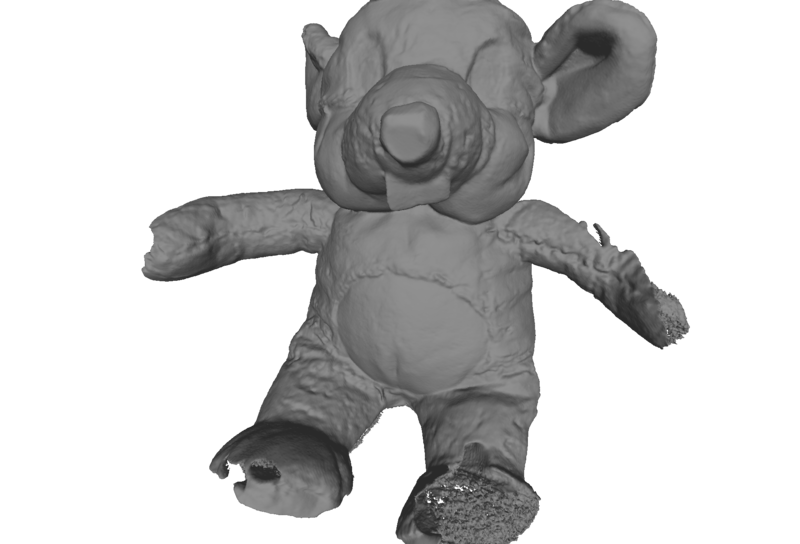} &
    \includegraphics[width=0.15\textwidth]{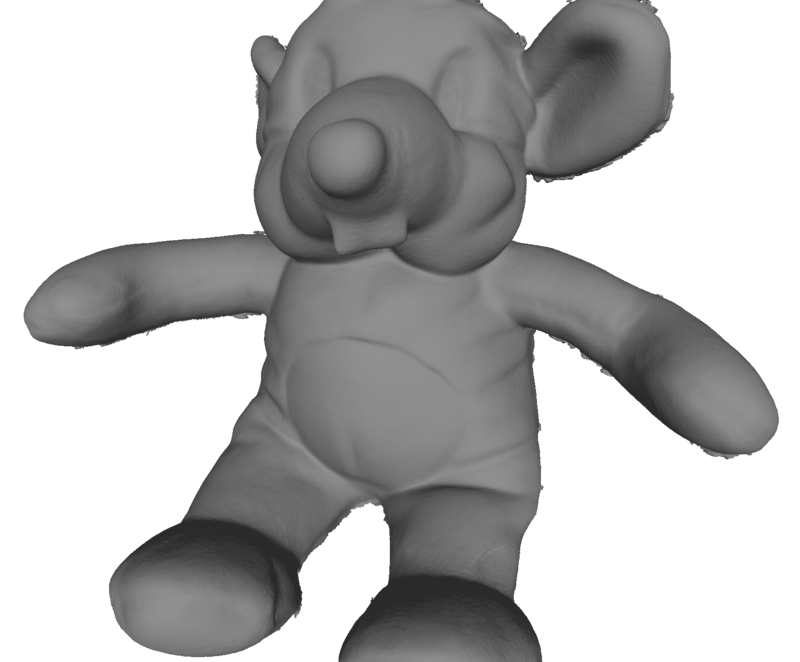} &
    \includegraphics[width=0.15\textwidth]{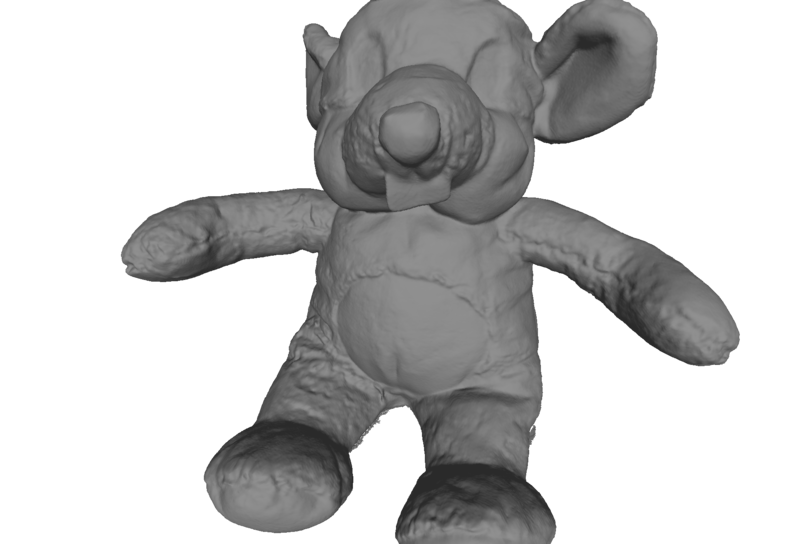} &
    \includegraphics[width=0.15\textwidth]{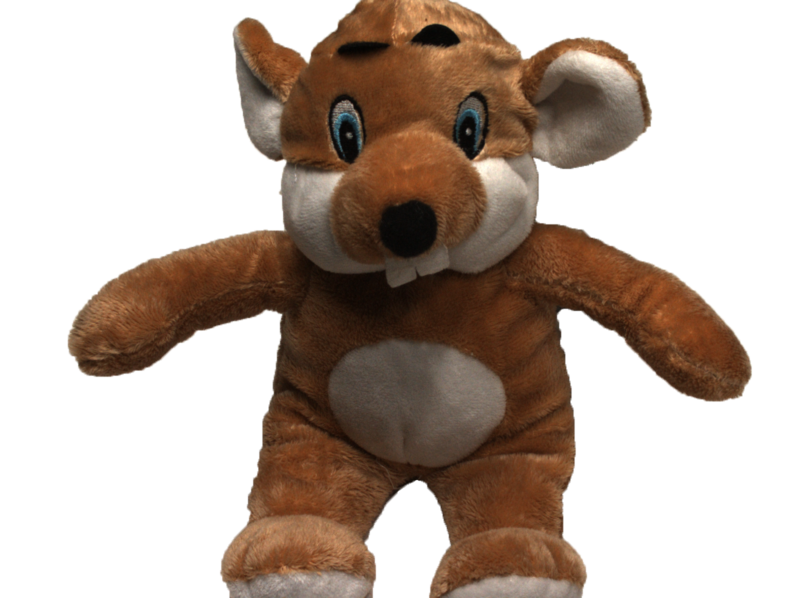} \\

    \includegraphics[width=0.15\textwidth]{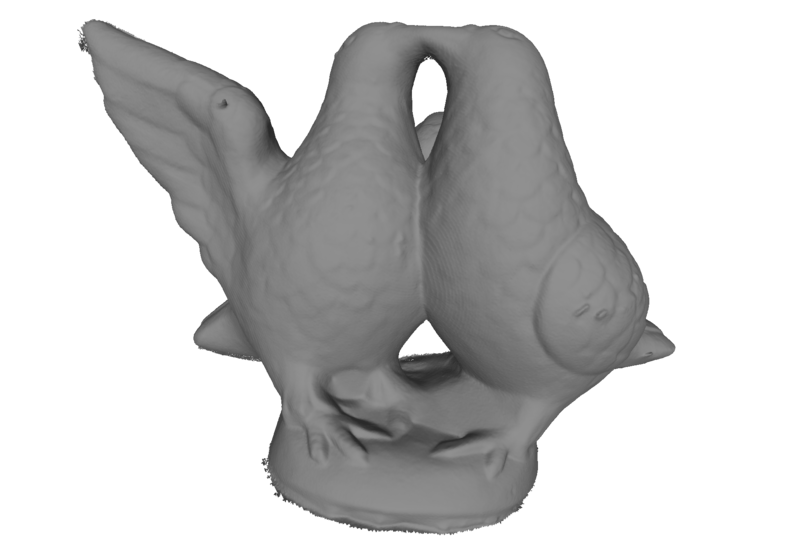} &
    \includegraphics[width=0.15\textwidth]{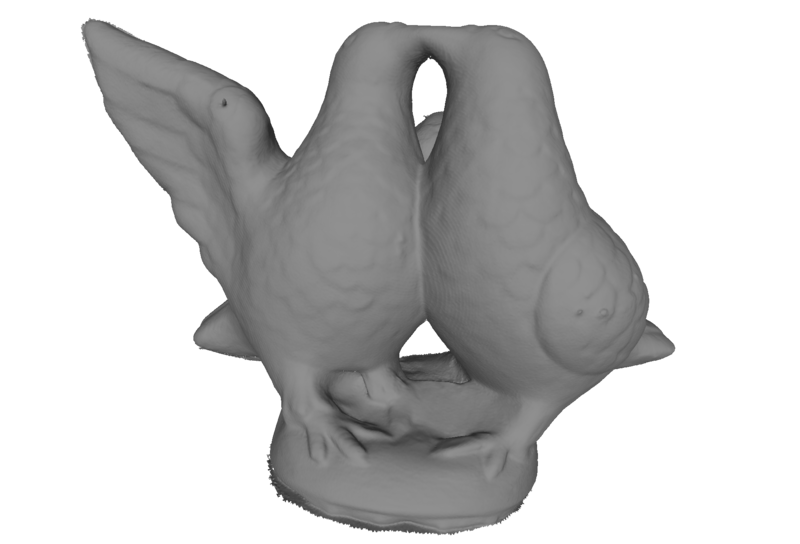} &
    \includegraphics[width=0.15\textwidth]{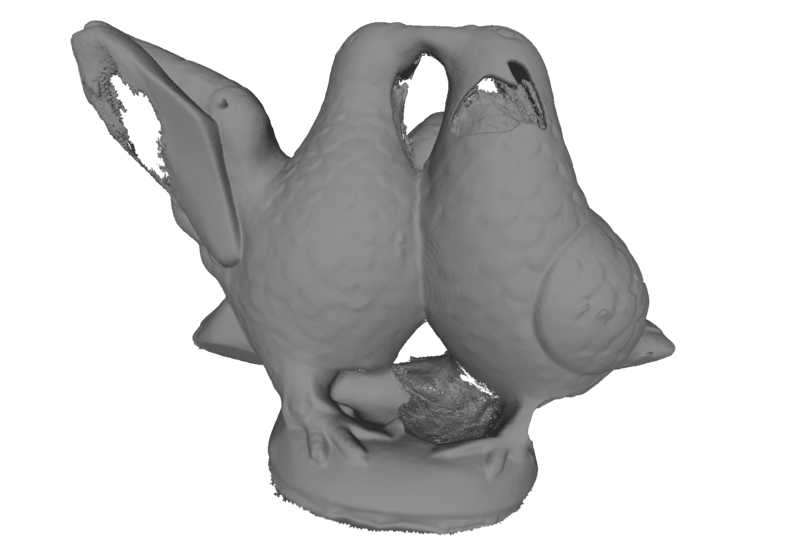} &
    \includegraphics[width=0.15\textwidth]{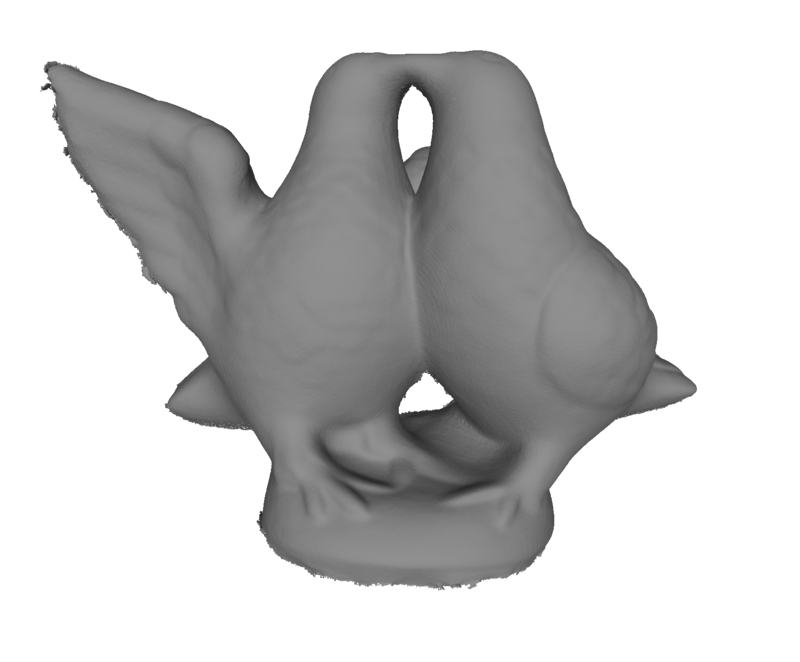} &
    \includegraphics[width=0.15\textwidth]{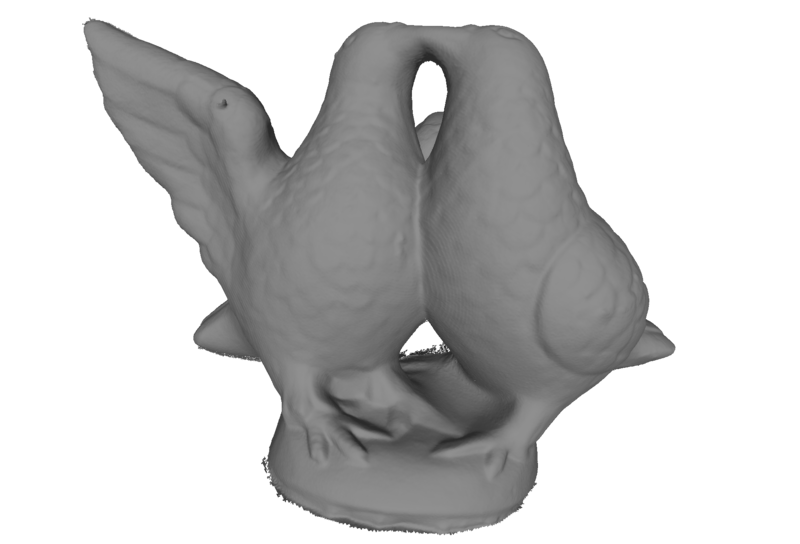} &
    \includegraphics[width=0.15\textwidth]{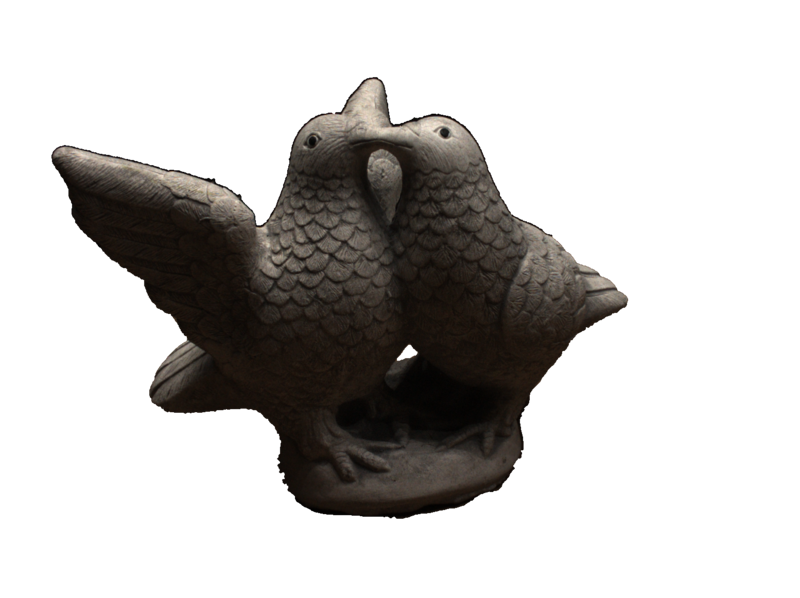} \\

    \includegraphics[width=0.15\textwidth]{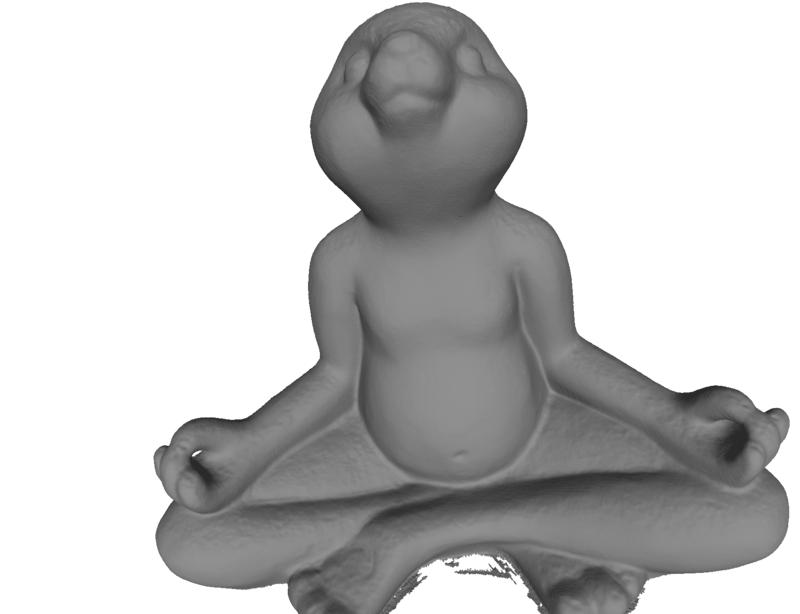} &
    \includegraphics[width=0.15\textwidth]{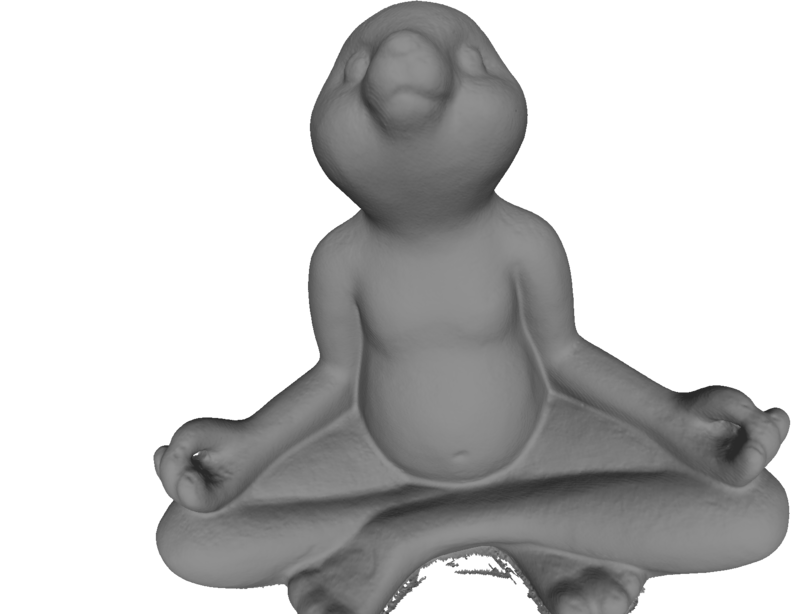} &
    \includegraphics[width=0.15\textwidth]{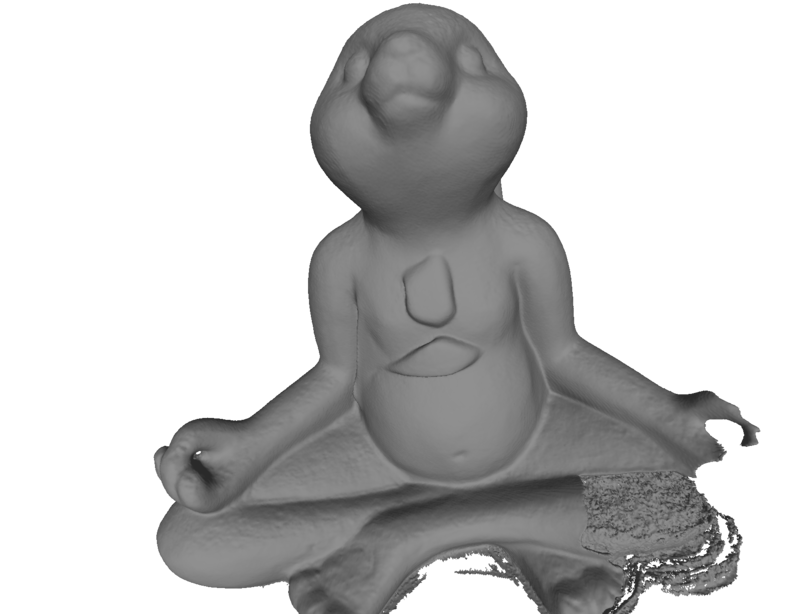} &
    \includegraphics[width=0.15\textwidth]{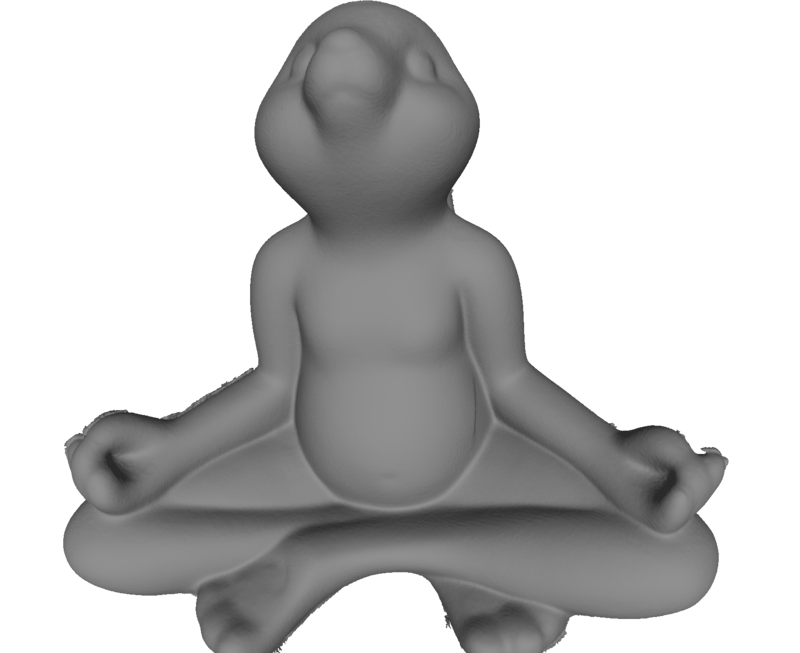} &
    \includegraphics[width=0.15\textwidth]{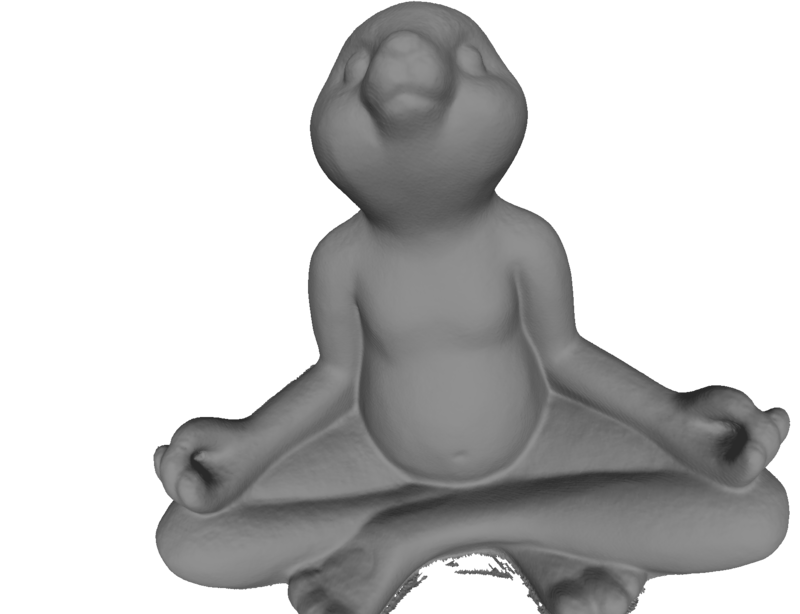} &
    \includegraphics[width=0.15\textwidth]{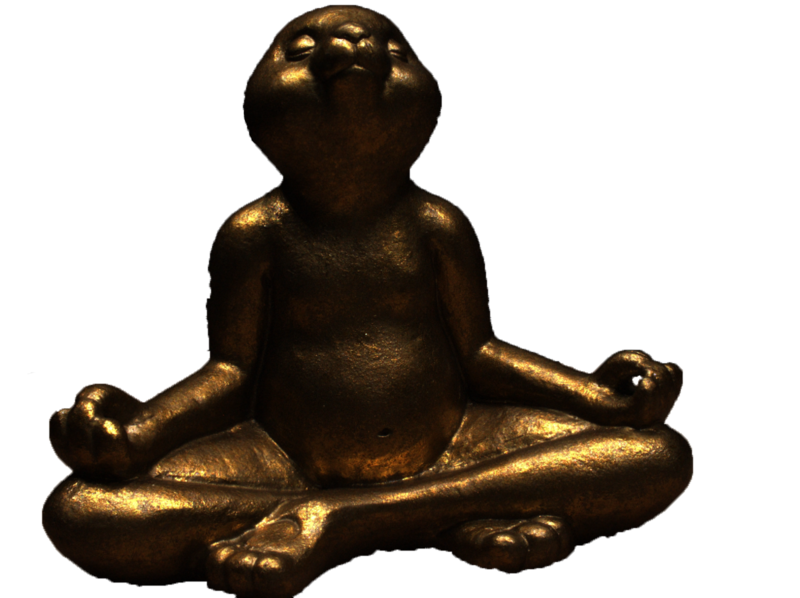} \\
    \end{tabular}
    \caption{Qualitative results on DTU from scan24 to scan110.}
    \label{supp:fig:DTU1}
\end{figure*}

\begin{figure*}
\setlength{\tabcolsep}{1pt} 
    \renewcommand{\arraystretch}{1} 
    \centering
    \begin{tabular}{cccccc}
    PGSR & +SN & +DA & \makecell[tc]{+VGGT\\(w/o conf)} & Ours & GT image\\
    \includegraphics[width=0.15\textwidth]{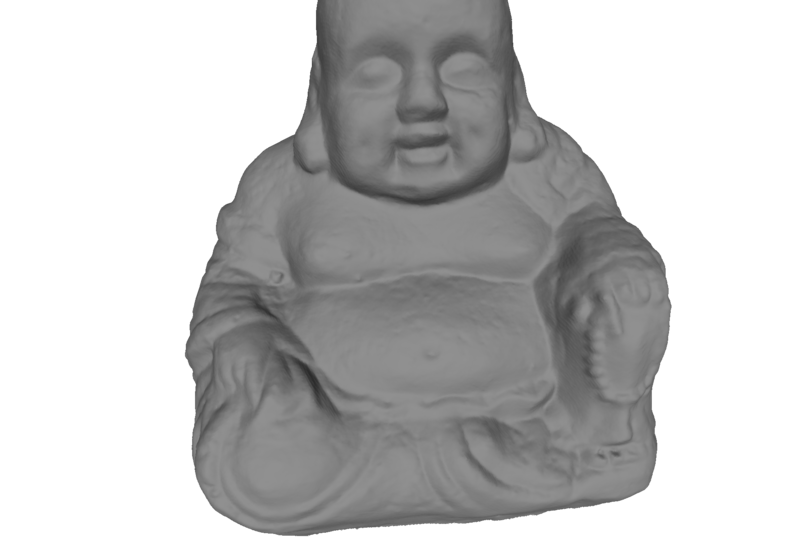} &
    \includegraphics[width=0.15\textwidth]{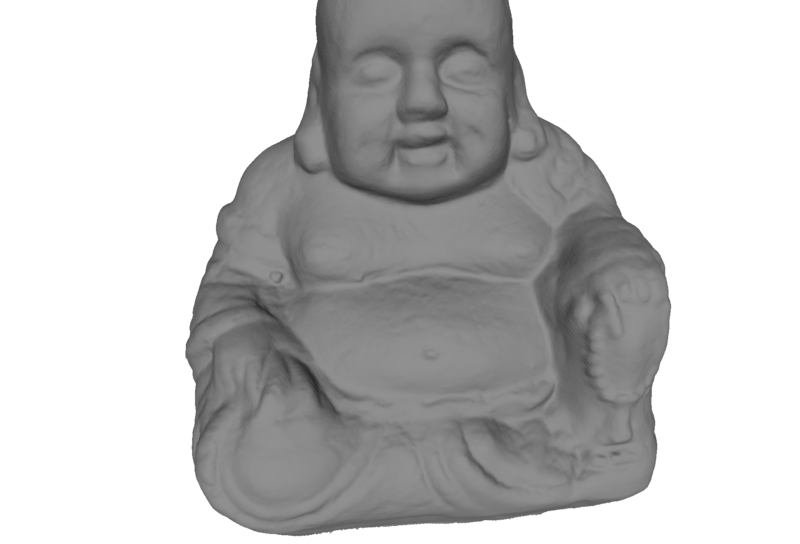} &
    \includegraphics[width=0.15\textwidth]{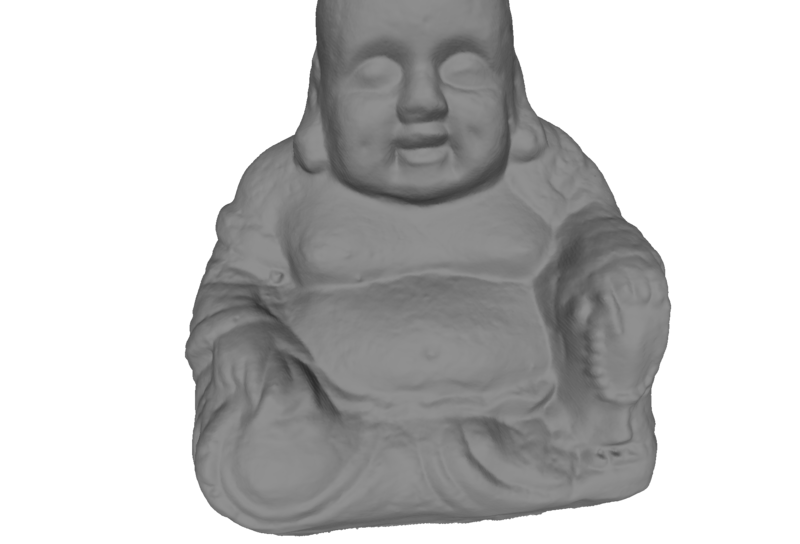} &
    \includegraphics[width=0.15\textwidth]{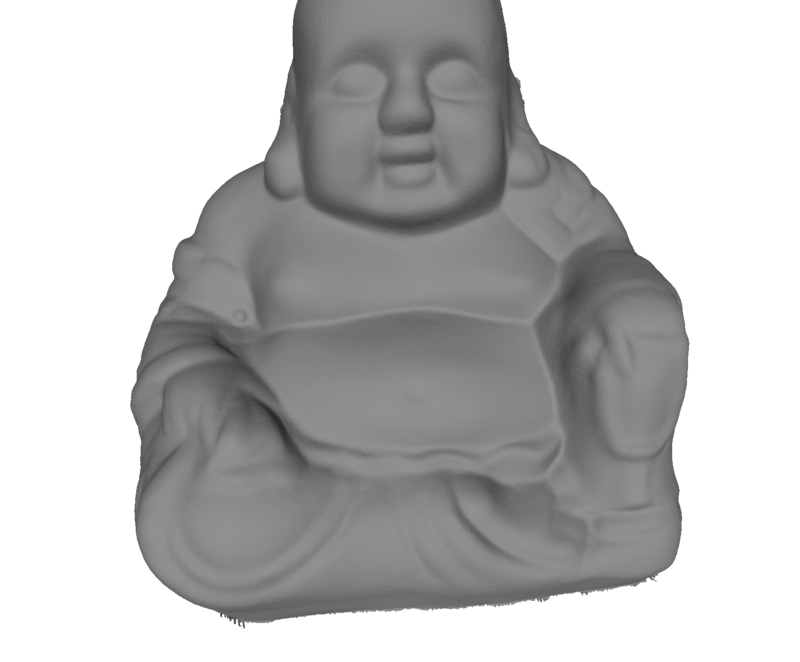} &
    \includegraphics[width=0.15\textwidth]{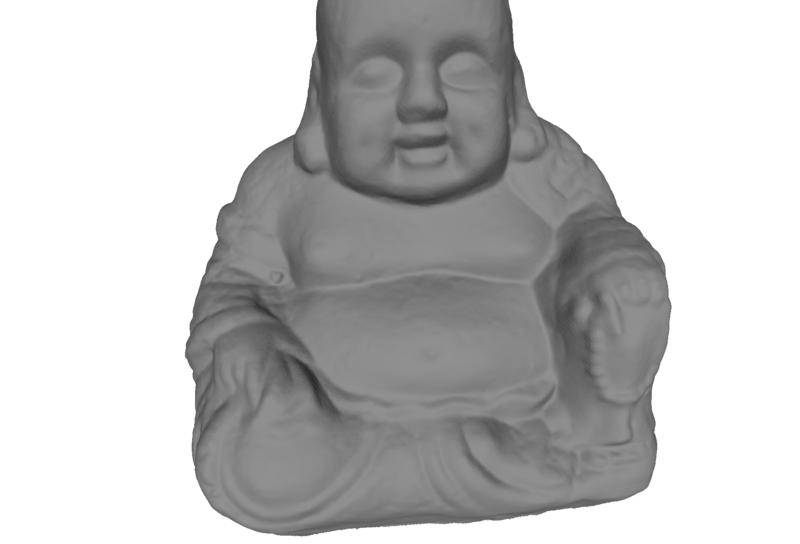} &
    \includegraphics[width=0.15\textwidth]{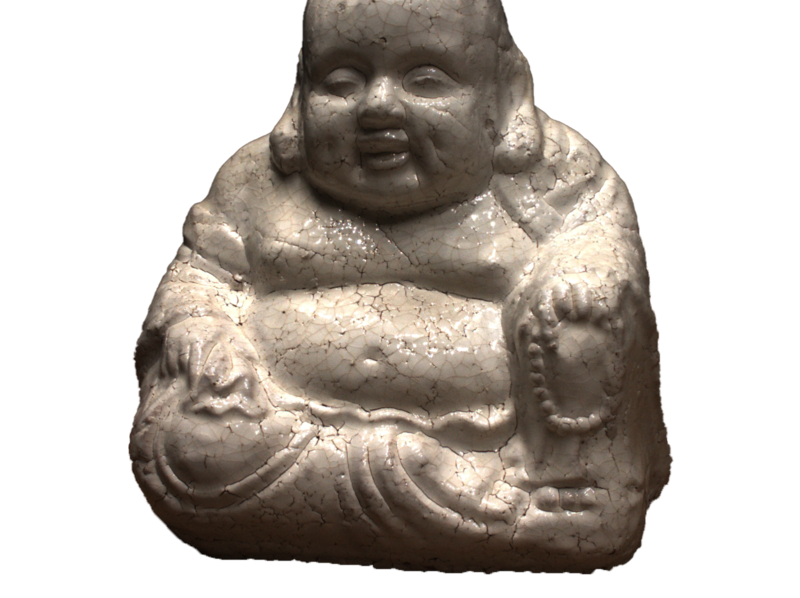} \\

    \includegraphics[width=0.15\textwidth]{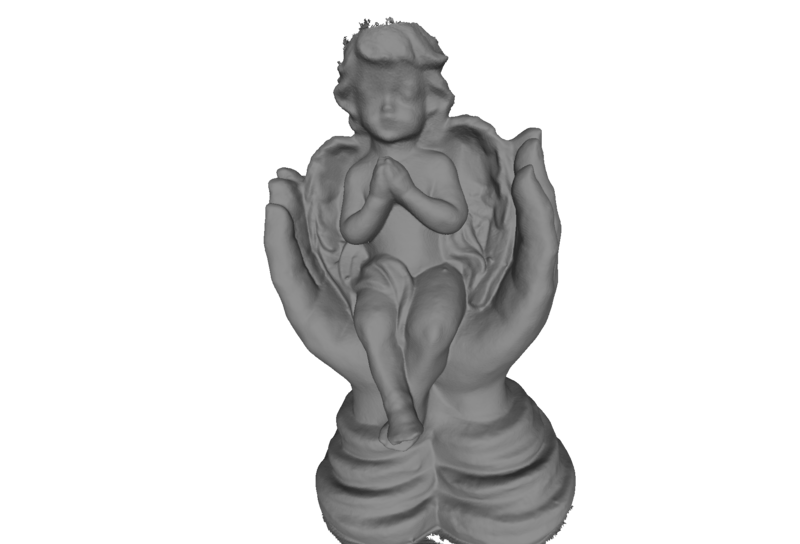} &
    \includegraphics[width=0.15\textwidth]{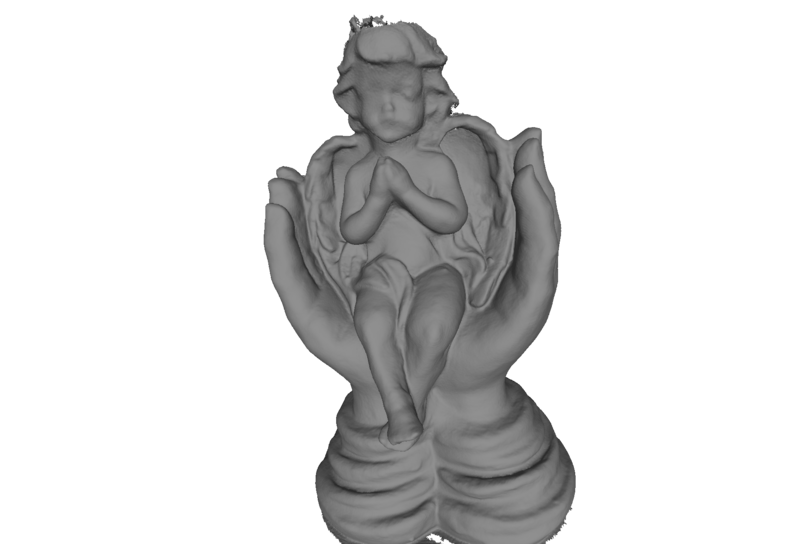} &
    \includegraphics[width=0.15\textwidth]{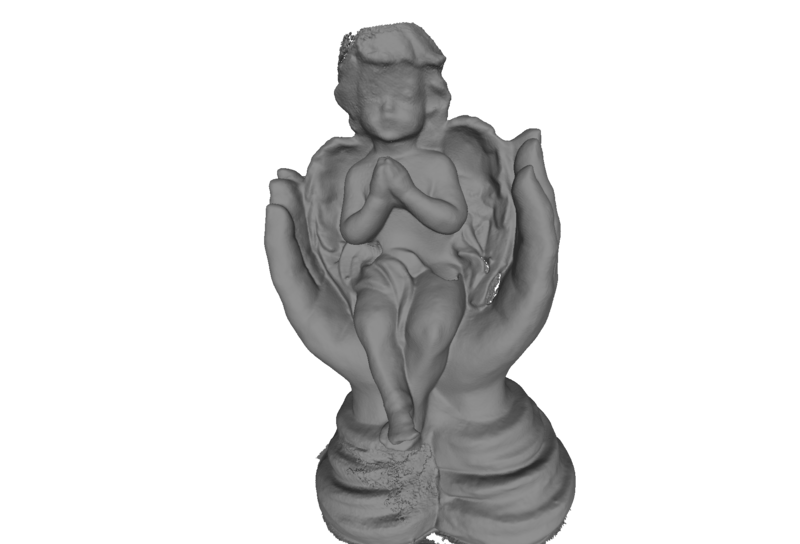} &
    \includegraphics[width=0.15\textwidth]{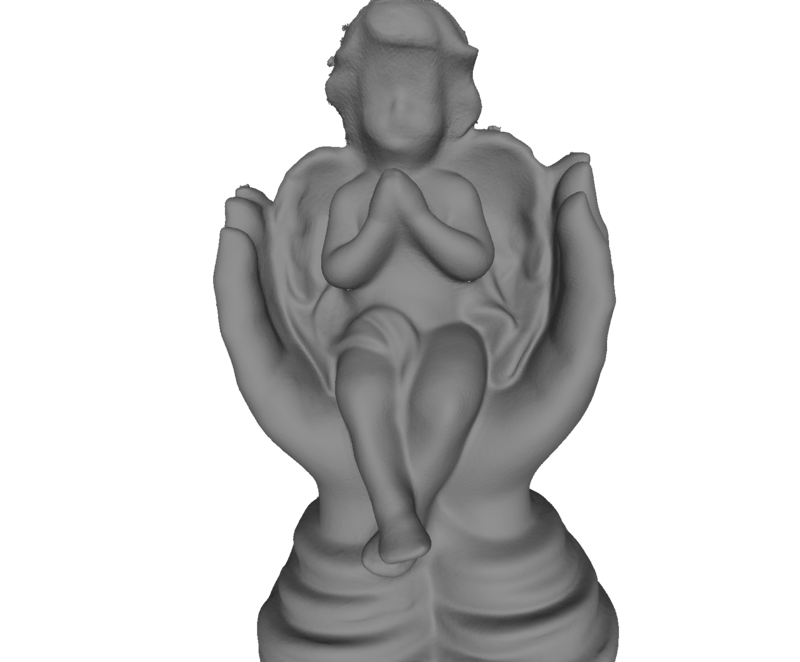} &
    \includegraphics[width=0.15\textwidth]{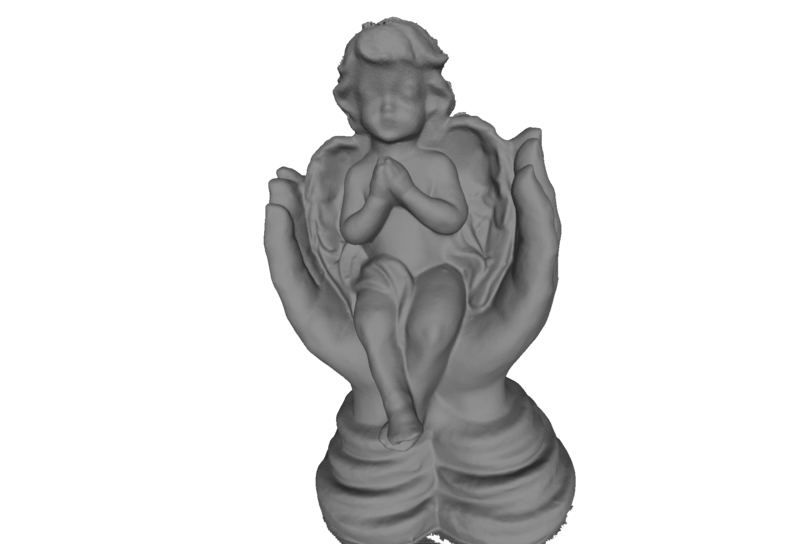} &
    \includegraphics[width=0.15\textwidth]{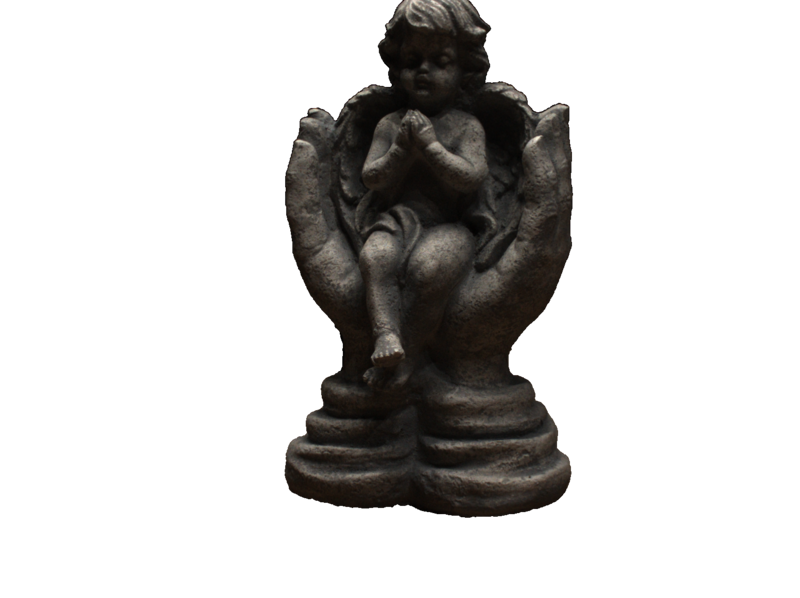} \\

    \includegraphics[width=0.15\textwidth]{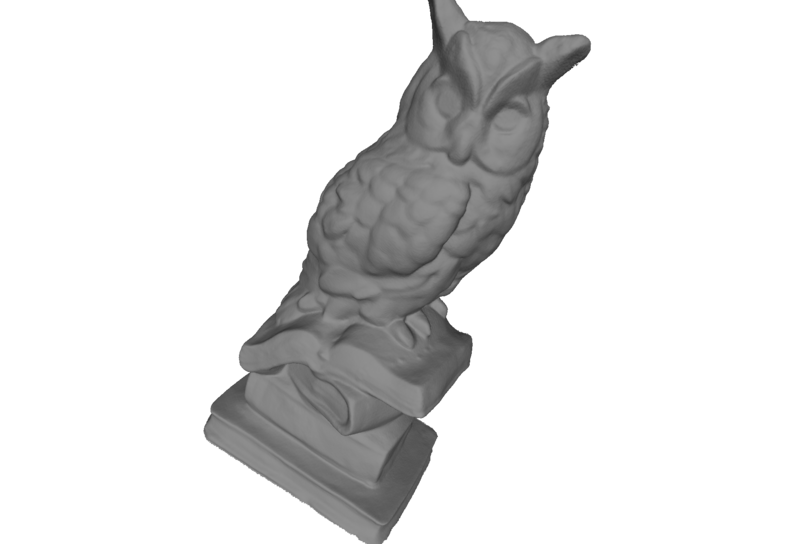} &
    \includegraphics[width=0.15\textwidth]{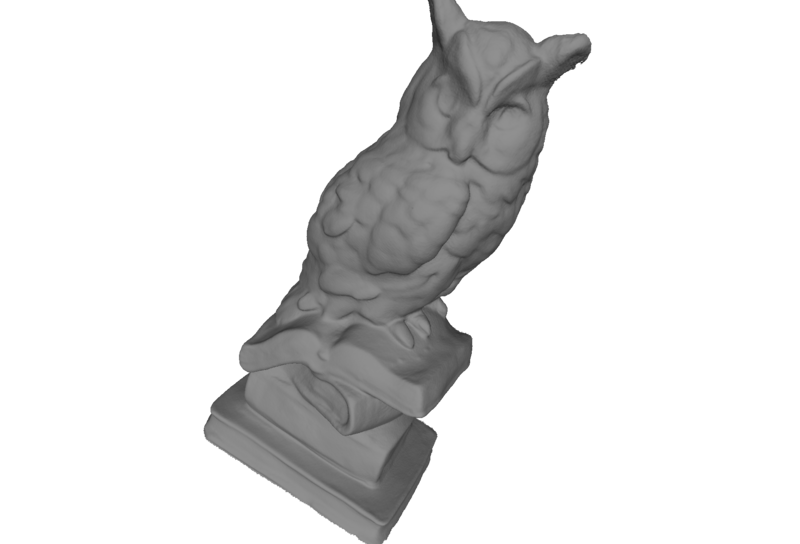} &
    \includegraphics[width=0.15\textwidth]{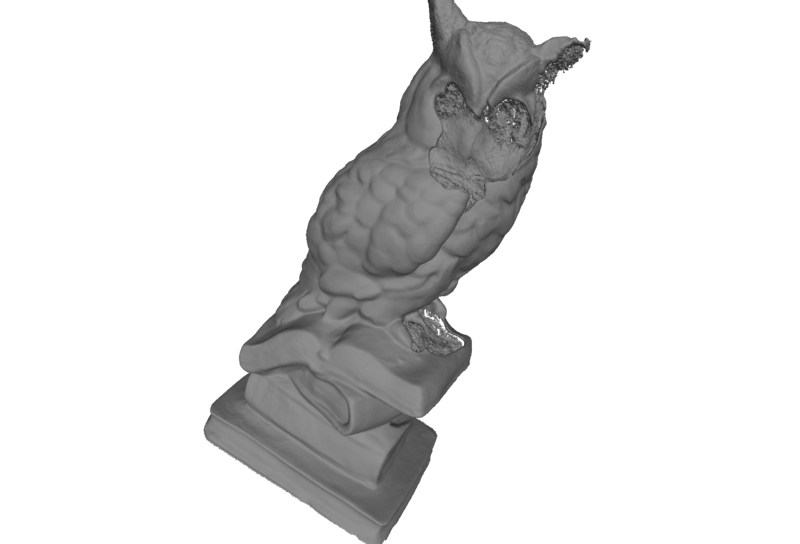} &
    \includegraphics[width=0.15\textwidth]{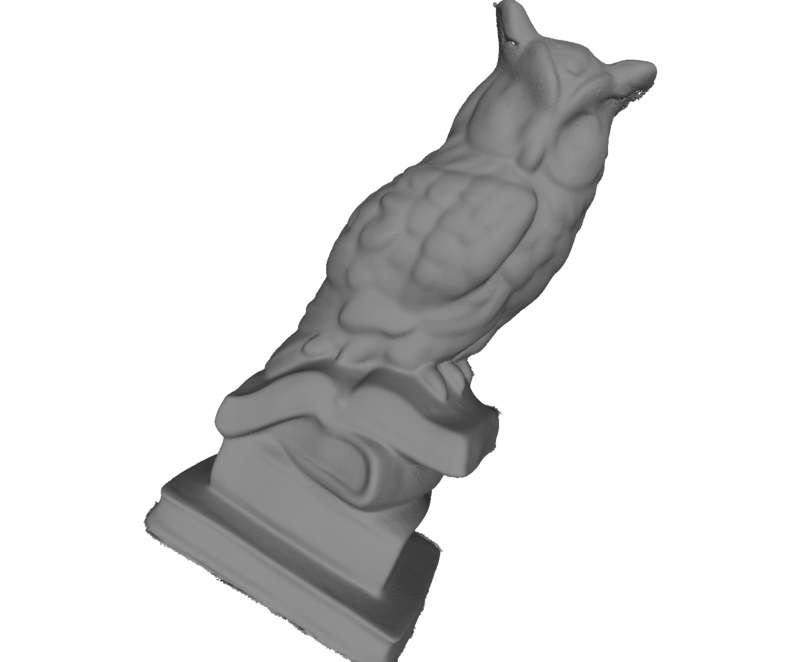} &
    \includegraphics[width=0.15\textwidth]{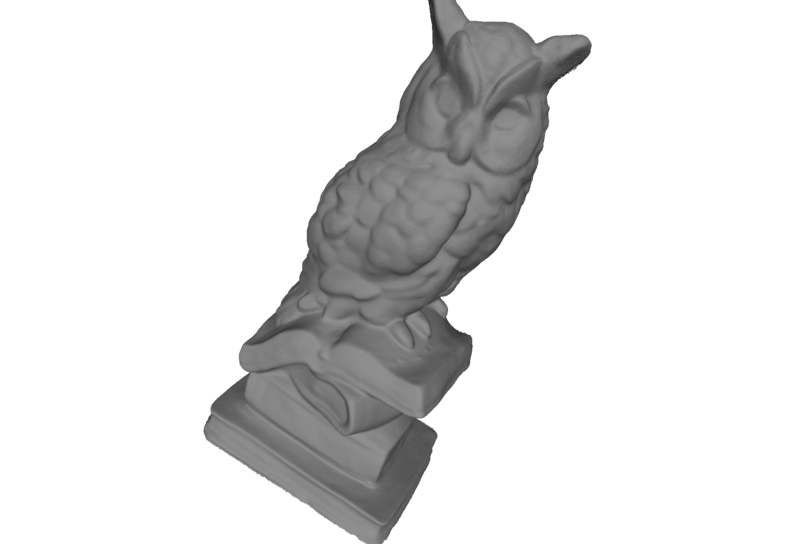} &
    \includegraphics[width=0.15\textwidth]{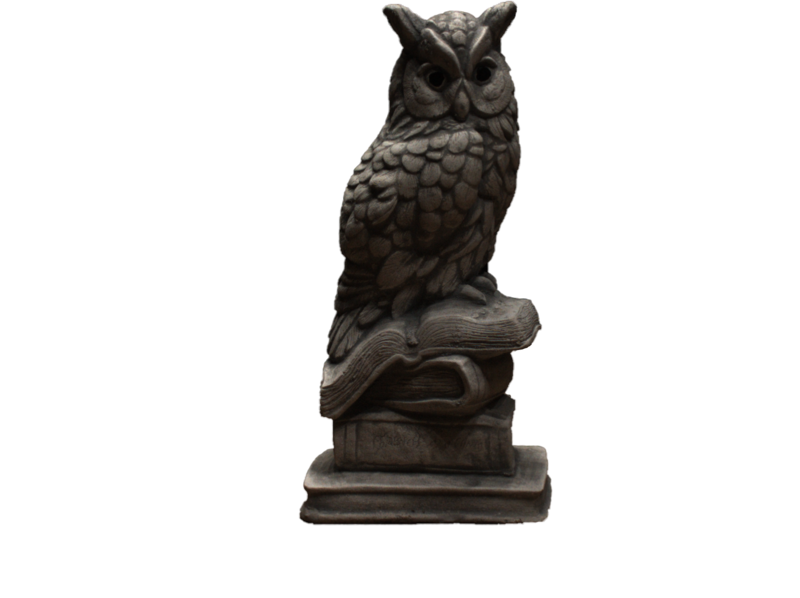} \\
    
    \end{tabular}
    \caption{Qualitative results on DTU from scan114 to scan122.}
    \label{supp:fig:DTU2}
\end{figure*}

\begin{figure*}
\setlength{\tabcolsep}{1pt} 
    \renewcommand{\arraystretch}{1} 
    \centering
    \begin{tabular}{cccccc}
    PGSR & +SN & \makecell[tc]{+VGGT\\(w/o conf)} & Ours & GT & GT image\\
    \includegraphics[width=0.15\textwidth]{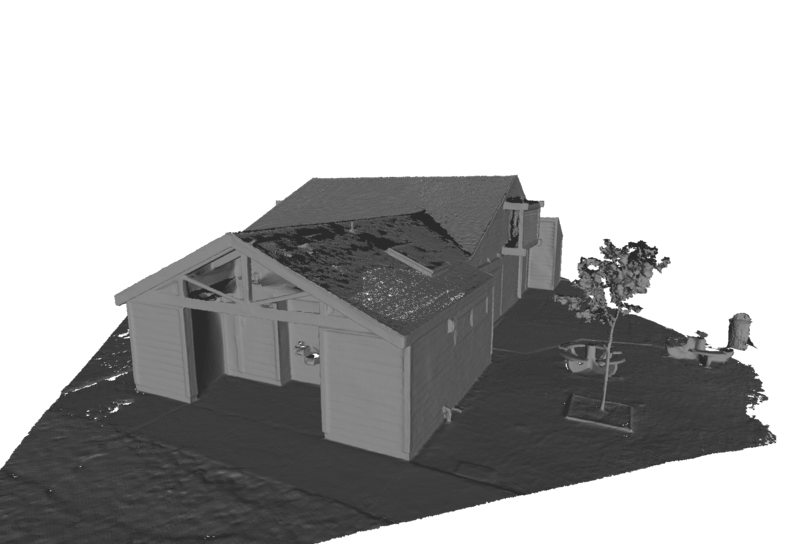} &
    \includegraphics[width=0.15\textwidth]{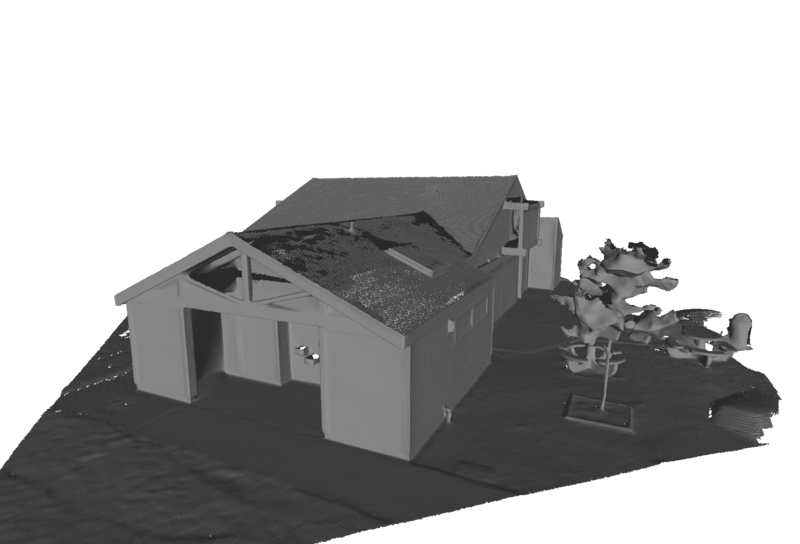} &
    \includegraphics[width=0.15\textwidth]{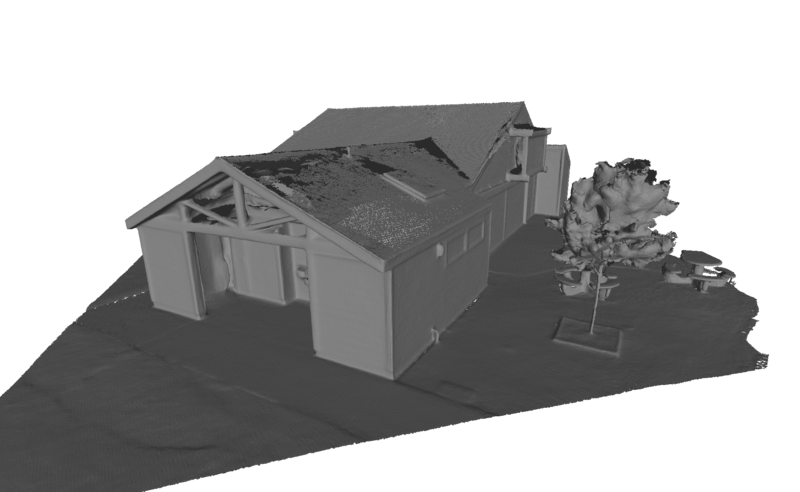} &
    \includegraphics[width=0.15\textwidth]{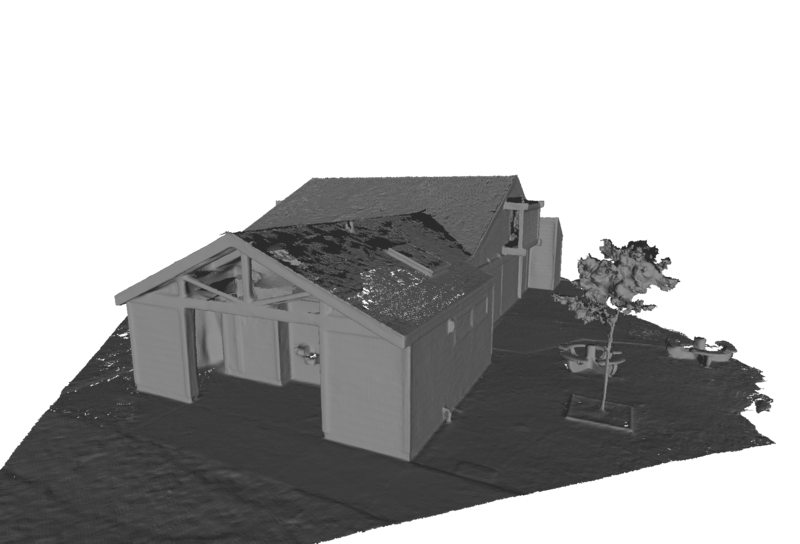} &
    \includegraphics[width=0.15\textwidth]{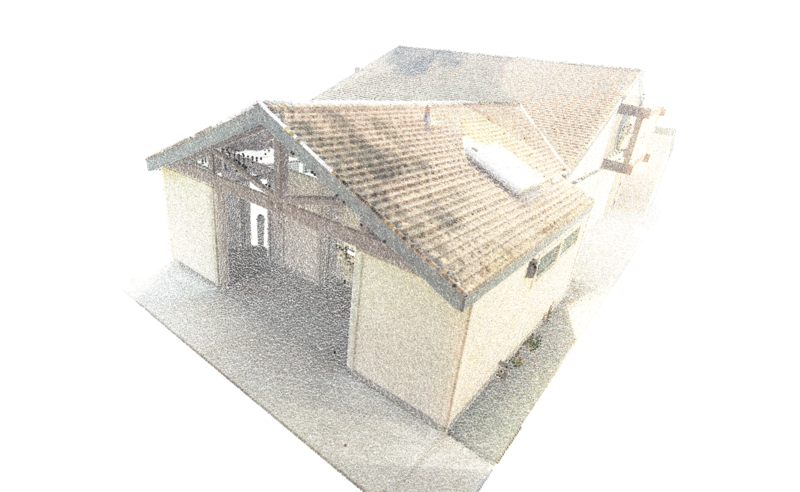} &
    \includegraphics[width=0.15\textwidth]{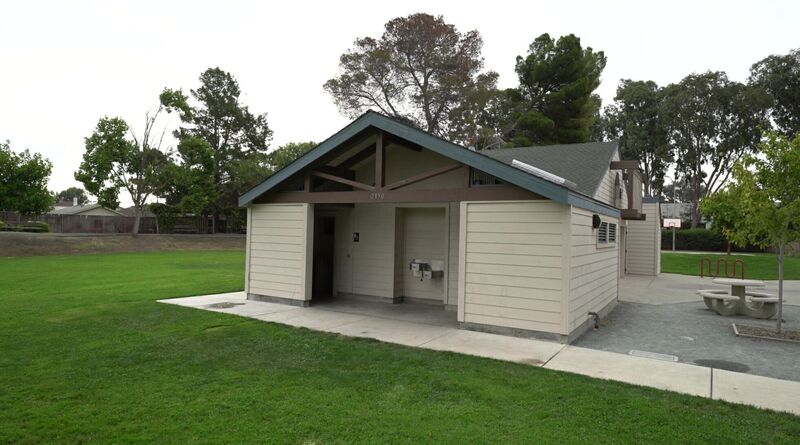} \\

    \includegraphics[width=0.15\textwidth]{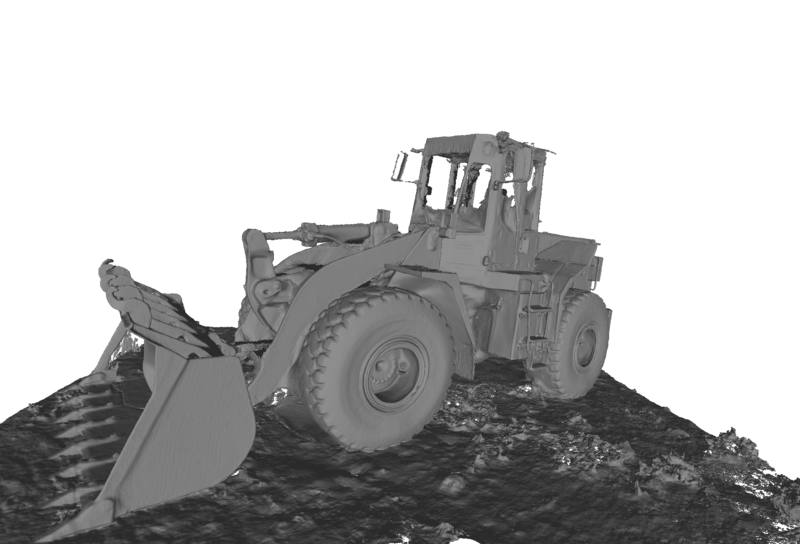} &
    \includegraphics[width=0.15\textwidth]{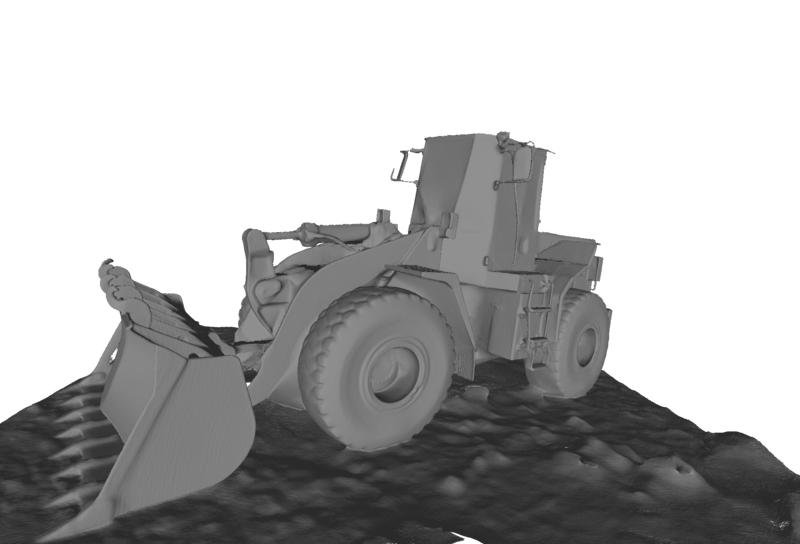} &
    \includegraphics[width=0.15\textwidth]{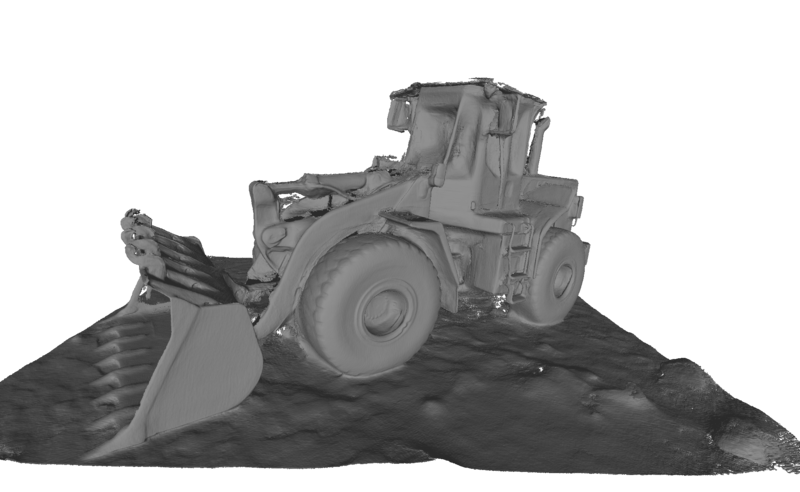} &
    \includegraphics[width=0.15\textwidth]{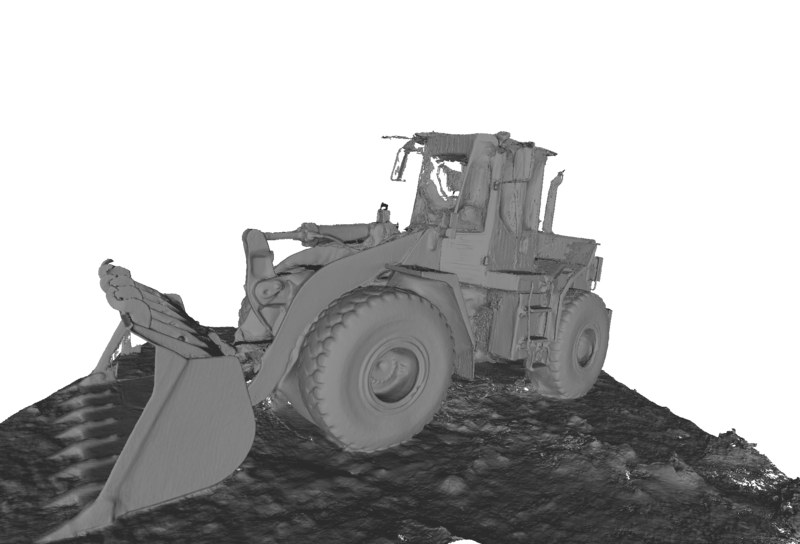} &
    \includegraphics[width=0.15\textwidth]{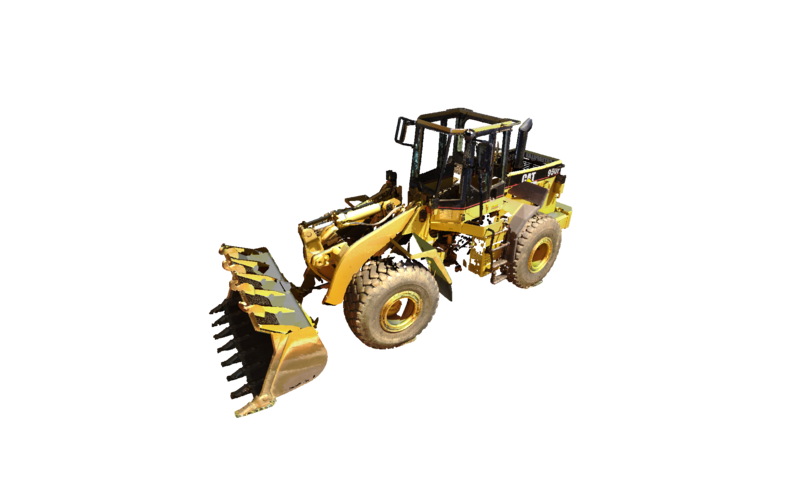} &
    \includegraphics[width=0.15\textwidth]{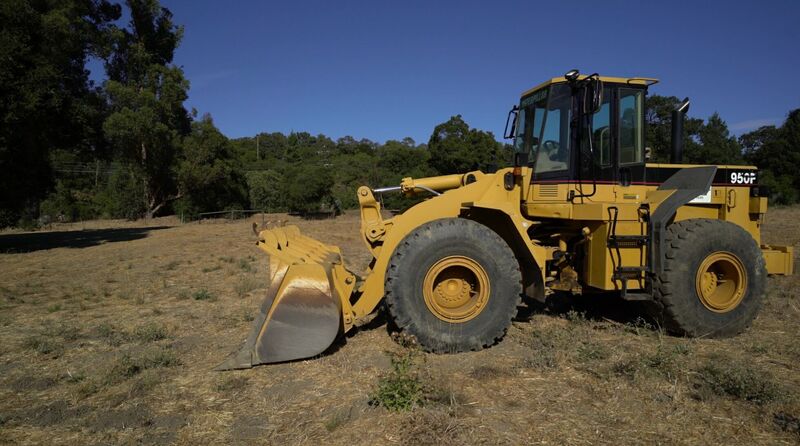} \\

    \includegraphics[width=0.15\textwidth]{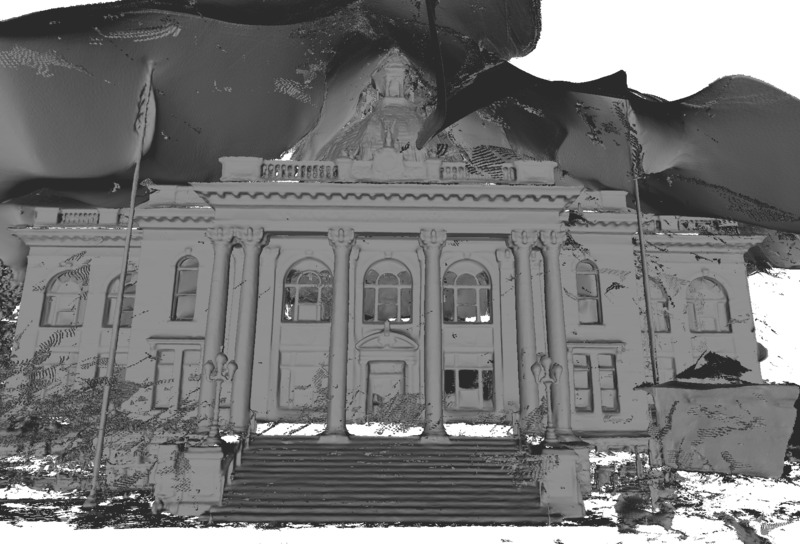} &
    \includegraphics[width=0.15\textwidth]{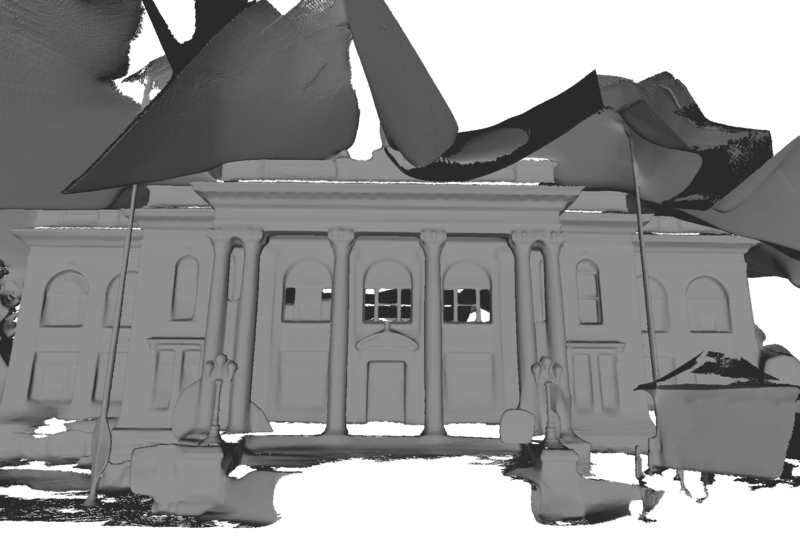} &
    \includegraphics[width=0.15\textwidth]{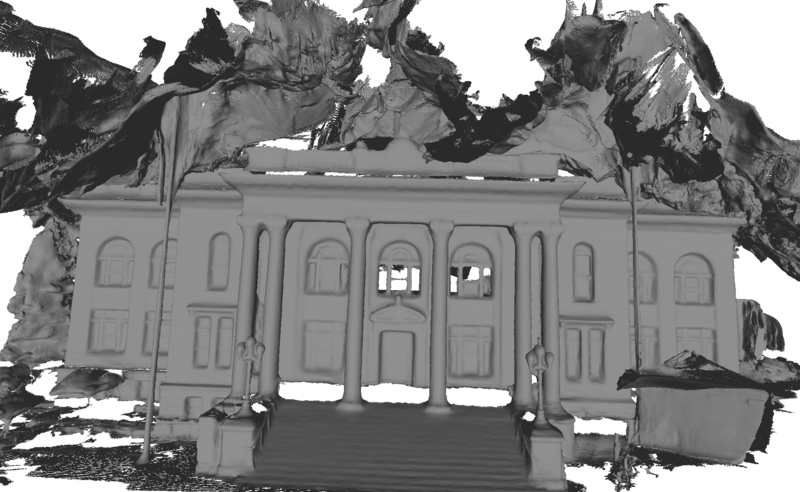} &
    \includegraphics[width=0.15\textwidth]{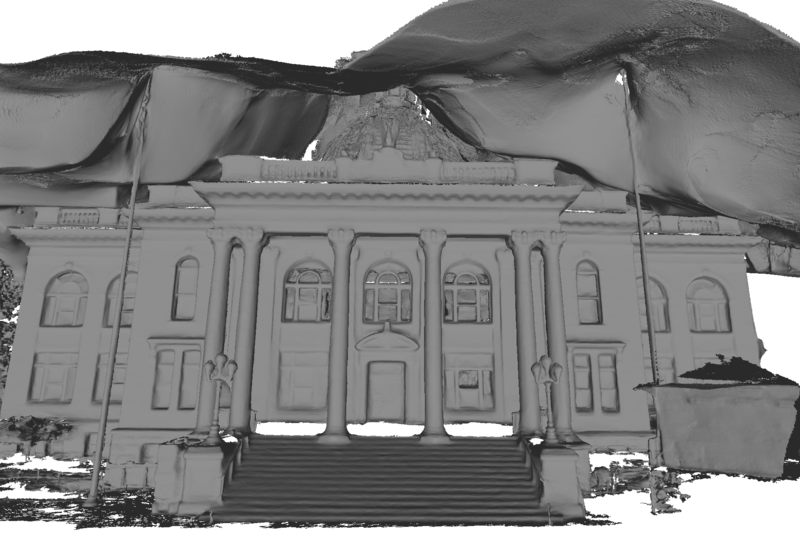} &
    \includegraphics[width=0.15\textwidth]{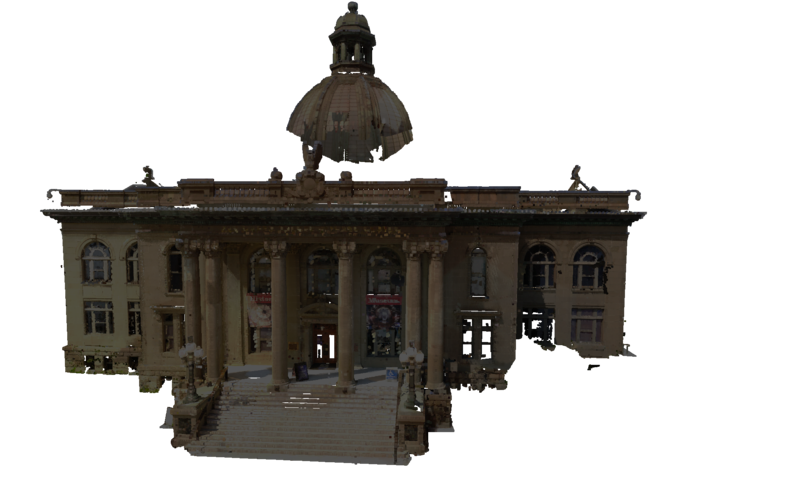} &
    \includegraphics[width=0.15\textwidth]{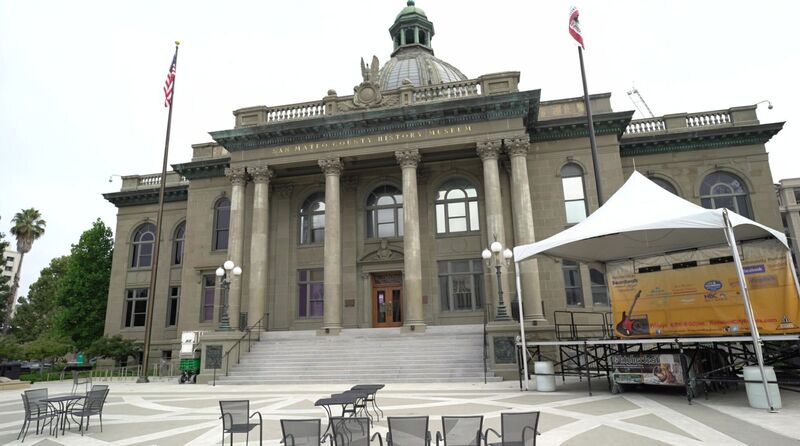} \\

    \includegraphics[width=0.15\textwidth]{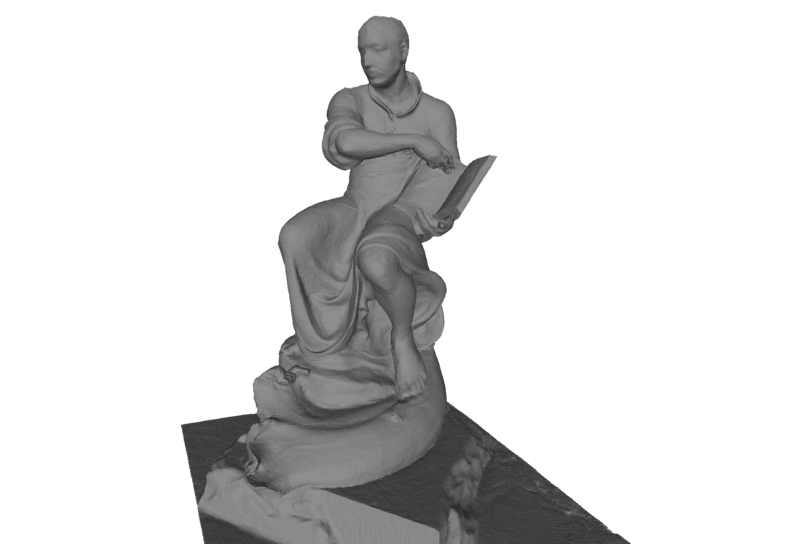} &
    \includegraphics[width=0.15\textwidth]{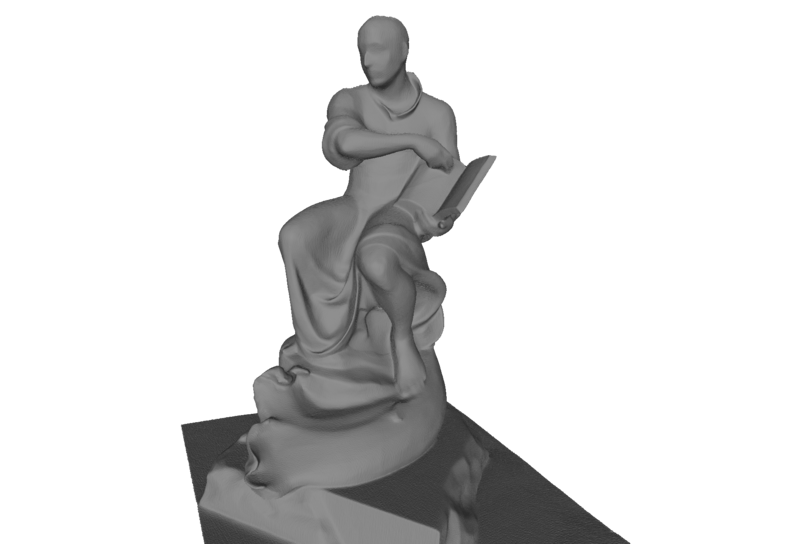} &
    \includegraphics[width=0.15\textwidth]{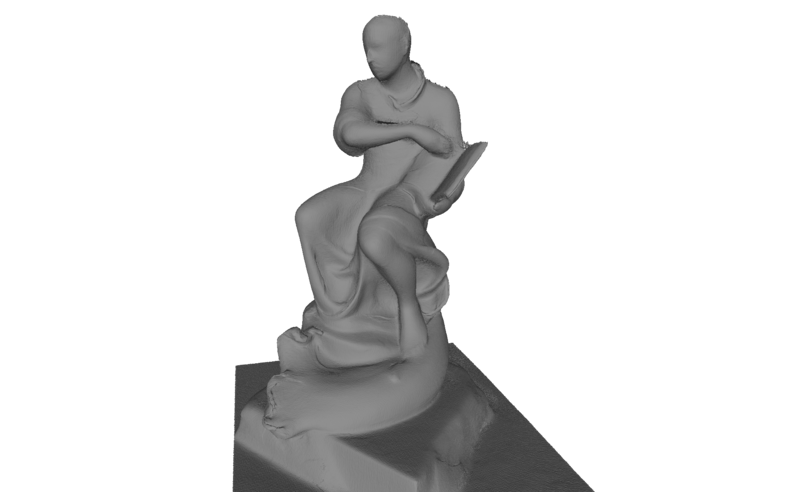} &
    \includegraphics[width=0.15\textwidth]{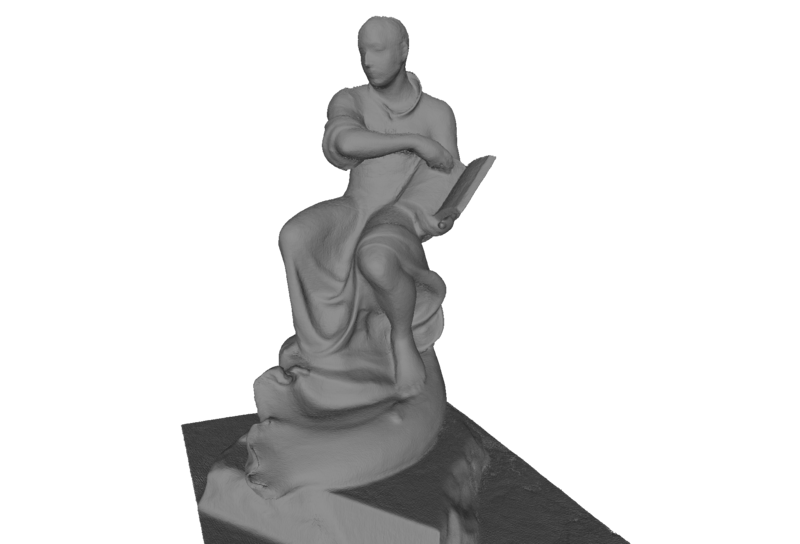} &
    \includegraphics[width=0.15\textwidth]{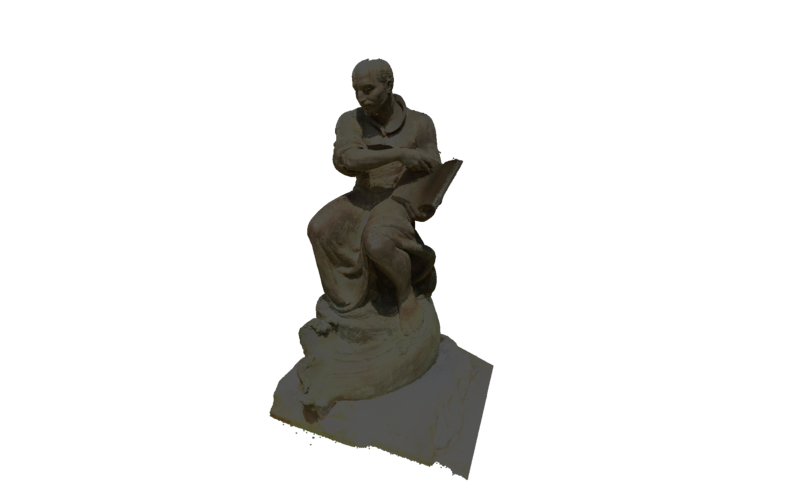} &
    \includegraphics[width=0.15\textwidth]{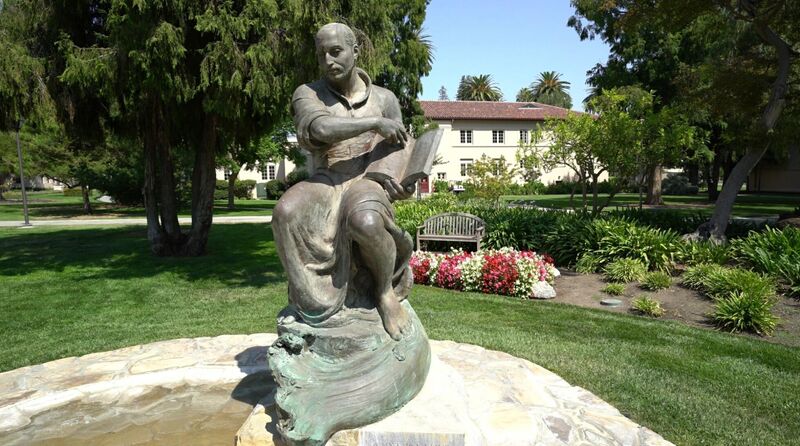} \\

    \includegraphics[width=0.15\textwidth]{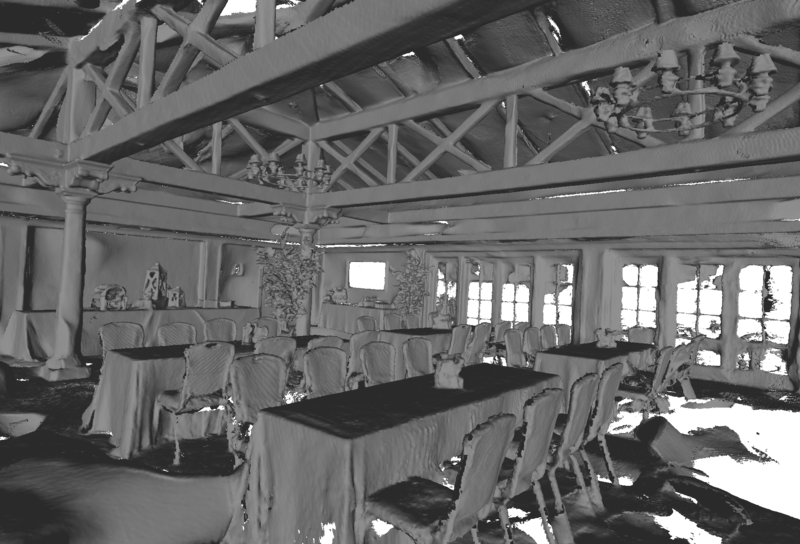} &
    \includegraphics[width=0.15\textwidth]{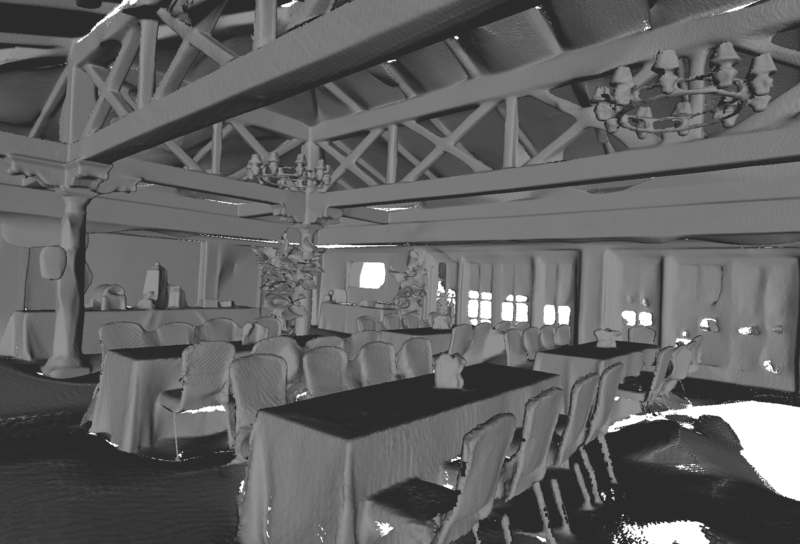} &
    \includegraphics[width=0.15\textwidth]{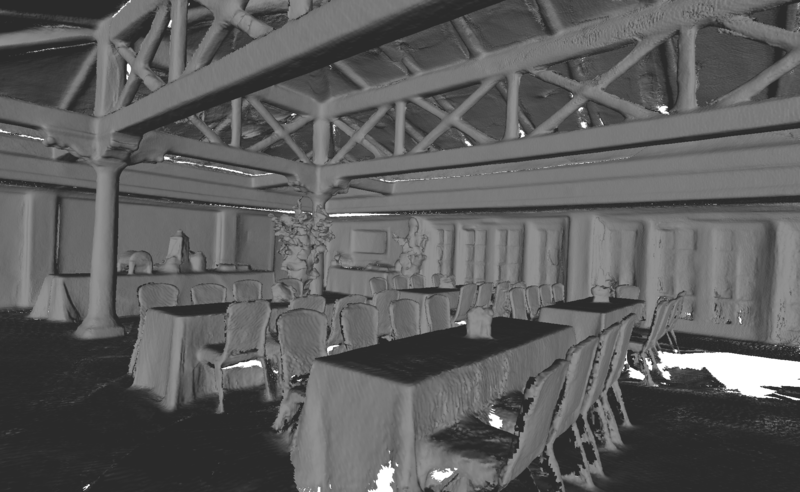} &
    \includegraphics[width=0.15\textwidth]{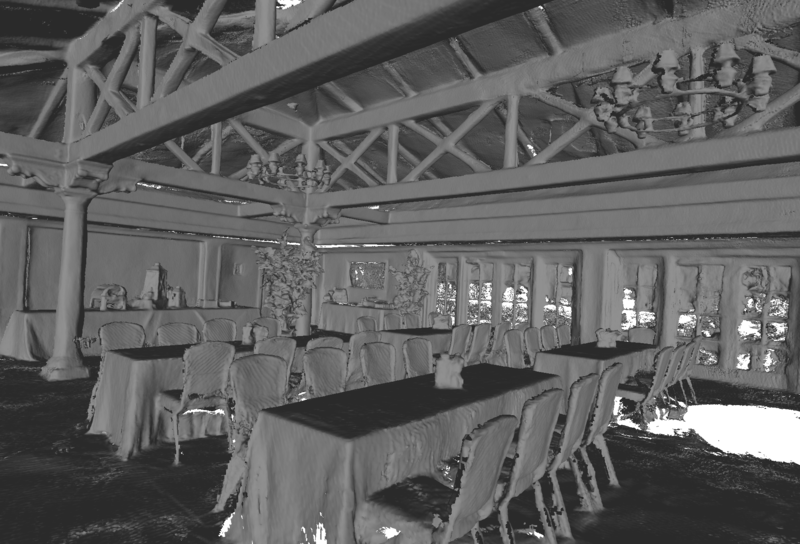} &
    \includegraphics[width=0.15\textwidth]{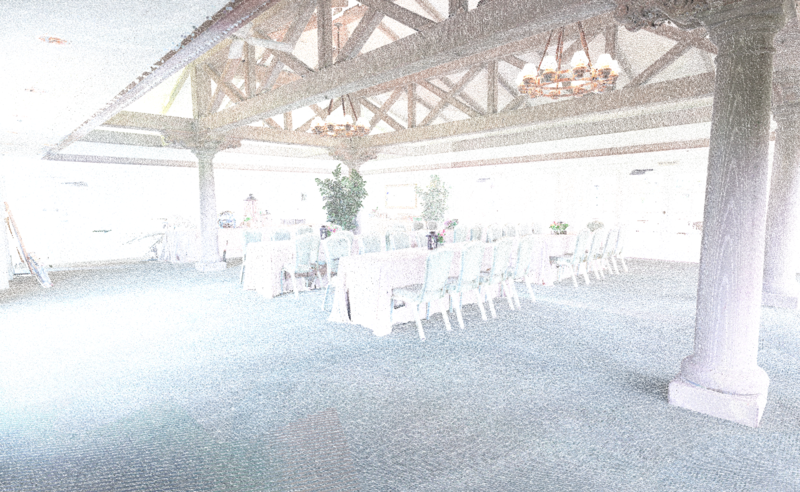} &
    \includegraphics[width=0.15\textwidth]{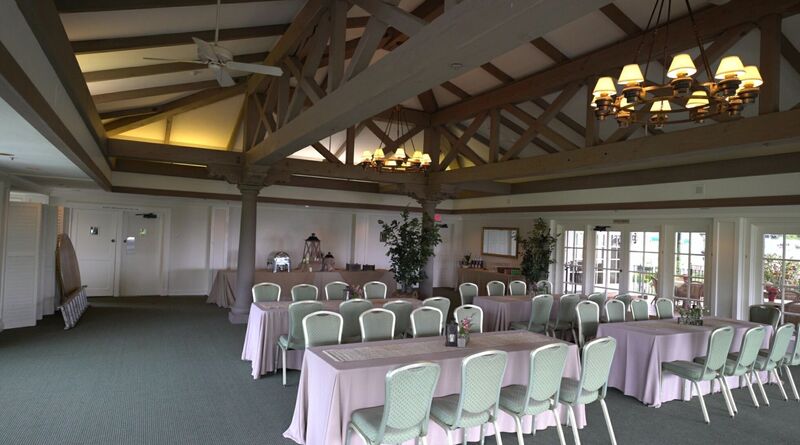} \\
    \end{tabular}
    \caption{Qualitative results on TnT.}
    \label{supp:fig:TnT}
\end{figure*}

\begin{figure*}
\setlength{\tabcolsep}{1pt} 
    \renewcommand{\arraystretch}{1} 
    \centering
    \begin{tabular}{ccccc}
    PGSR & +SN &  +DA & Ours & GT\\
    \includegraphics[width=0.18\textwidth]{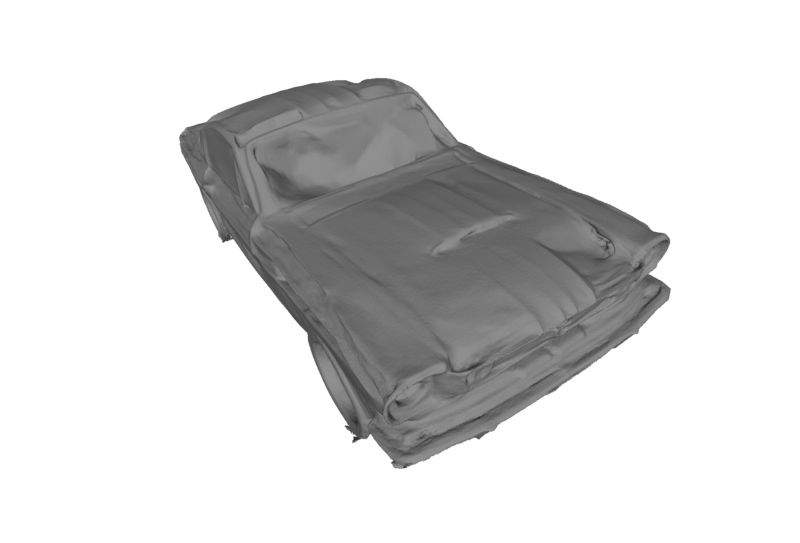} &
    \includegraphics[width=0.18\textwidth]{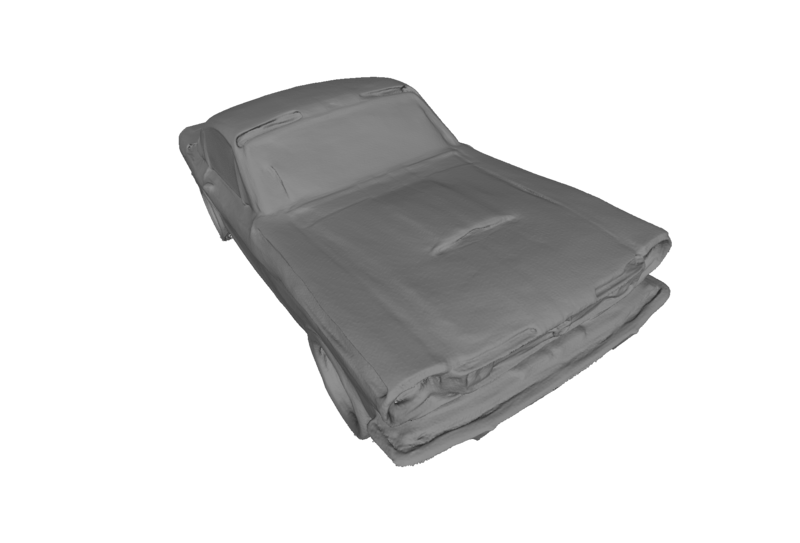} &
    \includegraphics[width=0.18\textwidth]{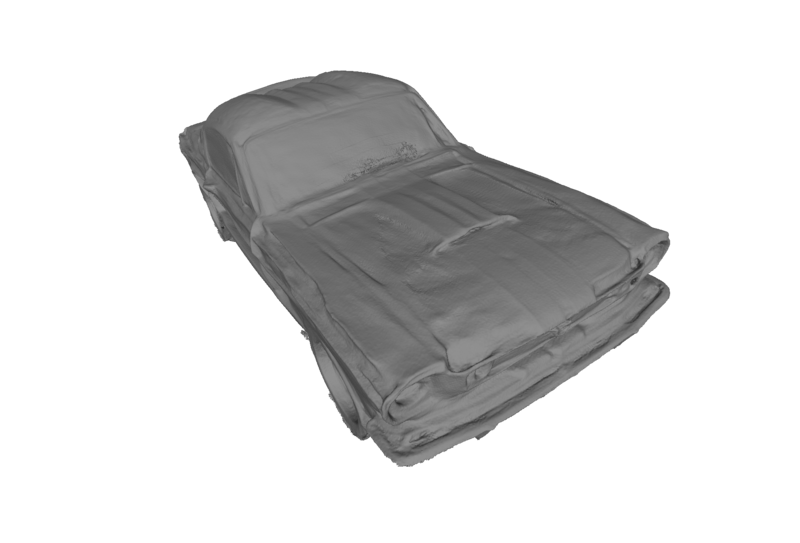} &
    \includegraphics[width=0.18\textwidth]{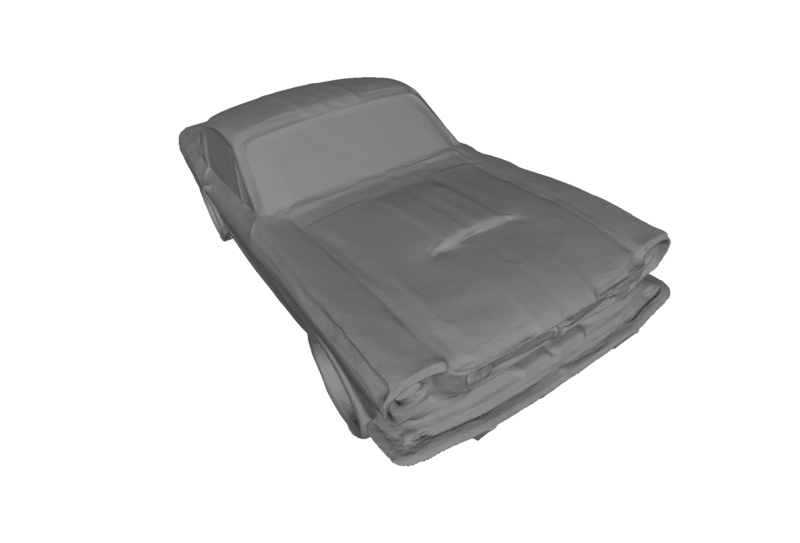} &
    \includegraphics[width=0.18\textwidth]{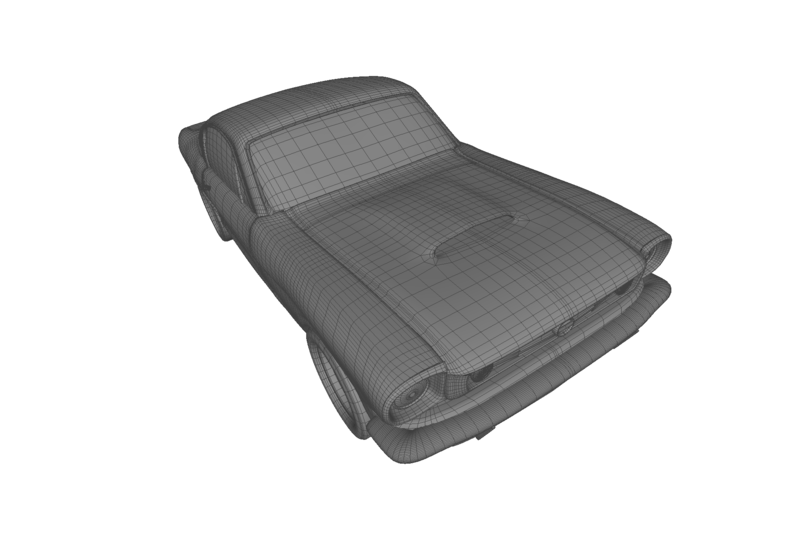}\\

    \includegraphics[width=0.18\textwidth]{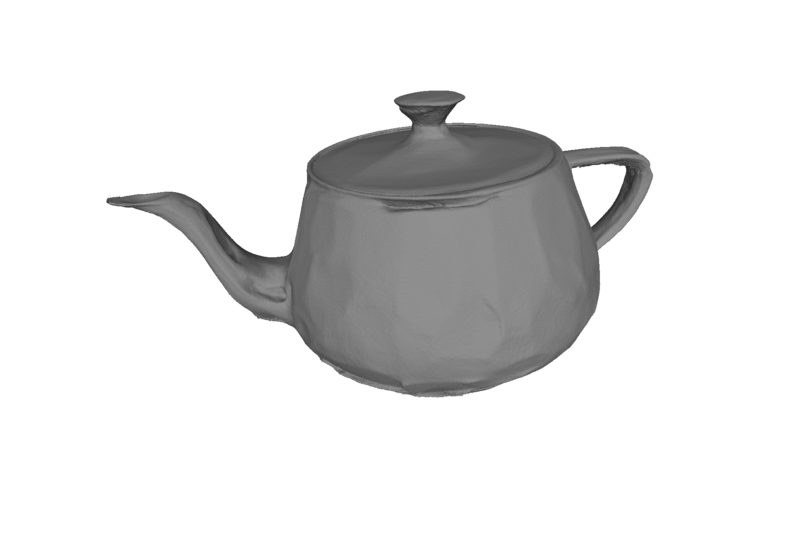} &
    \includegraphics[width=0.18\textwidth]{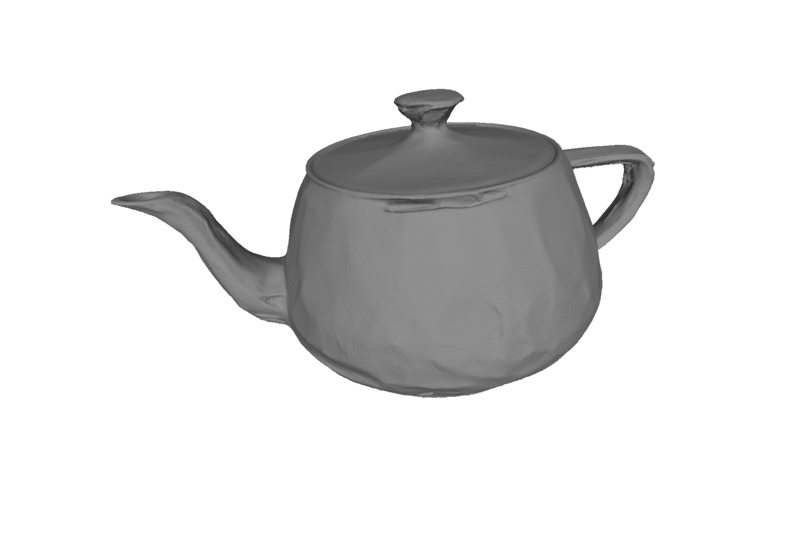} &
    \includegraphics[width=0.18\textwidth]{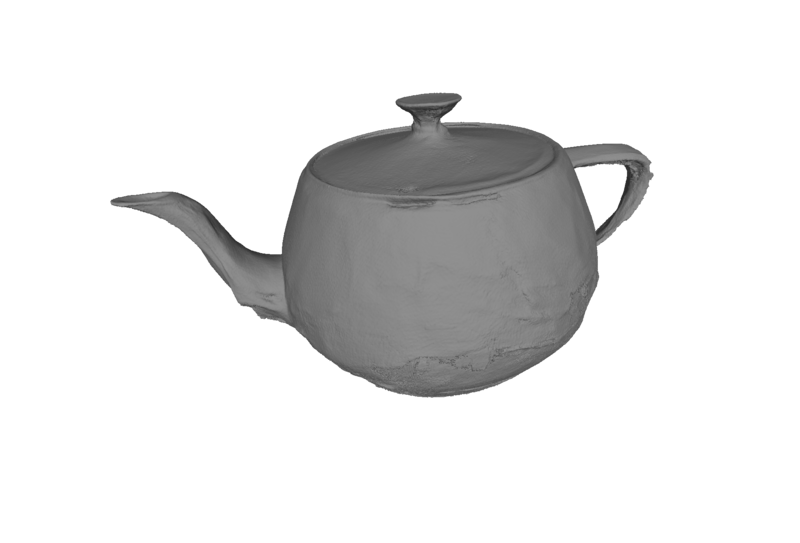} &
    \includegraphics[width=0.18\textwidth]{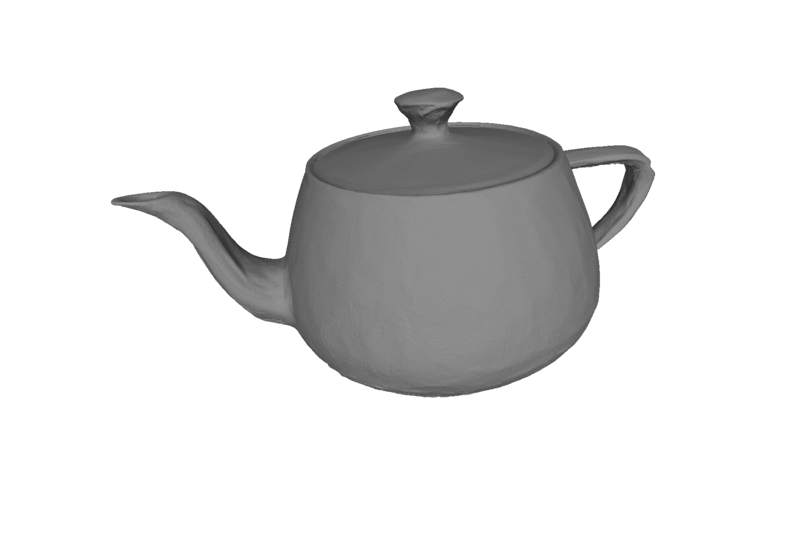} &
    \includegraphics[width=0.18\textwidth]{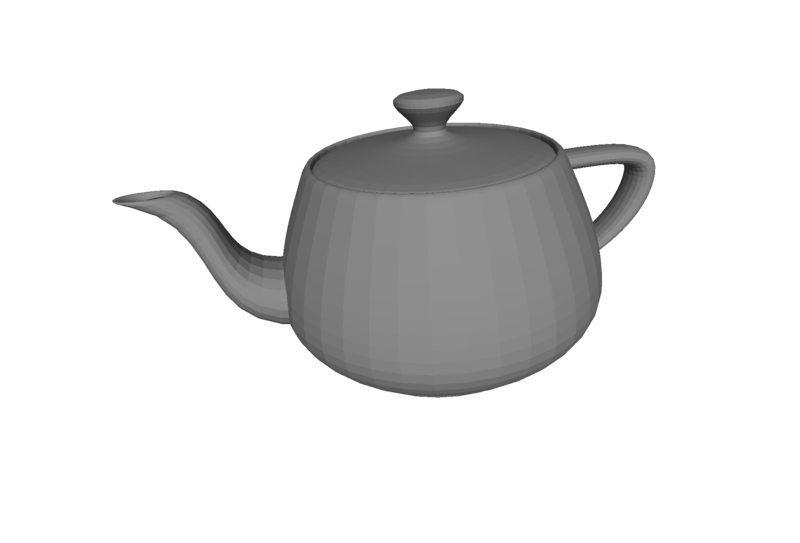}\\
    \end{tabular}
    \caption{Qualitative results on Shiny Blender.}
    \label{supp:fig:shiny blender}
\end{figure*}

We provide qualitative supplementary results on DTU~\cite{jensen2014large} in \cref{supp:fig:DTU1}, \cref{supp:fig:DTU2}, TnT~\cite{knapitsch2017tanks} in \cref{supp:fig:TnT}, and Shiny Blender~\cite{verbin2022refnerf} in \cref{supp:fig:shiny blender}. As for the TnT dataset, our qualitative performance looks very similar and sometimes even better compared to the baseline~(PGSR~\cite{chen2024pgsr}). For example, the geometry of the windows is predicted. The inside of the Barn is also predicted. The floor and the wall of the meeting room are flatter in our results. The improvement on Shiny Blender is significant in \cref{supp:fig:shiny blender}.

\end{document}